\documentclass[acmlarge,screen]{acmart}

\AtBeginDocument{%
  \providecommand\BibTeX{{%
    \normalfont B\kern-0.5em{\scshape i\kern-0.25em b}\kern-0.8em\TeX}}}

\setcopyright{acmcopyright}
\copyrightyear{2020}
\acmYear{2020}
\acmDOI{10.1145/1122445.1122456}

\def\eg{e.g.,}
\def\ie{i.e.,}

\def\lrpeps{LRP$_\varepsilon$}

\usepackage{ulem}
\definecolor{clr}{RGB}{0, 0, 0}

\begin{document}

\title{On the Explanation of Machine Learning Predictions in Clinical Gait Analysis}

\author{Djordje Slijepcevic}
\authornote{Both authors contributed equally to this research.}
\email{Djordje.Slijepcevic@fhstp.ac.at}
\orcid{0000-0002-2295-7466}
\affiliation{%
  \institution{Institute of Creative Media Technologies, Department of Media \& Digital Technologies, St. P\"olten University of Applied Sciences}
  \city{St. P\"olten}
  \country{Austria}
}

\author{Fabian Horst}
\authornotemark[1]
\email{horst@uni-mainz.de}
\orcid{0000-0002-3299-5896}
\affiliation{%
  \institution{Department of Training and Movement Science, Institute of Sport Science, Johannes Gutenberg-University Mainz}
  \city{Mainz}
  \country{Germany}
}

\author{Brian Horsak}
\affiliation{%
  \institution{Institute of Health Sciences, Department of Health Sciences, St. P\"olten University of Applied Sciences}
  \city{St. P\"olten}
  \country{Austria}
}

\author{Sebastian Lapuschkin}
\orcid{0000-0002-0762-7258}
\affiliation{%
  \institution{Department of Video Coding \& Analytics, Fraunhofer Heinrich Hertz Institute}
  \city{Berlin}
  \country{Germany}
}

\author{Anna-Maria Raberger}
\affiliation{%
  \institution{Institute of Health Sciences, Department of Health Sciences, St. P\"olten University of Applied Sciences}
  \city{St. P\"olten}
  \country{Austria}
}

\author{Wojciech Samek}
\affiliation{%
  \institution{Department of Video Coding \& Analytics, Fraunhofer Heinrich Hertz Institute}
  \city{Berlin}
  \country{Germany}
}

\author{Christian Breiteneder}
\affiliation{%
  \institution{Institute of Visual Computing and Human-Centered Technology, TU Wien}
  \city{Vienna}
  \country{Austria}
}

\author{Wolfgang Immanuel Sch\"ollhorn}
\affiliation{%
  \institution{Department of Training and Movement Science, Institute of Sport Science, Johannes Gutenberg-University Mainz}
  \city{Mainz}
  \country{Germany}
}

\author{Matthias Zeppelzauer}
\affiliation{%
  \institution{Institute of Creative Media Technologies, Department of Media \& Digital Technologies, St. P\"olten University of Applied Sciences}
  \city{St. P\"olten}
  \country{Austria}
}

\renewcommand{\shortauthors}{Slijepcevic and Horst, et al.}

\begin{abstract}
    
    \textcolor{clr}{Machine learning (ML) is increasingly used to support decision-making in the healthcare sector. While ML approaches provide promising results with regard to their classification performance, most share a central limitation, namely their black-box character. 
    Motivated by the interest to understand the functioning of ML models, methods from the field of \textit{Explainable Artificial Intelligence} (XAI) have recently become important. This article investigates the usefulness of XAI methods in \textit{clinical gait classification}. For this purpose, predictions of state-of-the-art classification methods are explained with an established XAI method, \ie~Layer-wise Relevance Propagation (LRP). We propose to evaluate the obtained explanations with two complementary approaches: a statistical analysis of the underlying data using Statistical Parametric Mapping and a qualitative evaluation by a clinical expert. A gait dataset comprising ground reaction force measurements from 132 patients with different lower-body gait disorders and 62 healthy controls is utilized. We investigate several gait classification tasks, employ multiple classification methods, and analyze the impact of data normalization and different signal components for classification performance and explanation quality. Our experiments show that explanations obtained by LRP exhibit promising statistical properties concerning inter-class discriminativity and are also in line with clinically relevant biomechanical gait characteristics. 
    }

\end{abstract}



\maketitle

\section{Introduction}
Artificial Intelligence~(AI) and machine learning~(ML) techniques have become almost ubiquitous in our daily lives by supporting or guiding our decisions and providing recommendations. \textcolor{clr}{Impressively, there are certain medical tasks, such as the detection of skin or breast cancer, that ML approaches have already been able to solve more efficiently and effectively than humans~\citep{haenssle2018man,mckinney2020international,Esteva.2017}.} Therefore, it is not surprising that ML approaches are currently becoming popular in the healthcare sector~\citep{topol_medicine_2019}. This trend has also been recognized in the field of clinical gait analysis (CGA)~\citep{schollhorn_applications_2004,figueiredo_automatic_2018}. CGA focuses on the quantitative description and analysis of human gait from a kinematic (\ie~joint angles), kinetic (\ie~ground reaction forces and joint moments), and muscular (\ie~electromyographic activity) point of view. Thereby, CGA produces a vast amount of data~\citep{phinyomark_analysis_2018,halilaj_machine_2018}, which are difficult to comprehend due to their multi-dimensional and multi-correlated nature~\citep{chau_review_2001,wolf_automated_2006}. In the last years, ML methods have been successfully employed in CGA for the classification of patient groups~\citep{schollhorn_applications_2004,figueiredo_automatic_2018} such as stroke~\citep{lau_support_2009}, Parkinson's disease~\citep{wahid_classification_2015}, cerebral palsy~\citep{van_gestel_probabilistic_2011}, multiple sclerosis~\citep{alaqtash2011automatic}, osteoarthritis~\citep{nuesch_gait_2012}, and patients suffering from different functional gait disorders~\citep{slijepcevic2017automatic}. While ML approaches yield promising results regarding classification performance, most share a central limitation, which is their black-box character~\citep{adadi_peeking_2018}. \textcolor{clr}{This means that even if the underlying mathematical principles in these methods are understood, it is often unclear why a particular prediction has been made and if meaningfully grounded patterns have led to this prediction.} Additionally, the black-box character also hinders ML approaches to provide justifications of their predictions. This is, however, necessary for compliance with legislation such as the General Data Protection Regulation (GDPR, EU 2016/679)~\citep{regulation2016regulation, adadi_peeking_2018, He_practical_2019}. These factors currently limit the application of ML-based decision-support systems in medical practice~\citep{holzinger_what_2017,samek_explaining_2017}.

Due to the aforementioned reasons, the field of \textit{Explainable Artificial Intelligence} (XAI) gained increasing attention in recent years. 
\textcolor{clr}{Different approaches have been proposed~(see Section~\ref{sec:relatedwork}: Related work). In general, XAI methods intend to illustrate how complex and non-linear ML models operate and how they produced their predictions. 
However, explanation is understood in the sense of providing more differentiated insights into the behaviour of ML models in order to fathom the dependence of the results on input variables (without claiming to give causation).
}
Even though research in XAI is still in an early stage, the application of such approaches in medicine has already raised attention~\citep{holzinger_what_2017,tjoa_survey_2019}. \textcolor{clr}{The motivation is to increase the traceability and trust of medical professionals in ML approaches~\citep{holzinger_causability_2019}. However, application of XAI methods to the field of CGA remains to be investigated.} A first step in that direction has recently been taken by Horst et al.~\cite{horst2019explaining} for explaining predictions in gait-based person recognition.

\textcolor{clr}{The primary aim of this article is to investigate to which extent XAI methods can help to explain ML predictions in the context of CGA. For this purpose, we compare the results of several classification models trained on different gait classification tasks and analyze the quality of explanations, which are derived from the trained models by an established XAI method, \ie~Layer-wise Relevance Propagation (LRP). The assessment of explanations' quality is, however, a challenge since no ground truth exists for automatically generated explanations. 
In contrast to images, which are more frequently subject to explainability studies~\cite{samek2016evaluating,fong2017interpretable,adebayo2018sanity,samek2020toward}, the evaluation of explanations becomes particularly challenging when the input signals are more abstract and thus not straightforward to interpret, as often is the case with biomedical signals.} 
\textcolor{clr}{Recently, it has been shown that XAI approaches do not necessarily refer to the actual prediction of the classification model and sometimes even build upon unrelated information~\cite{adebayo2018sanity}. Thus, a more comprehensive investigation of explanations obtained by XAI methods is necessary to verify whether they are meaningful and justified.}

\textcolor{clr}{
To account for the above-mentioned challenges, we suggest a two-step approach for the evaluation of the obtained explanations. First, we analyze the discriminatory power of the obtained explanations from a statistical perspective. For this purpose, we leverage Statistical Parametric Mapping (SPM)~\cite{pataky_generalized_2010} -- a method building upon random field theory -- to derive statistical measures along with the input signals and thereby investigate how statistically justified the obtained explanations are.
Second, an experienced domain expert interprets the explainability results from a clinical perspective, to verify whether obtained explanations match characteristics from clinical practice.
}

Our investigation focuses on \textcolor{clr}{two} leading research questions:

\begin{enumerate}
    
    \item \textcolor{clr}{Which input features or signal regions are most relevant for automatic gait classification?}

    \item \textcolor{clr}{To what extent are input features or signal regions identified as being relevant for a given gait classification task statistically justified and in line with clinical assessment?}
    
\end{enumerate}
\textcolor{clr}{
In addition to these two leading questions, we investigate several further aspects that may influence classification performance as well as explainability in more detail, including the influence of different classification methods, the impact of data normalization, and the role of different input signal components (\ie~the horizontal forces, measurements of the affected leg and measurements of the unaffected leg).}

\textcolor{clr}{
We perform our investigation on the {\sc GaitRec} dataset~\cite{horsak_gaitrec_2020}, which contains ground reaction force measurements from clinical practice. We design prediction models for different gait classification tasks and derive possible explanations from the resulting models that are based on relevance scores. These relevance scores are directly related to specific regions in the input signal. Subsequently, we analyze the explanations from a statistical as well as a clinical perspective. The results show that explanations share promising statistical properties concerning class discriminativity and thus indicate that predictions are grounded on statistically justified information for the task. Further, we show that input features considered as relevant can also be interpreted as meaningful and clinically relevant biomechanical gait characteristics. Overall, our investigation demonstrates the usefulness of XAI in the domain of gait classification, exemplifies how to apply XAI methods to gait measurement data, and suggest approaches to evaluate their quality. The performed study suggests that XAI methods can be useful to better understand and interpret automatic predictions in clinical gait analysis.}

\section{Related Work}
\label{sec:relatedwork}

Methods from XAI can be grouped according to the type of explanation they provide. We distinguish between XAI approaches for (i) \textbf{data exploration}, (ii) \textbf{prediction explanation} and (iii) \textbf{model explanation} based on an adaptation of the taxonomy introduced by Arya et al.~\cite{arya2019one}. In the following, we briefly introduce the three different types of approaches and their capabilities.

\textbf{Data exploration} includes methods from the fields of visual analytics, statistics and unsupervised machine learning. As such, the methods are not capable of explaining a model but rather the data on which the model is trained. These methods focus on projecting the data into a space where it is possible to find meaningful structures or clusters and thus understand the data in more detail. A popular approach for data exploration introduced by Maaten and Hinton~\cite{maaten2008visualizing} is T-distributed Stochastic Neighbor Embedding (t-SNE), which projects high-dimensional data into a lower-dimensional and visualizable space. The projection is performed in a way that the cluster structure in the original data space is optimally exposed. Thereby, an understanding of the data and the identification of typical patterns and clusters in the data is facilitated. Other approaches in this category are visual analytics approaches that employ advanced techniques for the interactive visualization of data to support data exploration, \ie~finding characteristic patterns or dependencies within data~\citep{wagner_KAVAG,wilhelm2015furyexplorer}.

\textbf{Prediction explanation} aims at explaining the local behavior of a model, \ie~the prediction for a given input instance. For a classification task, these methods can provide, for example, explanations about which part of the input influenced the classifier's prediction the most. For classification of gait data, the explanation should highlight all relevant signal regions and characteristic signal shapes in the input data, which are associated with a particular gait disorder. Two main categories can be distinguished for explaining the local behavior of a machine learning model: i) \textit{self-explaining} models and ii) \textit{post-hoc} methods.

Self-explaining models integrate components that learn relationships between input data and predictions during training. Simultaneously, they learn how these relationships relate to terms from a predefined dictionary and consequently generate explanations from them. A self-explaining approach which does not visually highlight relevant regions in input data but generates textual explanations was proposed by Hendricks et al.~\cite{hendricks2016generating}. This self-explaining model combines a Convolutional Neural Network (CNN) and a Recurrent Neural Network (RNN). The CNN learns discriminative features to perform a classification task, while the RNN generates textual explanations of the prediction. This approach cannot be applied to a previously trained model in a post-hoc manner, which limits the practical applicability of such approaches.

Post-hoc methods provide much greater applicability as they can be applied to already trained models. These methods can be further categorized into i) propagation-based, ii) perturbation-based, and iii) Shapley-value-based methods. \textit{Propagation-based methods} determine the contributions of each input feature by (back-)propagating some quantity of interest from the model's output layer to the input layer. Sensitivity Analysis~\citep{zurada1994sensitivity} has been introduced to Support Vector Machines~(SVM)~\citep{baehrens2010explain} and CNNs~\citep{simonyan2013deep} in the form of saliency maps. Layer-wise Relevance Propagation (LRP)~\citep{bach2015pixel,montavon2019layer} and Deep Learning Important FeaTures (DeepLIFT)~\citep{shrikumar2017learning} are methods that propagate importance scores from the output layer back to the input, thereby enabling the identification of positive and negative evidences for a specific prediction. Sensitivity Analysis and the therewith obtained explanations, in general, suffer from the effects of shattered gradients~\citep{balduzzi2017shattered}, especially so in more complex (deeper) networks. Modern approaches to CNN explainability, such as LRP or DeepLift, do not have this problem and work well for a wider range of network architectures and models in general~\citep{montavon2018methods,kohlbrenner2019towards}. \textit{Perturbation-based methods}, such as those introduced by Fong and Vedaldi~\cite{fong2017interpretable} or Zintgraf et al.~\cite{zintgraf2017visualizing}, treat the model as a black box and estimate the importance of input features by (partially) occluding the input and measuring the effect on the model output.
While some methods produce explanations directly from a perturbation process, others employ a learning component -- \eg~the Interpretable Model-agnostic Explanations (LIME) method~\citep{ribeiro2016model} -- to estimate locally interpretable surrogate models mimicking the prediction process of the black-box model. Perturbation-based methods can be considered to be model-agnostic, as they do not require access to internal model parameters or structures to operate. However, this model-agnosticism is bought at a considerable computational cost, compared to propagation-based approaches. \textit{Shapley-value-based methods} attempt to approximate the Shapley values of a given prediction. For this purpose, the effect of omitting an input feature is examined, taking into account all possible combinations of other input features, that can be included or excluded~\citep{vstrumbelj2014explaining}. Lundberg and Lee~\cite{lundberg2017unified} proposed the SHapley Additive exPlanations (SHAP) method, which is a unified approach building upon the theory of Shapley values and existing propagation-based and perturbation-based methods, \eg~LIME, DeepLIFT, and LRP.

\textbf{Model explanation} provides an interpretation of what a trained model has learned, \ie~the most characteristic representations or prototypes for an entire class are visualized (\eg~a class of gait disorders in CGA). These methods can indicate which classes overlap and point out ambiguous input features. In addition to saliency maps, Simonyan et al.~\cite{simonyan2013deep} proposed a method for generating a representative visualization for a specific class that was learned by a CNN. For this purpose, they applied activation maximization, \ie~starting with a blank image, each pixel is changed by utilizing backpropagation so that the activity of a neuron is increased. The resulting visualizations give a first impression about the patterns learned but are highly abstract and can only be interpreted to a limited extent. To generate visualizations that are easier to interpret, Nguyen et al.~\cite{nguyen2016synthesizing} proposed a method to constrain the optimization process by image priors that were learned automatically. Lapuschkin et al.~\cite{lapuschkin2019unmasking} proposed the Spectral Relevance Analysis (SpRAy) which summarizes a model's learned strategies by analyzing similarities and dissimilarities over large quantities of input relevance maps computed with respect to a category of interest.

\textcolor{clr}{For gait classification, prediction explanation is desirable to provide clinical experts with detailed information about which patterns in the input signals are important for a specific prediction. Additionally, based on aggregations of these explanations, differences between patient groups can be assessed. In this context, post-hoc methods are preferable because they provide a classifier-agnostic approach (can be applied to any classification model) and do not require retraining or additional labels. We, therefore, choose a established post-hoc explainability method, \ie~LRP, in our experiments.}

\section{Approach and Methodology}
\label{sec:approach}
\begin{figure}[ht!]
  \centering
	\includegraphics[width=\linewidth]{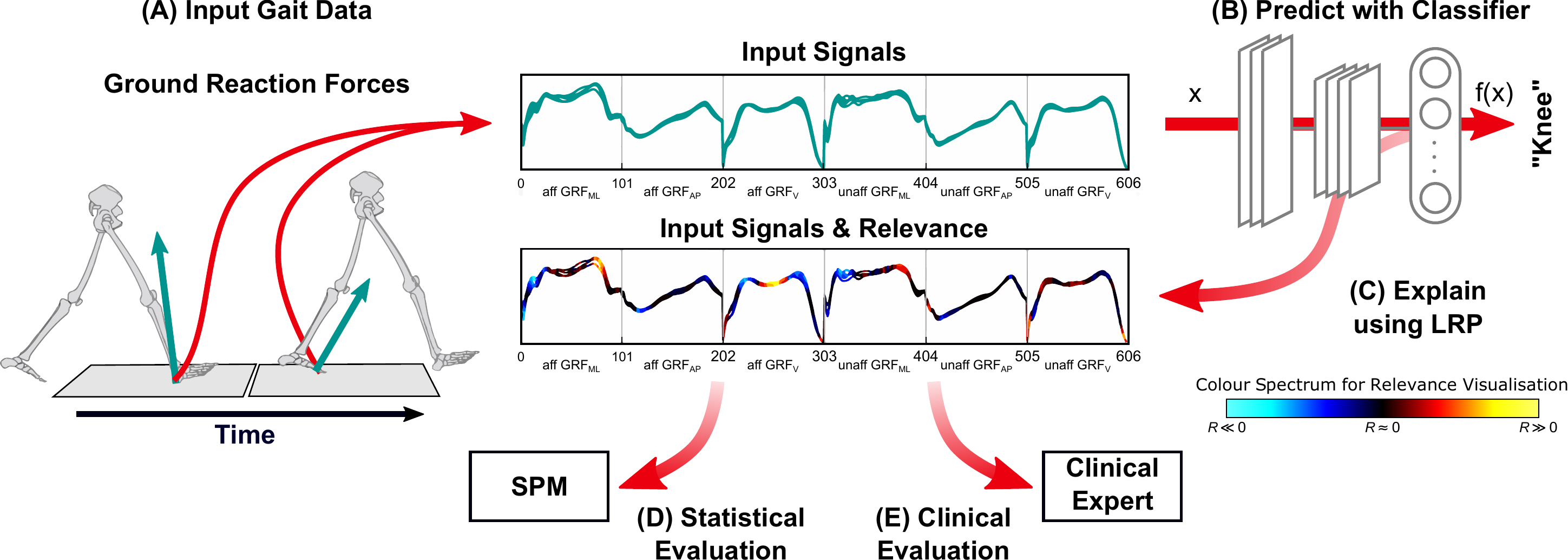}
    \caption{Overview of our proposed workflow for data acquisition, prediction and prediction explanation in automated gait classification, showing the data of subject 46 belonging to the knee disorder class. (A) The clinical gait analysis consists of five recordings of each subject walking barefoot (unassisted) a distance of 10~m at a self-selected walking speed. Two centrally-embedded force plates capture the three-dimensional ground reaction forces~(GRFs) during the stance phase of the right and left foot. (B) The GRF comprising the medio-lateral~($GRF_{ML}$), anterior-posterior~($GRF_{AP}$), and vertical~($GRF_{V}$) force components  of the affected and unaffected side are used as time-normalized and concatenated input vector $x$~(1$\times$606-dimensional) for the prediction of the knee disorder class using a classifier~(\eg~CNN). (C) Decomposition of input relevance scores is achieved using LRP. The color spectrum for the visualization of input relevance scores of the model predictions is shown in the bottom right corner. Black line segments are irrelevant to the model’s prediction. Warm hues identify input segments causing a prediction corresponding to the class label, while cool hues are features contradicting the class label. (D) Statistical and (E) Clinical evaluation of class-specific averaged relevance scores.}
    \label{img:overview_approach}
\end{figure}

\textcolor{clr}{
The general approach we followed in this study was to design and train classification models for automated gait classification tasks (see Figure~\ref{img:overview_approach}B) based on three-dimensional ground reaction forces~(GRFs) of both legs (see Figure~\ref{img:overview_approach}A), to explain the predictions of these models based on relevance scores that are related to the input signal space by using LRP~(see Figure~\ref{img:overview_approach}C), and to evaluate these results from a statistical (see Figure~\ref{img:overview_approach}D) and a clinical perspective (see Figure~\ref{img:overview_approach}E).     
}

\subsection{Gait Classification}
\textcolor{clr}{
The main task in automated gait classification is to determine whether a person has a healthy or pathological gait pattern based on gait measurements. We employed three-dimensional GRFs of the affected and unaffected side as input signals and investigated the classification performance of several state-of-the-art classification methods. Furthermore, the input signals were fed directly into the classification models. This ensures that the results of the employed explainability method (LRP) can be directly mapped to the original signals. 
For easier interpretation of the XAI results, we refrained from using data reduction techniques such as \eg~Principal Component Analysis (PCA), which are a common practice in automated gait classification~\cite{burdack2020,halilaj_machine_2018,slijepcevic2018p}.
}

\subsection{Prediction Explanation}
\textcolor{clr}{We employed Layer-wise Relevance Propagation (LRP) for prediction explanation~\cite{bach2015pixel} as a post-hoc method that provides explanations in the input space, which is the space where the signals are usually interpreted by experts in clinical practice.} LRP decomposes the prediction $f(x)$ of a learned function $f$ given an input vector $x$ into time- and component-resolved input relevance values $R_i$ for each individual input value $x_i$. This enables to explain the prediction of an ML model as partial contributions of an individual input value. LRP indicates which information a model uses to predict in favor or against an output class. Thereby, it enables the interpretation of input relevance scores and their dynamics as representation for a certain class (\ie~healthy controls or functional disorders in ankle, knee or hip).

\textcolor{clr}{For the explanation of predictions, we decomposed the input relevance scores of each gait trial with LRP. In order to analyse patterns learned for a specific class, we used LRP to decompose the ground truth label (and not the predicted value) of the trial. For the visualization of the explanations, we averaged the underlying GRF signals and the resulting input relevance scores over all trials of a class.} 

Given that the models investigated in this study are comparatively shallow and are largely unaffected by detrimental effects such as gradient shattering~\citep{balduzzi2017shattered,montavon2019layer}, we performed relevance decomposition according to \lrpeps~with $\varepsilon=10^{-5}$ in all layers across the different models (except for the CNN for which we employed the \textit{flat} rule at the input layer)~\citep{kohlbrenner2019towards}.

\subsection{\textcolor{clr}{Statistical Evaluation}}
\textcolor{clr}{To evaluate the derived relevance scores of LRP, we employ Statistical Parametric Mapping (SPM)~\cite{pataky_generalized_2010,pataky_one-dimensional_2012} which recently received increased attention in the gait analysis community~\cite{nieuwenhuys2017statistical,booth2018stapp}}. While standard inference statistical approaches tend to reduce time-continuous signals to single time-discrete values for statistical testing, SPM allows to use the entire time-continuous signals to make probabilistic conclusions. It follows the same notion and logic as classical inference statistics. The main advantages of SPM are that the statistical results are presented in the original sampling space and that there is no need for a (potentially biasing) parameterization technique~\citep{pataky_one-dimensional_2012,pataky_generalized_2010}. \textcolor{clr}{In comparison to the XAI methods, SPM is completely data-centric. Therefore, SPM is suited for our investigation, as an model-independent method to assess the quality of derived explanations. Since the LRP explanations and the results of SPM reside in the same space (the input signal space), we can leverage SPM to verify the meaningfulness of LRP explanations from a statistical point of view.}
For the statistical evaluation we compute independent \textit{t}-tests using the {\sc SPM1D}\footnote{
    {\sc SPM1D} v.0.4, \hyperlink{http://www.spm1d.org/}{http://www.spm1d.org/}
} 
package provided by Pataky~\cite{pataky_one-dimensional_2012} for Matlab and investigate differences between \textcolor{clr}{each GRF signal between two classes (for visualization purposes we concatenated the results obtained on each GRF component)}. \textcolor{clr}{To take into account the dependence of SPM results on the choice of a distinct alpha level, we performed experiments with three different alpha levels: 0.01, 0.05, and 0.1.} The output of SPM provides \textit{t}-values for each point of the investigated time series and the threshold corresponding to the chosen alpha level. \textcolor{clr}{The \textit{t}-values exceeding this threshold indicate statistically significant differences in the corresponding sections of the time series. For a better visibility, we marked these significant sections as gray-shaded areas in Figures~\ref{img:cnn-nonorm-NGD},~\ref{img:cnn-norm-NGD}, and~\ref{img:cnn-svm-mlp-norm-NGD}. We used three different shades of gray for the three different alpha levels, \ie~dark gray for 0.01, gray for 0.05, and light gray for 0.1.} Additionally, we computed the \textit{effect size} by transforming the resulting \textit{t}-values to Pearson's correlation coefficient~\textit{r} using the definition by Rosenthal~\cite{rosenthal_meta-analytic_1986}. \textcolor{clr}{The effect size provides an indicator for the discriminativeness of a given signal region independent of the alpha level.}


\subsection{\textcolor{clr}{Clinical Evaluation}}
\textcolor{clr}{To evaluate the derived relevance scores of LRP from a clinical perspective, a clinical expert with more than ten years' experience in human gait analysis analyzed the explainability results. The expert verified the extent to which regions with the highest input relevance scores correspond to GRF characteristics from clinical practice.}

\section{Experimental Setup}
\label{sec:experimentalSetuo}

\subsection{Data Recording and Dataset}
For the gait classification task we utilized a subset of the large-scale {\sc GaitRec} dataset~\cite{horsak_gaitrec_2020}. This dataset is part of an existing clinical gait database maintained by a local Austrian rehabilitation center. Before conducting our experiments approval was obtained from the local Ethics Committee (\#GS1-EK-4/299-2014). The employed dataset contains bilateral three-dimensional ground reaction force (GRF) recordings of patients and healthy controls walking unassisted at self-selected walking speed on an approximately 10~m walkway with two centrally-embedded force plates (Kistler, Type 9281B12, Winterthur, CH). Data were recorded at 2000~Hz, filtered with a zero-lag Butterworth filter of 2nd order with a cut-off frequency of 20~Hz, time-normalized to 101 points (100\% stance phase), and amplitude-normalized to 100\% body weight. During one session subjects walked barefoot or in socks until a minimum number of 5 valid recordings were available. Recordings were defined as valid by an experienced assessor.

\begin{table*}[ht!]
\centering
\caption{Demographic details of the employed dataset for each pre-defined class.}
\label{table:dataset}
\begin{tabular}{lcccccc}
\hline
Classes & N & \begin{tabular}[c]{@{}c@{}}Age (yrs.)\\ Mean (SD)\end{tabular} & \begin{tabular}[c]{@{}c@{}}Body Mass (kg)\\ Mean (SD)\end{tabular} & \begin{tabular}[c]{@{}c@{}}Gender\\ (m/f)\end{tabular} & \begin{tabular}[c]{@{}c@{}}Walking Speed\\ (m/s)\end{tabular} & \begin{tabular}[c]{@{}c@{}}Num.\\ Trials\end{tabular} \\
\hline
Healthy Control & 62 & 36.0 (10.8) & 72.3 (15.0) & 28/34 & 4.1 (0.3) &  310 \\
Hip		        & 37 & 44.2 (12.5) & 81.4 (14.1) & 31/6  & 3.7 (0.3) &  185 \\
Knee 			& 52 & 43.5 (13.8) & 85.6 (16.4) & 37/15 & 3.5 (0.4) &  260 \\
Ankle			& 43 & 42.6 (10.9) & 91.6 (20.4) & 36/7  & 3.4 (0.4) &  215 \\
\hline
\textbf{Total}		& \textbf{194} & \textbf{41.1 (12.4)} & \textbf{81.9 (18.0)} & \textbf{132/62} & \textbf{3.7 (0.5)} & \textbf{970} \\ \hline
\end{tabular}
\end{table*}

\begin{figure}[ht!]
  \centering
	\includegraphics[width=1\linewidth]{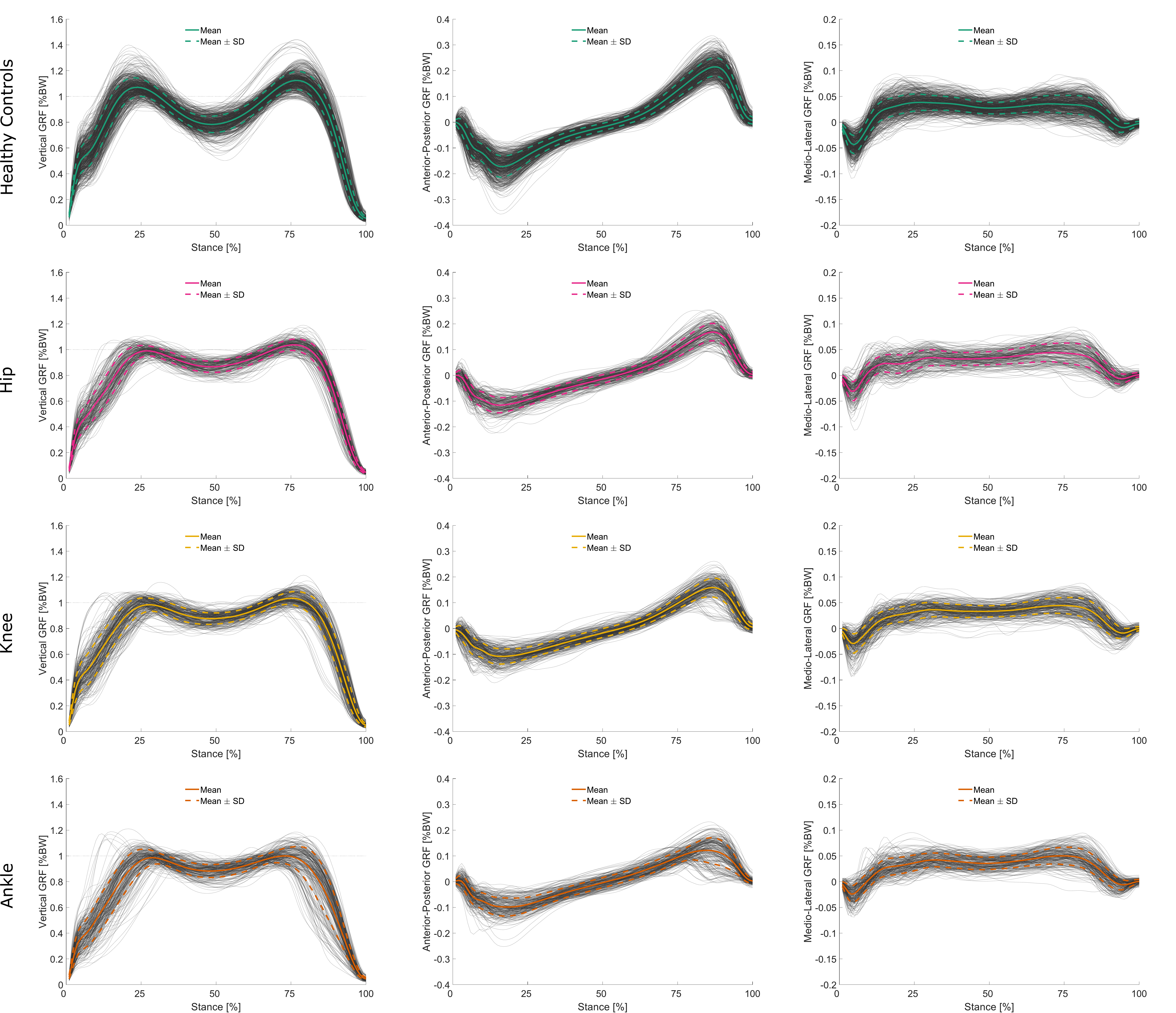}
    \caption{Visualization of vertical (left panel), anterior-posterior (central panel), and medio-lateral (right panel) force components of the body weight-normalized GRF measurements of the affected side available per subject and class. For healthy controls all available measurements are visualized. Mean and standard deviation signals (calculated per class) are highlighted as solid and dashed colored lines.}
    \label{img:waveforms}
\end{figure}

In total, the dataset comprises GRF measurements from 132 patients with lower-body gait disorders ($GD$) and data from 62 healthy controls ($HC$), both of various physical composition and gender. The dataset includes three classes of orthopaedic gait disorders associated with the hip ($H$, N=37), knee ($K$, N=52), and ankle ($A$, N=43). For class-specific demographic details of the data refer to Table~\ref{table:dataset}. The dataset is balanced regarding the number of recorded sessions per person and the number of trials per person. Figure~\ref{img:waveforms} shows an overview of all GRF measurements of the affected side (except for healthy controls where each step is visualized) per class and the associated mean and standard deviation. The $GD$ classes ($A$, $H$, and $K$) include patients after joint replacement surgery, fractures, ligament ruptures, and related disorders associated with the above-mentioned anatomical areas. A well-experienced physical therapist with more than a decade of clinical experience manually labeled the dataset based on the available medical diagnosis of each patient.

\subsection{\textcolor{clr}{Input Data Preparation}}

\textcolor{clr}{The input data for each classification task is a concatenated version of the three-dimensional GRF signals from both force plates (see Figure \ref{img:overview_approach}). The concatenation of all six GRF signals (three force components per force plate) results in a 1$\times$606-dimensional input vector for each gait trial}. The three-dimensional GRF signals are the medio-lateral horizontal force ($GRF_{ML}$), anterior-posterior horizontal force ($GRF_{AP}$), and vertical force ($GRF_{V}$). The dataset includes only unilateral gait disorders, \textcolor{clr}{\ie~disorders where the main physical limitation can be attributed to one leg (the \textit{affected leg/side} in the following). The signal components of the affected leg (input features: 1 to 303) are concatenated first and are followed by the signal components of the unaffected leg (input features: 304 to 606) in the input vector. For the healthy controls there is no affected and unaffected side (both sides are unaffected). Thus, the order of the signals was randomly assigned, while ensuring an equal distribution, to avoid any bias regarding the side. 
}

\subsection{\textcolor{clr}{Data Normalization}}
Normalization of input vectors is applied to ensure an equal contribution of all six GRF signals to the classification models and thus avoids that signals with larger numeric ranges dominate those with smaller numeric ranges~\citep{Hsu.2016,francois2017deep}.
To evaluate the robustness of our models' predictions and the derived relevance estimates concerning normalization, we firstly conducted experiments without normalization. In a second step, we conducted the same experiments applying min-max normalization to the input signals and thereby scaled each signal to the range $[-1,1]$. The global minimum and maximum values were determined separately for each of the six GRF signals over all trials.

\subsection{\textcolor{clr}{Classification Tasks}}

\textcolor{clr}{ 
We investigate six different classification tasks on the dataset introduced above to provide a more comprehensive picture on the investigated problem and to enable the differentiation between task-specific and general observations. Classification tasks include:}
\begin{itemize}
    \item binary classification between healthy controls and all gait disorders ($HC/GD$),
    \item binary classification between healthy controls and each gait disorder separately (\ie~$HC/H$, $HC/K$, and $HC/A$),
    \item multi-class classification between healthy controls and all gait disorders ($HC/H/K/A$),
    \item and multi-class classification between all gait disorders ($H/K/A$).
\end{itemize}

\subsection{\textcolor{clr}{Classification Methods}}


In our experiments, three representative machine learning approaches,
\ie~(linear) Support Vector Machine (SVM),
Multi-layer Perceptron (MLP),
and Convolutional Neural Network (CNN)
were compared in terms of prediction accuracy and learned input relevance patterns.
The SVM models were trained using a standard quadratic optimization algorithm, with an error penalty parameter $C=0.1$ and $\ell_2$-constrained regularization of the learned weight vector $w$.
The MLP models comprised of three consecutive fully connected layers with ReLU non-linearities activating the hidden neurons and a final SoftMax activation in the output layer. The size of both hidden layers is 768 whereas the size of the output layer is $c$, where $c$ is the number of target classes.
The CNN models process the given data via three consecutive convolutional layers, with a 
$<$filter size$>$-$<$stride$>$-$<$output channel$>$
configuration 
of 8-2-24, 8-2-24 and 6-3-48, and ReLUs for non-linear neuron activation.
The resulting 48$\times$48 feature mapping is then unrolled into a 2304-dimensional vector,
and fed into a fully-connected layer, which directly maps to the model output. This fully connected layer is topped with a SoftMax output activation, which is 
acting as a multi-class predictor output towards the $c$ target classes. Both, the MLP and CNN models, have been trained via standard error back-propagation using stochastic gradient descent~\citep{lecun2012efficient} and a mean absolute ($\ell_1$) loss function. The training procedure was executed for $3\cdot10^{4}$ iterations of mini batches of five randomly selected training samples and an initial learning rate of $5\cdot10^{-3}$. The learning rate was gradually decreased after every $10^{4}$-th training iteration to $10^{-3}$ by a factor of $0.2$ and then to $5\cdot10^{-4}$ by a factor of $0.5$. Model weights were initialized with random values drawn from a normal distribution with $\mu=0$ and $\sigma=m^{-\frac{1}{2}}$, where $m$ is the number of inputs to each output neuron of the layer~\citep{lecun2012efficient}.
Since the CNN receives as input a 1$\times$606-dimensional input vector,
its convolution operations can be understood as 1D convolutions, moving over the time axis only. We used 1D convolutions to maintain comparability with the two other classification methods (MLP and SVM). Preliminary experiments demonstrated negligible differences between 1D and 2D CNNs.

\subsection{\textcolor{clr}{Performance Evaluation}}
The prediction accuracies were reported over a stratified ten-fold cross-validation configuration, where eight partitions of the data are used for training, one partition is used as validation set and the remaining partition is reserved for testing. The samples from each class were distributed evenly while ensuring that all gait trials from an individual subject are placed in the same partition of the data to rule out subject-related information influencing the measured model performance during testing. All results are reported as mean with standard deviation (SD), unless otherwise stated. Additionally, we calculated the Zero Rule baseline (ZRB) for each classification task. The ZRB refers to the theoretical accuracy obtained by assigning class labels according to the prior probabilities of the classes, \ie~the target labels are always set to the class with the greatest cardinality in the training dataset.

\subsection{\textcolor{clr}{Implementation}}
The implementation of the three ML methods and the LRP method was conducted within the software framework Python~3.7 (Python Software Foundation, USA). Data preprocessing, SPM, and the visualization of the results were performed in Matlab 2017b (MathWorks, USA).

\section{Results}
\label{sec:results}

\textcolor{clr}{We first present the results obtained in our classification experiments as well as from the explainability analysis and then discuss them in detail in Section \ref{sec:Discussion}. We start with a presentation of the classification accuracies achieved for the different classification methods, tasks, and normalization methods (Section \ref{subsec:resultClassification}) and continue with a presentation of the explainability results obtained by LRP (Section \ref{subsec:resultLRP}).} 

\subsection{\textcolor{clr}{Classification Results}}
\label{subsec:resultClassification}
\begin{figure}[ht!]
  \centering
	\includegraphics[width=0.6\linewidth]{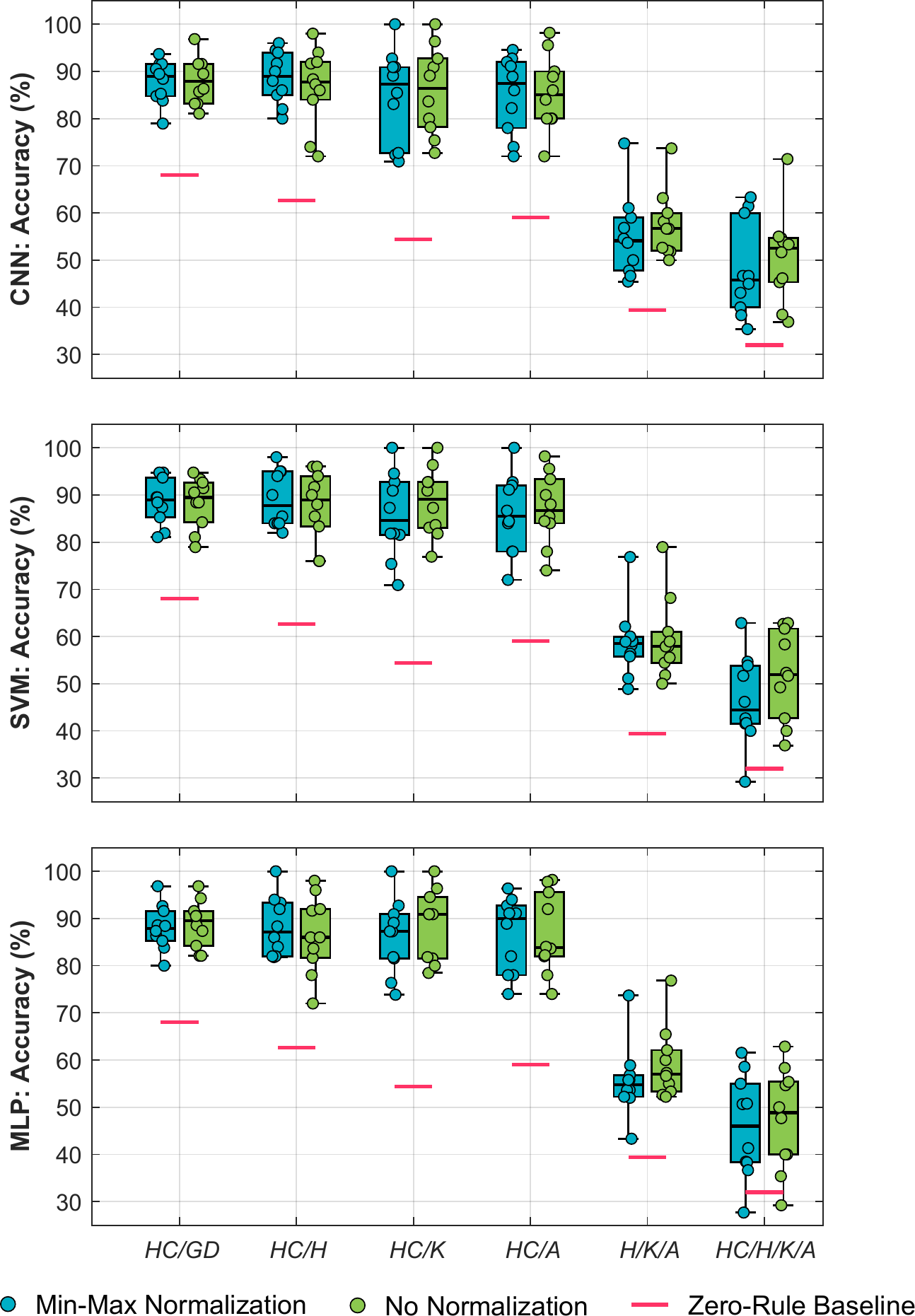} 
    \caption{Overview of the prediction accuracy obtained for the three employed classification methods (CNN, SVM and MLP) and all six classification tasks with min-max normalized and non-normalized input signals, reported as boxplots enhanced with the classification accuracies obtained over ten-fold cross validation (represented as individual dots).}
    \label{img:classacc}
\end{figure}

The mean prediction accuracy showed a clear superiority over the ZRB for all three classification methods (CNN, SVM, and MLP) and all classification tasks (see Figure~\ref{img:classacc} and Supplementary Table~S1). A 2$\times$2 repeated measures analysis of variance (ANOVA) (classification method: CNN, SVM, and MLP; normalization: min-max and non-normalized) conducted for each classification task only indicated a significant difference in classification accuracy between the three classifiers for task $HC/H/K/A$ ($F_{2,18}$ = 5.251, p = 0.016, $\eta_{p}^{2}$ = 0.368). Additional pairwise and Bonferroni-corrected post-hoc tests revealed that the CNN resulted in a marginally ($\sim$3\%), but significantly (p = 0.029) lower accuracy than the SVM for task $HC/H/K/A$. No other significant differences were found for the classifiers’ performances. Regarding normalization, no significant differences were found. Only for task $HC/H/K/A$~($F_{1,9}$ = 4.670, p = 0.059, $\eta_{p}^{2} $ = 0.342) the ANOVA slightly missed the alpha level. For the other tasks no significant effects and differences were found.

\subsection{\textcolor{clr}{Explainability Results}}
\label{subsec:resultLRP} 

\textcolor{clr}{
In the following, we present in detail the results for classification task $HC/GD$ together with respective result visualizations. Results for the other five investigated classification tasks can be found in the supplementary material~(see Supplementary Figures~S1--S24).}

Figure~\ref{img:cnn-nonorm-NGD} shows an exemplary result for prediction explanation by LRP, \ie the averaged signals together with the color-coded averaged relevance values for each of the 606 input values for task $HC/GD$ with non-normalized GRF signals. The input relevance values point out which GRF characteristics were most relevant for (or contradictory to) the classification of a certain class ($HC$ or $GD$). For visualization, input values neutral to the prediction ($R_i \approx 0$) are shown in black color, while warm hues indicate input values supporting the prediction ($R_i \gg 0 $) of the analyzed class and cool hues identify contradictory input values ($R_i \ll 0 $). For binary classification tasks ($HC/GD$, $HC/H$, $HC/K$, and $HC/A$), note that a high input relevance value for one class results in a contradictory input relevance value for the other class. Therefore, the total relevance, \textcolor{clr}{which is the absolute sum of the relevance scores of both classes} is a good indicator for the overall relevance of an input value for a respective classification task. \textcolor{clr}{The higher the total relevance at a certain signal region, the more discriminative is this region for the two underlying classes.}

\begin{figure}[ht!]
  \centering
	\includegraphics[width=1\linewidth]{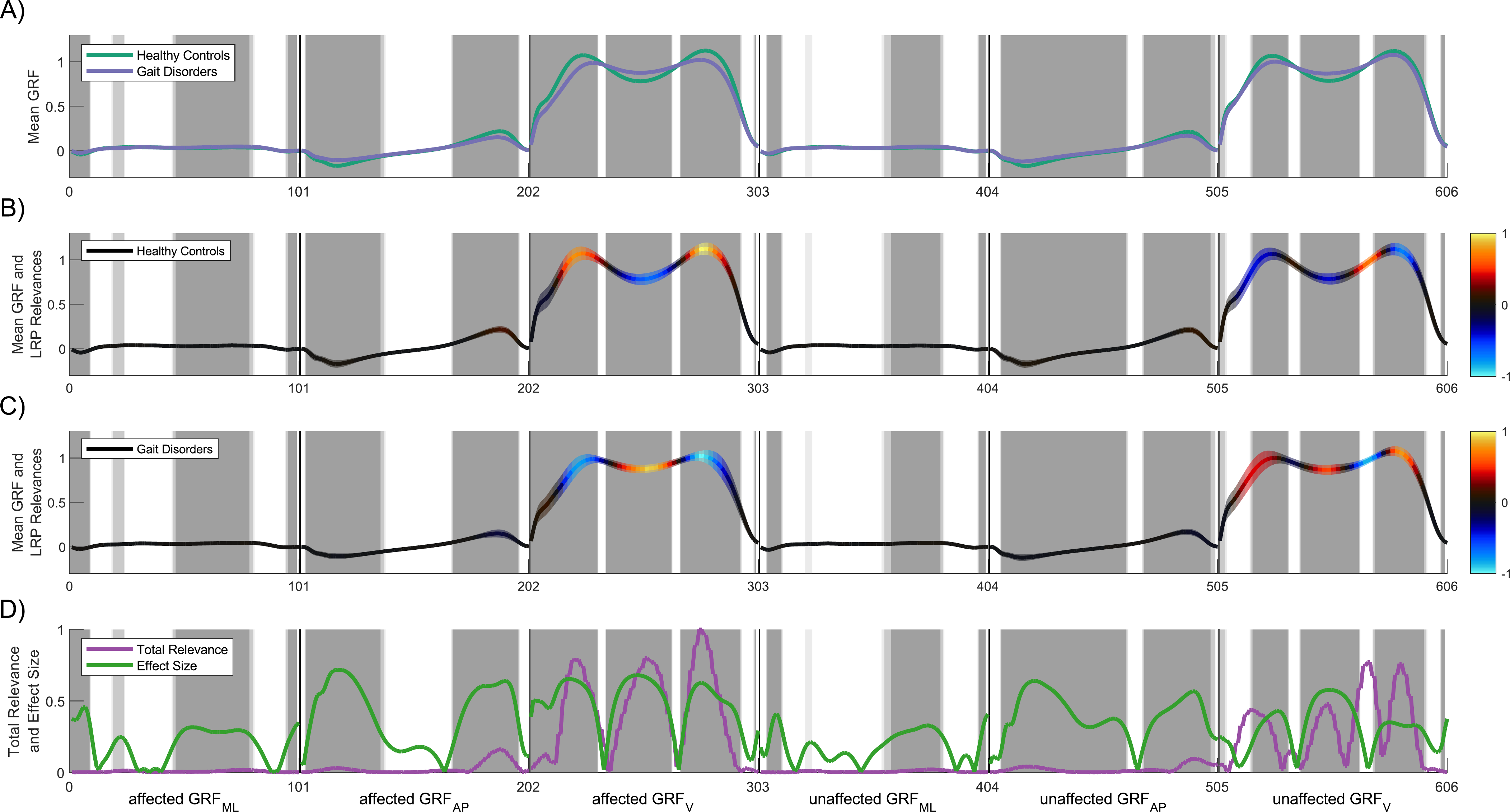} 
    \caption{Results overview for the classification of healthy controls~($HC$)  and the aggregated class of all three gait disorders~($GD$) based on non-normalized GRF signals using a CNN as classifier. (A) Averaged GRF signals for $HC$ and $GD$. The first three signals represent the three GRF components of the affected side and are followed by the three GRF components of the unaffected side. Note that the data for both sides are composed of three GRF components (\eg~input features of the affected side: 1 to 101 ($GRF_{ML}$), 102 to 202 ($GRF_{AP}$), and 203 to 303 ($GRF_{V}$)). This means, for example, that input features 21 ($GRF_{ML}$), 122 ($GRF_{AP}$) and 233 ($GRF_{V}$) all correspond to the relative time of 20\% of the same stance phase. \textcolor{clr}{The areas, which are colored in three different shades of grey for the three different alpha levels, \ie~dark grey for 0.01, grey for 0.05, and light grey for 0.1, highlight regions in the input signals where SPM indicates statistically significant differences between both classes~(\ie~$HC$ and $GD$)}. (B) Averaged GRF signals of all test trials as a line plot for the healthy controls class, with a band of one standard deviation, color coded via input relevance values for the class ($HC$) obtained by LRP. (C) Averaged GRF signals of all test trials are shown as a line plot for the class of all the gait disorders ($GD$), in the same format as in (B). (D) Line plot showing the effect size obtained from SPM and total relevance based on the absolute sum of the LRP relevance values of both classes ($HC$ and $GD$). \textcolor{clr}{The total relevance correlates with the local discriminativity of the input signal for the classification task.}}
    \label{img:cnn-nonorm-NGD}
\end{figure}

The highest input relevance values were observed in $GRF_{V}$ of the affected side, as illustrated in Figure~\ref{img:cnn-nonorm-NGD}. In the same way, the mean GRF signals of both classes~(see Figure~\ref{img:cnn-nonorm-NGD}A) as well as the SPM analysis (gray-shaded areas in Figure~\ref{img:cnn-nonorm-NGD}) highlighted statistically significant differences between both classes in the same regions of $GRF_{V}$ of the affected side. \textcolor{clr}{While LRP identified hardly any relevant regions in both horizontal forces ($GRF_{AP}$ and $GRF_{ML}$), it is noticeable that the SPM analysis highlighted a lot more regions within $GRF_{AP}$ and $GRF_{ML}$ as statistically significantly different between the two classes.} This shows that relevant regions identified by LRP mostly correlated with significant signal regions. However, not all significant regions are necessarily used by the classification model.

\begin{figure}[ht!]
  \centering
	\includegraphics[width=1\linewidth]{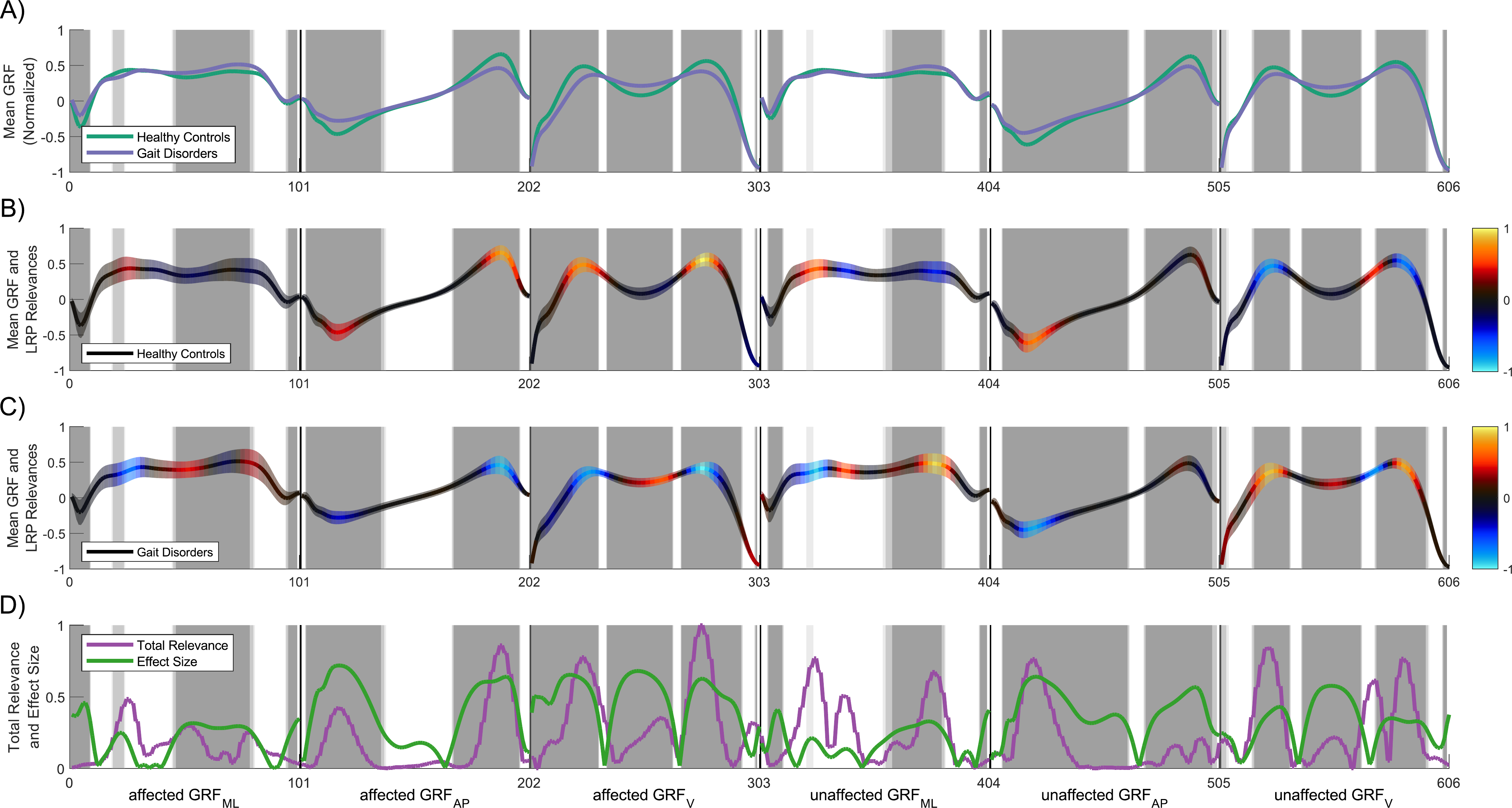} 
    \caption{Same experiment as shown in Figure~\ref{img:cnn-nonorm-NGD} but \textcolor{clr}{with min-max normalized GRF signals. SPM regions are not affected by the normalization. High LRP relevance scores are also present in the horizontal forces $GRF_{AP}$ and $GRF_{ML}$.}}
    \label{img:cnn-norm-NGD}
\end{figure}

\textcolor{clr}{When min-max normalization is applied to the input data of the classification task from above ($HC/GD$), the identified regions of high relevance in $GRF_{V}$ are similar to those obtained from non-normalized signals (see Figure~\ref{img:cnn-nonorm-NGD}). However, there are many additional regions of high relevance in both horizontal forces (see Figure~\ref{img:cnn-norm-NGD}). 
Note that the regions identified by SPM are the same since they are not affected by the normalization. The additionally identified regions of high relevance according to LRP agree to a large extent with the SPM results. The regions with the highest input relevance values for the prediction can be observed at approximately 20\% and 80\% of the stance phase in the affected and unaffected GRF signals. 
For the $GRF_{ML}$ a relevant region can also be observed at 10\% of the stance phase. In addition, high input relevance scores can be observed in the $GRF_{V}$ during the first and last $\sim$5\% of the stance phase of the affected and unaffected side, as well as during the middle of the stance phase of the affected side.}

\begin{figure}[ht!]
  \centering
	\includegraphics[width=1\linewidth]{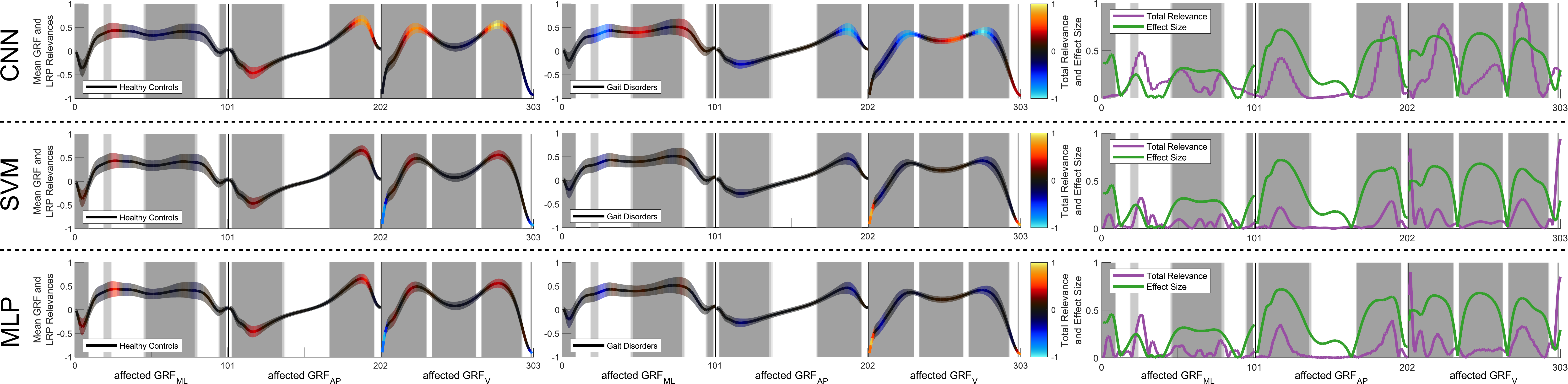} 
    \caption{Comparison of different methods (CNN, SVM, and MLP) for the classification of healthy controls and the class of all three gait disorders ($HC/GD$) based on min-max normalized GRF signals (only the signals of the affected side are shown). The comparison is based on the decomposed input relevance values for both classes ($HC$ and $GD$), their total relevance determined by LRP as well as statistically significant differences (gray-shaded areas) and effect sizes obtained by SPM. Note that the effect size (green curve) is the same for all three classifiers but the total relevance varies.}
    \label{img:cnn-svm-mlp-norm-NGD}
\end{figure}

\textcolor{clr}{
Figure~\ref{img:cnn-svm-mlp-norm-NGD} shows the results of task $HC/GD$ (with min-max normalized GRF signals as in Figure~\ref{img:cnn-norm-NGD}) for all three employed classification methods~(CNN, SVM, and MLP). The relevance scores agree to a large extent. However, with respect to $GRF_{V}$, the highest input relevance values can be observed in the peak regions for the CNN, while the highest input relevance values for SVM and MLP are present during the first and last $\sim$5\% of the stance phase~(beginning and end of the $GRF_{V}$ signal). These results show that the investigated classification methods rely on the same regions in the input data with only small exceptions.}

\textcolor{clr}{
For the sake of brevity, only the results for the classification task $HC/GD$ were presented. For results of the other classification tasks we refer the reader to the supplementary material~(see Supplementary Figures~S1--S24). The discussion in the following will incorporate all six classification tasks.}

\section{Discussion}
\label{sec:Discussion}

The primary aim of this article is to investigate whether XAI methods can enhance explainability of ML predictions in clinical gait classification. \textcolor{clr}{In this section, the classification results are analyzed, compared and interpreted in terms of classification accuracy and relevance-based explanations. These explanations are, furthermore, evaluated from a statistical and clinical viewpoint. Additionally, we discuss dependencies, influences and interesting observations with respect to different classification methods, tasks, normalization methods, and signal components (horizontal forces and affected/unaffected leg signals).}

\subsection{Classification Results}
\label{subsec:classificationResults}
The results expressed in terms of classification accuracy (presented in  Figure~\ref{img:classacc} and Supplementary Table~S1) demonstrate a comparable level of performance between the three different machine learning methods (CNN, SVM, and MLP). \textcolor{clr}{The achieved performance level is not only interesting by itself but also important information for further explainability experiments. The reason is that an objective analysis of explainability by a post-hoc method like LRP is only meaningful if the classification model can robustly differentiate between the target classes,~\ie~a certain model quality is necessary to draw meaningful conclusions from explainability results.} An analysis of unreliable classification models bears the potential risk that unstable patterns, noise, and spurious correlations bias the explainability results. For this reason, we excluded the classification tasks $HC/H/K/A$ and $H/K/A$ from our further investigation, \textcolor{clr}{as the tasks could not be solved with sufficient accuracy (average classification accuracy above 80\%).} For the binary classification tasks this risk is much lower, because the higher classification accuracies (and deviations from ZRB) obtained suggest that robust features can be found in the input data.

Another aspect we assessed is the influence of normalization on the input data (see Figure~\ref{img:classacc}). The normalization of the input data is important for machine learning since highly differing value ranges can have a negative influence on the classification model,~\ie~input variables with a higher value range have a stronger influence on the predictions~\citep{Hsu.2016,francois2017deep}. \textcolor{clr}{The same appears to be the case for the gait data, where the normalization of the input data strongly influences the classification models, as can be observed from the relevance scores of the horizontal forces (compare Figures \ref{img:cnn-nonorm-NGD} and \ref{img:cnn-norm-NGD}). 
Surprisingly, however, min-max normalization does not significantly improve the classification results (see Figure~\ref{img:classacc}) for the investigated classification tasks}. This raises the question of whether the use of $GRF_{V}$ alone would already be sufficient to solve the classification tasks. We discuss this seemingly contradictory behavior in the following section.

\subsection{Explainability Results}
\label{subsec:explResults}

\textcolor{clr}{In the following, we discuss different related aspects with regard to our first leading research question: ``\textit{\textbf{Which input features or signal regions are most relevant for automatic gait classification?}}''.} The visualizations for all classification tasks and classification methods can be found in the supplementary material~(see Supplementary Figures~S1--S24).

\textbf{Which input features and signal regions are most relevant for the automatic classification of functional gait disorders?} For the classification of non-normalized GRF signals (\eg~Figure~\ref{img:cnn-nonorm-NGD}), the most relevant input values are mainly located in $GRF_{V}$ of the affected side, \ie~especially the two peaks and the valley in between are relevant for the tasks. This shows that the classification models learned that the $HC$ class and the gait disorder classes ($GD$, $H$, $K$, and $A$) differ most in these three sections of the signals. 
These results were also confirmed in our earlier studies, \eg~where the peaks of $GRF_{V}$ and the valley between them were most important for discriminating the classes~\citep{slijepcevic2017automatic}.

\textbf{Is the unaffected side important?} Identified relevant regions are considerably less pronounced in $GRF_{V}$ of the unaffected side, but they correlate to a large extent with those of the affected side, except that only the rear part of the valley and not the entire valley is relevant (best recognized from the total relevance curve in \eg~Figure~\ref{img:cnn-nonorm-NGD}D). In earlier studies~\citep{slijepcevic2018p, SLIJEPCEVIC2019}, we showed that the omission of the unaffected side during classification negatively affected classification accuracy. The explainability results confirm this observation. The unaffected side seems to capture complementary information relevant to the classification task. 

\textbf{Are the anterior-posterior and medio-lateral forces relevant for the task?}
A minimal degree of relevance can be observed in the peaks of the non-normalized affected and unaffected $GRF_{AP}$ signals. The absence of relevant regions from the horizontal forces ($GRF_{AP}$ and $GRF_{ML}$) does not confirm our results from previous studies using normalized GRF signals~\citep{slijepcevic2018p,SLIJEPCEVIC2019}, where we showed that adding horizontal forces indeed improved classification performance (leading even to peak performance). The question of whether or not horizontal forces are beneficial for the task cannot be answered conclusively. Interestingly, the statistical analysis via SPM highlights regions in the horizontal forces that differ significantly between the $HC$ class and the gait disorder classes ($GD$, $H$, $K$, and $A$).

\textbf{What is the impact of normalization on explainability?}
The reason for the absence of relevant regions in the horizontal forces could be their small value range. The rather small range compared to the $GRF_{V}$ component may lead to a smaller influence on the training of the classification models. Explainability results for min-max normalized input data show that more relevant regions are identified by LRP in the horizontal forces of the affected and unaffected side (\eg~Figure~\ref{img:cnn-norm-NGD}). Normalization amplifies the relevance of values in the horizontal forces and thereby makes them similarly important as $GRF_{V}$. Based on the LRP relevance scores, we conclude that normalization is important to obtain unbiased predictions (bias introduced by different signal amplitudes).

\textbf{Are all identified relevant regions \textit{necessary} for the task?} For all classification tasks and classification methods, with min-max normalized input data, many regions of the GRF signals are identified to be relevant for the classification of a particular class by LRP. The classification performance with and without normalization does, however, not vary significantly~(see classification results in Section~\ref{subsec:resultClassification}). This raises the question of whether all identified regions are necessary to achieve peak performance in classification or whether some of them are redundant~(\ie~not yielding an increase in classification performance when combined). Note that the assumption of redundancy is supported by the fact that the three GRF components represent individual dimensions of the same three-dimensional physical process. Thus, a strong correlation is a priori given in the data.
To answer the question we occluded parts of the input vector in the classification experiment and evaluated the changes in classification performance. Occlusion is realized by replacing the horizontal forces~($GRF_{AP}$ and $GRF_{ML}$) of both sides~(affected and unaffected) with zero values and retraining the classification model. Table~\ref{table:horizontalzero-classification-results} shows the classification results for the occluded input. To enable easier comparison with the previous results, the deviation from the mean classification accuracy of the non-occluded experiments~(from Figure~\ref{img:classacc} and Supplementary Table~S1) are displayed for all binary classification tasks (see Table~\ref{table:horizontalzero-classification-results}). 
The results decrease on average when the horizontal forces are occluded, except for task $HC/A$ with min-max normalized input data. Furthermore, the decrease is more pronounced for min-max normalized input data than for non-normalized input data. This further corroborates our assumption that normalization is important to take information from horizontal forces into account.
\textcolor{clr}{However, the classification results of the binary classification tasks are not influenced by the occlusion of horizontal forces in a statistically significant way.} This was confirmed by several dependent t-tests~(p $>$ 0.05) with Bonferroni correction. Our results indicate that the relevant regions identified by LRP may represent an over-complete set, which exhibits a certain degree of redundancy, as removing one section does not necessarily lead to reduced classification performance. \textcolor{clr}{However, redundancy is not necessarily a negative property. It may be an important property to achieve higher robustness to noise and possibly also to outliers and missing data~\citep{horst2019explaining}.}

\textbf{Are the anterior-posterior and medio-lateral forces relevant for the task (question revisited)?} The effects of occluded horizontal forces are illustrated in Table~\ref{table:horizontalzero-classification-results}.
\textcolor{clr}{Especially for experiments with min-max normalized input data a decrease in mean performance can be observed~(\eg~$HC/H$ and $HC/K$).} Thus, the relevant regions in the horizontal forces cannot be completely redundant to those in $GRF_{V}$ and, therefore, represent also complementary information. This is also in line with our previous quantitative performance evaluations~\citep{slijepcevic2018p,SLIJEPCEVIC2019}.

\begin{table*}[ht!]
\centering
\caption{Classification results for the experiment with occluded horizontal forces ($GRF_{AP}$, $GRF_{ML}$), in percent. The results are reported as mean deviation from the prediction accuracy of the original input signals presented in Figure~\ref{img:classacc}, \ie~negative values signify a decrease and positive values an improvement in classification performance.}
\label{table:horizontalzero-classification-results}
\begin{tabular}{lcccc}
\hline
Task        & Normalization         & SVM & MLP & CNN \\
\hline
HC/GD        & no norm.            & -0.9 & -1.0 & -0.5 \\
HC/GD 	    & min-max             & -1.1 & -2.0 & -0.3 \\
\hline
HC/H         & no norm.            & -1.8 & -2.1 & -0.8 \\
HC/H 	    & min-max             & -2.2 & -3.7 & -0.3 \\
\hline
HC/K         & no norm.            & -1.8 & -1.9 & -2.1 \\
HC/K 	    & min-max             & -4.1 & -5.0 & -2.9 \\
\hline
HC/A         & no norm.            & -1.6 & -1.8 & -2.1 \\
HC/A 	    & min-max             & 1.2 & 0.9 & 0.5 \\
\hline
\end{tabular}
\end{table*}

\textbf{Do different classifiers rely on different patterns?} A condensed comparison of the three employed classification methods is depicted in Figure~\ref{img:cnn-svm-mlp-norm-NGD}. The LRP relevance values are consistent for non-normalized and normalized input data. For the former (\eg~for task $HC/GD$ see Supplementary Figures~S1, S3, and S5), the relevant regions for SVM and MLP largely correspond across all binary classification tasks. The CNN matches the relevant regions of SVM and MLP in broad terms. 
The relevant regions in $GRF_V$ of the unaffected side for the CNN are considerably lower compared to SVM and MLP~(compare Supplementary Figures~S1, S3, and S5), \eg~for task $HC/GD$ the valley in $GRF_V$ of the unaffected side is hardly relevant and the second peak in $GRF_V$ of the unaffected side is considerably less relevant compared to SVM and MLP. 
For min-max normalized data~(see Figure~\ref{img:cnn-svm-mlp-norm-NGD}), the relevant regions for SVM and MLP coincide also to a large extent. The relevant regions of CNN correspond to those of SVM and MLP with regard to their location, but are considerably more relevant (best visible in the total relevance curves in the right part of Figure~\ref{img:cnn-svm-mlp-norm-NGD}).
The most pronounced difference between the classification methods can be observed in the estimated relevance scores at the beginning and end of $GRF_{V}$ signals. While LRP indicates that those regions are relevant for SVM and MLP, the total relevance curve of the CNN does not show any correspondence in those regions. 
The remaining binary classification tasks,~\ie~$HC/H$ (see Supplementary Figures~S7--S12), $HC/K$ (see Supplementary Figures~S13--S18) and $HC/A$ (see Supplementary Figures~S19--S24) confirm these findings. \textcolor{clr}{A reason for the strong difference in relevance scores at the beginning and end of the stance phase might be the higher degree of inter- and intra-subject variability due to balance instabilities. Additionally, measurement noise may bias the rather small force values during these phases of stance~\citep{BIZOVSKA2014399}.}


In the absence of ground truth information for automatically generated explanations, it is difficult to assess whether regions \textcolor{clr}{considered as relevant for SVM and MLP are related to (biomechanically) meaningful gait characteristics. While LRP clearly shows where the prediction is grounded, it cannot explain \textit{why} these patterns are important. However, it allows to identify and compare the learning strategies of different classification methods and, thus, points out potential degeneration effects, influences of noise, and spurious correlations.}

\subsection{\textcolor{clr}{Statistical Verification of Relevance-based Explanations}}
\label{subsec:statisticalResults}

\textcolor{clr}{In the following, we investigate the statistical properties of the signal regions found to be relevant by LRP to answer the second leading research question: \textit{\textbf{´´To what extent are input features or signal regions identified as being relevant for a given gait classification task statistically justified?''}.}} 
%
\textcolor{clr}{
To answer this question, we leverage SPM, which provides significance estimates for each sample of the input signals. We compare the LRP regions with those considered as significantly different by SPM.} 
Results show that in the vast majority of cases, the SPM analysis shows statistically significant differences in regions which are also highly relevant for classification according to LRP. Thus, for binary classification tasks, it seems that machine learning models base their predictions primarily on features that are also significantly different between the two classes. This can be observed,~\eg~in the $HC/GD$ classification for both, min-max normalized and non-normalized GRF signals in Figure~\ref{img:cnn-nonorm-NGD}D and Figure~\ref{img:cnn-norm-NGD}D. As the total relevance increases, the effect size usually also increases. \textcolor{clr}{We performed a cross-correlation to determine the relationship between the effect size and the total relevance. Both curves show highly correlated behavior for the normalized case (r $=$ 0.73), whereas for the non-normalized case a moderate positive correlation (r $=$ 0.61) could be observed. For the latter, the disagreement between SPM estimates and LRP results indicated the presence of bias in the analysis. This bias, which was introduced by different amplitude ranges in the input signals, prevented the classification models to learn relevant patterns in the horizontal forces. SPM clearly showed this bias, which underlines the role of SPM as a suitable statistical reference for our investigation.}

\textcolor{clr}{
Concerning our second research question, we conclude that the relevance estimates are to the greatest extent statistically justified, when the data is correctly preprocessed (normalized). The second part of the research question regarding the validity of the explanations with respect to clinical assessment is investigated in the following section.}


\subsection{Clinical Evaluation of Relevance-based Explanations}
\label{subsec:clinicalEval}

\begin{figure}[ht!]
  \centering
	\includegraphics[width=1\linewidth]{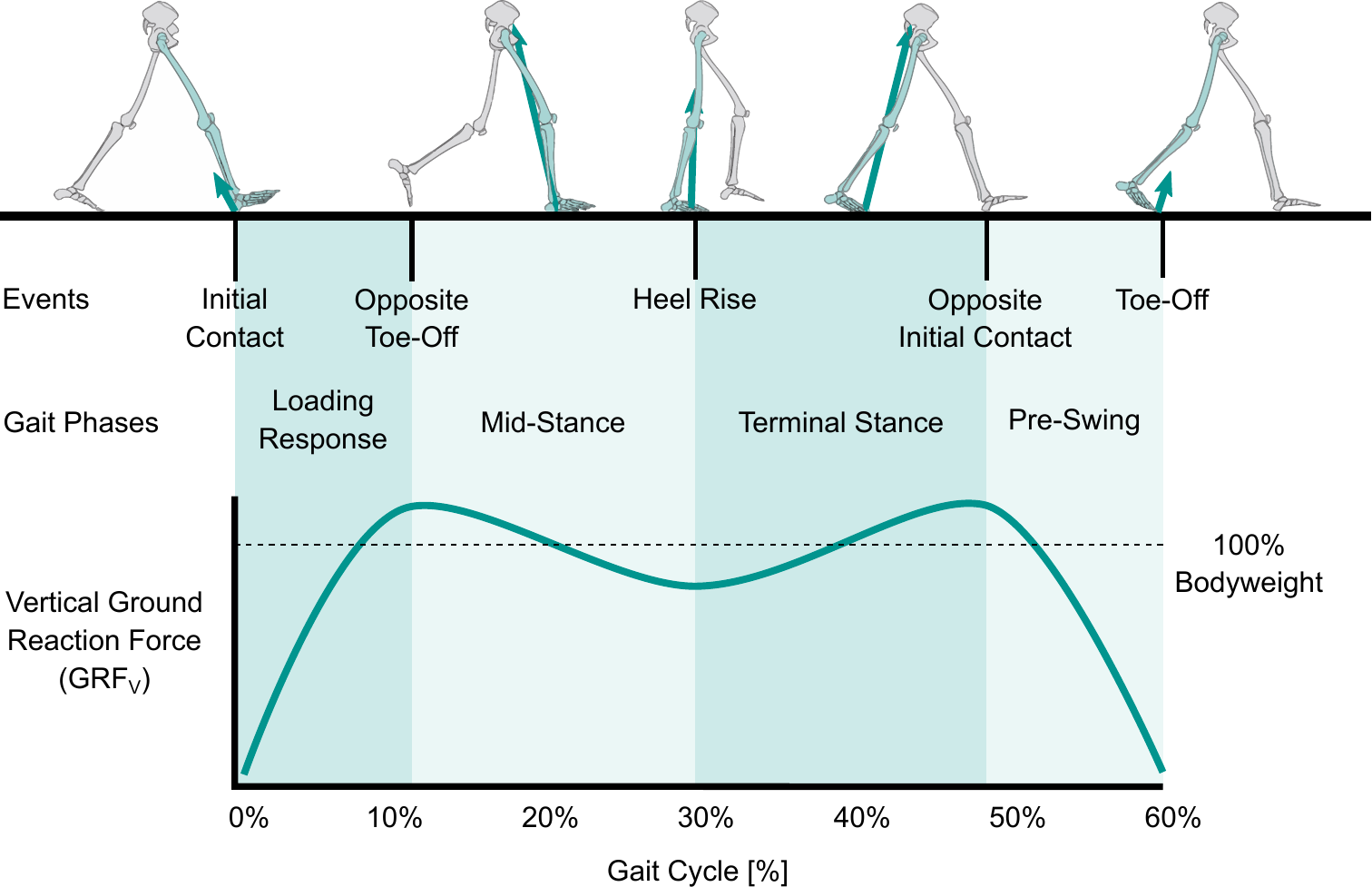} 
    \caption{
    Overview of the most relevant gait events during the stance phase. In clinical gait analysis, a gait cycle (100\%) is defined from the initial contact of one leg to the subsequent initial contact of the same leg. During the first 60\% of the gait cycle, also called stance phase (relevant time range for the present work) the foot has contact to the ground. The beginning of the stance phase is defined as initial contact with the ground (typically by the heel), then body weight is shifted to the supporting leg (loading response and mid-stance), followed by terminal stance (forward propulsion), pre-swing (preparation of the swing phase), and toe-off. Adapted from~\cite{shi2018effect,bakerMeasuringWalkingHandbook2013}.}
    \label{img:gait_phases}
\end{figure}

\textbf{\textit{\textbf{``To what extent are input features or signal regions identified as being relevant for a given gait classification task in line with clinical assessment?''}}} \textcolor{clr}{This question is answered in the following by a clinical expert with more than ten years' experience in human gait analysis. To assist the reader in following the discussion and to facilitate the interpretation of the input signals, 
the domain-specific terms and gait cycle definitions are described in Figure~\ref{img:gait_phases}
.}

The explainability results for min-max normalized input data illustrate clinically meaningful patterns. Across all classification tasks and models high LRP relevance scores \textcolor{clr}{occurred mainly during loading response and terminal stance phase in $GRF_{AP}$ and in loading response, mid stance, and terminal stance in $GRF_{V}$. These phases are especially sensitive toward gait anomalies as loading response requires the absorption of body weight and terminal stance plays an essential role for forward propulsion. Both aspects are affected in case of gait impairments due to a diminished walking speed (requiring less absorption or push-off) as well as factors that go along with an injury, such as the presence of pain, a decreased range of motion and/or lessened muscle strength.}

When analyzing the explainability results in more detail, one can identify specific gait dynamics that can be traced back to an impairment at a certain joint level. For classification task $HC/A$ (see Supplementary Figure~S20) we can observe pronounced peaks in the total relevance curves of $GRF_{AP}$, which may be caused by alterations in the pre-swing phase of the affected side and the consecutive breaking impulse during initial contact on the contralateral (unaffected) side.
For classification task $HC/K$, the highest LRP relevance scores are present in $GRF_{V}$, $GRF_{AP}$, and $GRF_{ML}$ (see Supplementary Figure~S14). Changes in $GRF_{V}$ may result from lessened knee flexibility that hinders typical knee dynamics over the entire course of the stance phase. More precisely, healthy walking requires an almost fully extended knee joint during initial contact followed by a slight knee flexion thereafter, by definition called loading response. During the mid stance phase the walker's center of gravity is shifted forward and thus demands further knee extension. Highest LRP relevance values for the classification task $HC/H$ are obtained during loading response and terminal stance in $GRF_{V}$ of the affected side (see Supplementary Figure~S8). These results may be ascribed to lowered impact and weight-bearing in the early stance phase (due to a potentially more cautious walking strategy) to avoid excessive load on the affected hip joint. \textcolor{clr}{These observations further corroborate the idea that LRP explanations agree well with clinical assessment (given the input data is normalized).}

As the results presented above are based on min-max normalized data, the question arises whether similar observations can be derived from the experiments without normalization. The prediction explanations for non-normalized data (see Figure~\ref{img:cnn-nonorm-NGD} and Supplementary Figures~S7,~S13,~S19) clearly show a different picture, even if classification results are comparable to those obtained with min-max normalized data. Relevant regions can only be found in $GRF_{V}$. \textcolor{clr}{These observations highlight that normalization strongly affects the quality of classification models and, therefore, needs to be considered to derive clinically meaningful explanations.}

\subsection{Limitations and Future Work}
A fundamental problem in evaluating the explainability results is the absence of a ground truth. A challenge in interpreting the explainability results is that alterations of the input signals can be caused not only by the influence of a pathology, but also by other independent parameters, \eg~a lower walking speed or an increased body mass. 
To minimize potential biases introduced by independent parameters on prediction explanations, future research should attempt to develop normalization procedures for input signals that compensate such influencing factors or develop classification models that inherently learn the relationship between influencing factors and input signals.

A limitation regarding the applicability of the present work in a clinical setting is its sole use of GRF signals. Kinematic data would allow a more comprehensive view on the biomechanical processes during walking and therefore simplify the interpretation of explainability results.

\textcolor{clr}{
Besides visual explanations as presented in this paper, a translation into human understandable textual explanations would be desired for clinical application. An interesting direction for future research is the generation of textual explanations based on biomechanical parameters estimated from the input signals. This would enable approaches that exceed pure explainability and provide deeper interpretations for clinical experts in the form of,~\eg~"there is a high probability of a pathology in the knee due to a limited knee extension during the mid stance phase".}

\section{Conclusion}
The present findings highlight that machine learning models base their predictions on meaningful features of GRF signals in clinical gait classification tasks that are in accordance with a statistical and clinical evaluation. Hence, XAI methods which provide explainability for predictions made by machine learning models, such as LRP, can be promising solutions to increase justification of automatic classification predictions in CGA and can help to make the prediction processes comprehensible to clinical and legal experts. Thereby, XAI may facilitate the application of ML-based decision-support systems in clinical practice. Within the scope of our analysis we were able to show that:

\begin{itemize}

    \item Highly relevant regions were identified in the signals of the affected and unaffected side. Thus, the unaffected side captures additional information which are relevant for automated gait classifications. 

    
    \item{ \textcolor{clr}{For time-series data such as GRF signals, SPM has shown to be a suitable statistical reference. Relevant regions in the input data (according to LRP) are in the most cases also significantly different and in line with clinical evaluation.}}
    
    \item Input data normalization allows machine learning models to consider features from various input signals for their predictions (especially if the value ranges differ as much as for the three force components of the GRF). Without normalization, only relevant regions in the $GRF_V$ were identified which is not reasonable from a clinical point of view. Therefore, data normalization is highly recommended for ML-based gait classification.
    
    \item For the investigated binary gait classification tasks, machine learning models seem to learn an over-complete set of features that may contain redundant information. This might explain why the occlusion of horizontal forces in our experiments had negligible influence on the classification accuracies and also why the classification accuracies for the classification of normalized and non-normalized GRF signals were similar.

\end{itemize}

This paper can be considered as a first step towards explainability of ML approaches in clinical gait analysis. We will conduct further research to compare different explanation methods and rule-based approaches~\citep{kohlbrenner2019towards} for different classification tasks and datasets. In addition, we want to point out that quantitative and objective methods are necessary to assess the quality of prediction explanations~\citep{samek_2017_evaluating} including datasets with respective ground truth explanations. 

\section*{Conflict of Interest Statement}

The authors declare that the research was conducted in the absence of any commercial or financial relationships that could be construed as a potential conflict of interest.

\section*{Author Contributions}

DS, A-MR, MZ, BH prepared the dataset.
FH, DS, SL, WS, WIS, BH conceived the presented idea.
MZ, WS, WIS, BH raised the funding.
FH, DS, SL, A-MR, MZ, WS, CB, WIS, BH participated in the data analysis.
FH, DS, SL, A-MR, MZ, BH wrote the manuscript.
FH, DS, SL, A-MR, MZ, WS, BH designed the figures.
FH, DS, SL, A-MR, MZ, WS, CB, WIS, BH reviewed and approved the final manuscript.

\section*{Funding}
This work was partly funded by the Austrian Research Promotion Agency (FFG) and the BMDW within the COIN-program (\#866855 and \#866880),
the Lower Austrian Research and Education Company (NFB), the Provincial Government of Lower Austria (\#FTI17-014). 
Further support was received from the German Ministry for Education and Research as BIFOLD (\#01IS18025A and \#01IS18037A) and TraMeExCo (\#01IS18056A).

\section*{Acknowledgments}
We want to thank Marianne Worisch, Szava Zolt\'{a}n, and Theresa Fischer for their great assistance in data preparation and their support in clinical and technical questions.

\section*{Data Availability Statement}
For our analyses, we used a subset of the {\sc GaitRec} dataset~\cite{horsak_gaitrec_2020}. The data and the experimental code will be made publicly available on GitHub after publication.

\bibliographystyle{ACM-Reference-Format}
\bibliography{sample-base}


\begin{thebibliography}{64}


\ifx \showCODEN    \undefined \def \showCODEN     #1{\unskip}     \fi
\ifx \showDOI      \undefined \def \showDOI       #1{#1}\fi
\ifx \showISBNx    \undefined \def \showISBNx     #1{\unskip}     \fi
\ifx \showISBNxiii \undefined \def \showISBNxiii  #1{\unskip}     \fi
\ifx \showISSN     \undefined \def \showISSN      #1{\unskip}     \fi
\ifx \showLCCN     \undefined \def \showLCCN      #1{\unskip}     \fi
\ifx \shownote     \undefined \def \shownote      #1{#1}          \fi
\ifx \showarticletitle \undefined \def \showarticletitle #1{#1}   \fi
\ifx \showURL      \undefined \def \showURL       {\relax}        \fi
\providecommand\bibfield[2]{#2}
\providecommand\bibinfo[2]{#2}
\providecommand\natexlab[1]{#1}
\providecommand\showeprint[2][]{arXiv:#2}

\bibitem[\protect\citeauthoryear{Adadi and Berrada}{Adadi and Berrada}{2018}]%
        {adadi_peeking_2018}
\bibfield{author}{\bibinfo{person}{Amina Adadi} {and} \bibinfo{person}{Mohammed
  Berrada}.} \bibinfo{year}{2018}\natexlab{}.
\newblock \showarticletitle{Peeking Inside the Black-Box: A Survey on
  Explainable Artificial Intelligence ({XAI})}.
\newblock \bibinfo{journal}{\emph{IEEE Access}}  \bibinfo{volume}{6}
  (\bibinfo{year}{2018}), \bibinfo{pages}{52138--52160}.
\newblock
\showISSN{2169-3536}
\urldef\tempurl%
\url{https://doi.org/10.1109/ACCESS.2018.2870052}
\showDOI{\tempurl}


\bibitem[\protect\citeauthoryear{Adebayo, Gilmer, Muelly, Goodfellow, Hardt,
  and Kim}{Adebayo et~al\mbox{.}}{2018}]%
        {adebayo2018sanity}
\bibfield{author}{\bibinfo{person}{Julius Adebayo}, \bibinfo{person}{Justin
  Gilmer}, \bibinfo{person}{Michael Muelly}, \bibinfo{person}{Ian Goodfellow},
  \bibinfo{person}{Moritz Hardt}, {and} \bibinfo{person}{Been Kim}.}
  \bibinfo{year}{2018}\natexlab{}.
\newblock \showarticletitle{Sanity Checks for Saliency Maps}.
\newblock In \bibinfo{booktitle}{\emph{Advances in Neural Information
  Processing Systems 31}}, \bibfield{editor}{\bibinfo{person}{S.~Bengio},
  \bibinfo{person}{H.~Wallach}, \bibinfo{person}{H.~Larochelle},
  \bibinfo{person}{K.~Grauman}, \bibinfo{person}{N.~Cesa-Bianchi}, {and}
  \bibinfo{person}{R.~Garnett}} (Eds.). \bibinfo{publisher}{Curran Associates,
  Inc.}, \bibinfo{pages}{9505--9515}.
\newblock
\urldef\tempurl%
\url{http://papers.nips.cc/paper/8160-sanity-checks-for-saliency-maps.pdf}
\showURL{%
\tempurl}


\bibitem[\protect\citeauthoryear{Alaqtash, Sarkodie-Gyan, Yu, Fuentes, Brower,
  and Abdelgawad}{Alaqtash et~al\mbox{.}}{2011}]%
        {alaqtash2011automatic}
\bibfield{author}{\bibinfo{person}{Murad Alaqtash}, \bibinfo{person}{Thompson
  Sarkodie-Gyan}, \bibinfo{person}{Huiying Yu}, \bibinfo{person}{Olac Fuentes},
  \bibinfo{person}{Richard Brower}, {and} \bibinfo{person}{Amr Abdelgawad}.}
  \bibinfo{year}{2011}\natexlab{}.
\newblock \showarticletitle{Automatic classification of pathological gait
  patterns using ground reaction forces and machine learning algorithms}. In
  \bibinfo{booktitle}{\emph{2011 Annual International Conference of the IEEE
  Engineering in Medicine and Biology Society (EMBS)}}.
  \bibinfo{publisher}{IEEE}, \bibinfo{pages}{453--457}.
\newblock
\showISBNx{978-1-4577-1589-1}
\urldef\tempurl%
\url{https://doi.org/10.1109/IEMBS.2011.6090063}
\showDOI{\tempurl}


\bibitem[\protect\citeauthoryear{Arya, Bellamy, Chen, Dhurandhar, Hind,
  Hoffman, Houde, Liao, Luss, Mojsilovi{\'c}, et~al\mbox{.}}{Arya
  et~al\mbox{.}}{2019}]%
        {arya2019one}
\bibfield{author}{\bibinfo{person}{Vijay Arya}, \bibinfo{person}{Rachel~KE
  Bellamy}, \bibinfo{person}{Pin-Yu Chen}, \bibinfo{person}{Amit Dhurandhar},
  \bibinfo{person}{Michael Hind}, \bibinfo{person}{Samuel~C Hoffman},
  \bibinfo{person}{Stephanie Houde}, \bibinfo{person}{Q~Vera Liao},
  \bibinfo{person}{Ronny Luss}, \bibinfo{person}{Aleksandra Mojsilovi{\'c}},
  {et~al\mbox{.}}} \bibinfo{year}{2019}\natexlab{}.
\newblock \showarticletitle{One Explanation Does Not Fit All: A Toolkit and
  Taxonomy of AI Explainability Techniques}.
\newblock \bibinfo{journal}{\emph{arXiv:1909.03012 [Preprint].}}
  (\bibinfo{year}{2019}).
\newblock
\newblock
\shownote{Available at: https://arxiv.org/abs/1909.03012.}


\bibitem[\protect\citeauthoryear{Bach, Binder, Montavon, Klauschen, M{\"u}ller,
  and Samek}{Bach et~al\mbox{.}}{2015}]%
        {bach2015pixel}
\bibfield{author}{\bibinfo{person}{Sebastian Bach}, \bibinfo{person}{Alexander
  Binder}, \bibinfo{person}{Gr{\'e}goire Montavon}, \bibinfo{person}{Frederick
  Klauschen}, \bibinfo{person}{Klaus-Robert M{\"u}ller}, {and}
  \bibinfo{person}{Wojciech Samek}.} \bibinfo{year}{2015}\natexlab{}.
\newblock \showarticletitle{On pixel-wise explanations for non-linear
  classifier decisions by layer-wise relevance propagation}.
\newblock \bibinfo{journal}{\emph{PloS One}} \bibinfo{volume}{10},
  \bibinfo{number}{7} (\bibinfo{year}{2015}), \bibinfo{pages}{e0130140}.
\newblock
\urldef\tempurl%
\url{https://doi.org/10.1371/journal.pone.0130140}
\showDOI{\tempurl}


\bibitem[\protect\citeauthoryear{Baehrens, Schroeter, Harmeling, Kawanabe,
  Hansen, and M{\"{u}}ller}{Baehrens et~al\mbox{.}}{2010}]%
        {baehrens2010explain}
\bibfield{author}{\bibinfo{person}{David Baehrens}, \bibinfo{person}{Timon
  Schroeter}, \bibinfo{person}{Stefan Harmeling}, \bibinfo{person}{Motoaki
  Kawanabe}, \bibinfo{person}{Katja Hansen}, {and}
  \bibinfo{person}{Klaus-Robert M{\"{u}}ller}.}
  \bibinfo{year}{2010}\natexlab{}.
\newblock \showarticletitle{How to Explain Individual Classification
  Decisions}.
\newblock \bibinfo{journal}{\emph{Journal of Machine Learning Research}}
  \bibinfo{volume}{11} (\bibinfo{year}{2010}), \bibinfo{pages}{1803--1831.}
\newblock
\urldef\tempurl%
\url{http://portal.acm.org/citation.cfm?id=1859912}
\showURL{%
\tempurl}


\bibitem[\protect\citeauthoryear{Baker}{Baker}{2013}]%
        {bakerMeasuringWalkingHandbook2013}
\bibfield{author}{\bibinfo{person}{Richard Baker}.}
  \bibinfo{year}{2013}\natexlab{}.
\newblock \bibinfo{booktitle}{\emph{Measuring Walking: A Handbook of Clinical
  Gait Analysis}}.
\newblock \bibinfo{publisher}{Mac Keith Press}, \bibinfo{address}{{London}}.
\newblock
\showISBNx{978-1-908316-66-0 978-1-908316-69-1}


\bibitem[\protect\citeauthoryear{Balduzzi, Frean, Leary, Lewis, Ma, and
  McWilliams}{Balduzzi et~al\mbox{.}}{2017}]%
        {balduzzi2017shattered}
\bibfield{author}{\bibinfo{person}{David Balduzzi}, \bibinfo{person}{Marcus
  Frean}, \bibinfo{person}{Lennox Leary}, \bibinfo{person}{J.~P. Lewis},
  \bibinfo{person}{Kurt Wan-Duo Ma}, {and} \bibinfo{person}{Brian McWilliams}.}
  \bibinfo{year}{2017}\natexlab{}.
\newblock \showarticletitle{The Shattered Gradients Problem: If resnets are the
  answer, then what is the question?}. In \bibinfo{booktitle}{\emph{Proceedings
  of the 34th International Conference on Machine Learning}}.
  \bibinfo{publisher}{PMLR}, \bibinfo{pages}{342--350.}
\newblock


\bibitem[\protect\citeauthoryear{Bizovska, Svoboda, Kutilek, Janura, Gaba, and
  Kovacikova}{Bizovska et~al\mbox{.}}{2014}]%
        {BIZOVSKA2014399}
\bibfield{author}{\bibinfo{person}{Lucia Bizovska}, \bibinfo{person}{Zdenek
  Svoboda}, \bibinfo{person}{Patrik Kutilek}, \bibinfo{person}{Miroslav
  Janura}, \bibinfo{person}{Ales Gaba}, {and} \bibinfo{person}{Zuzana
  Kovacikova}.} \bibinfo{year}{2014}\natexlab{}.
\newblock \showarticletitle{Variability of centre of pressure movement during
  gait in young and middle-aged women}.
\newblock \bibinfo{journal}{\emph{Gait {\&} Posture}} \bibinfo{volume}{40},
  \bibinfo{number}{3} (\bibinfo{year}{2014}), \bibinfo{pages}{399 -- 402}.
\newblock
\showISSN{0966-6362}
\urldef\tempurl%
\url{https://doi.org/10.1016/j.gaitpost.2014.05.065}
\showDOI{\tempurl}


\bibitem[\protect\citeauthoryear{Booth, Keijsers, Sijbers, and Huysmans}{Booth
  et~al\mbox{.}}{2018}]%
        {booth2018stapp}
\bibfield{author}{\bibinfo{person}{Brian~G Booth}, \bibinfo{person}{No{\"e}l~LW
  Keijsers}, \bibinfo{person}{Jan Sijbers}, {and} \bibinfo{person}{Toon
  Huysmans}.} \bibinfo{year}{2018}\natexlab{}.
\newblock \showarticletitle{STAPP: spatiotemporal analysis of plantar pressure
  measurements using statistical parametric mapping}.
\newblock \bibinfo{journal}{\emph{Gait \& Posture}}  \bibinfo{volume}{63}
  (\bibinfo{year}{2018}), \bibinfo{pages}{268--275}.
\newblock


\bibitem[\protect\citeauthoryear{Burdack, Horst, Giesselbach, Hassan, Daffner,
  and Schöllhorn}{Burdack et~al\mbox{.}}{2020}]%
        {burdack2020}
\bibfield{author}{\bibinfo{person}{Johannes Burdack}, \bibinfo{person}{Fabian
  Horst}, \bibinfo{person}{Sven Giesselbach}, \bibinfo{person}{Ibrahim Hassan},
  \bibinfo{person}{Sabrina Daffner}, {and} \bibinfo{person}{Wolfgang~I.
  Schöllhorn}.} \bibinfo{year}{2020}\natexlab{}.
\newblock \showarticletitle{Systematic Comparison of the Influence of Different
  Data Preprocessing Methods on the Performance of Gait Classifications Using
  Machine Learning}.
\newblock \bibinfo{journal}{\emph{Frontiers in Bioengineering and
  Biotechnology}}  \bibinfo{volume}{8} (\bibinfo{year}{2020}),
  \bibinfo{pages}{260}.
\newblock
\showISSN{2296-4185}
\urldef\tempurl%
\url{https://doi.org/10.3389/fbioe.2020.00260}
\showDOI{\tempurl}


\bibitem[\protect\citeauthoryear{Chau}{Chau}{2001}]%
        {chau_review_2001}
\bibfield{author}{\bibinfo{person}{Tom Chau}.} \bibinfo{year}{2001}\natexlab{}.
\newblock \showarticletitle{A review of analytical techniques for gait data.
  {Part} 1: fuzzy, statistical and fractal methods}.
\newblock \bibinfo{journal}{\emph{Gait \& Posture}} \bibinfo{volume}{13},
  \bibinfo{number}{1} (\bibinfo{date}{Feb.} \bibinfo{year}{2001}),
  \bibinfo{pages}{49--66}.
\newblock
\showISSN{09666362}
\urldef\tempurl%
\url{https://doi.org/10.1016/S0966-6362(00)00094-1}
\showDOI{\tempurl}


\bibitem[\protect\citeauthoryear{Chollet}{Chollet}{2017}]%
        {francois2017deep}
\bibfield{author}{\bibinfo{person}{Fran{\c{c}}ois Chollet}.}
  \bibinfo{year}{2017}\natexlab{}.
\newblock \bibinfo{booktitle}{\emph{Deep Learning with Python}}.
\newblock \bibinfo{publisher}{Manning Publications Company},
  \bibinfo{address}{Shelter Island (NY)}.
\newblock
\showISBNx{9781617294433}


\bibitem[\protect\citeauthoryear{Esteva, Kuprel, Novoa, Ko, Swetter, Blau, and
  Thrun}{Esteva et~al\mbox{.}}{2017}]%
        {Esteva.2017}
\bibfield{author}{\bibinfo{person}{Andre Esteva}, \bibinfo{person}{Brett
  Kuprel}, \bibinfo{person}{Roberto~A. Novoa}, \bibinfo{person}{Justin Ko},
  \bibinfo{person}{Susan~M. Swetter}, \bibinfo{person}{Helen~M. Blau}, {and}
  \bibinfo{person}{Sebastian Thrun}.} \bibinfo{year}{2017}\natexlab{}.
\newblock \showarticletitle{Dermatologist-level classification of skin cancer
  with deep neural networks}.
\newblock \bibinfo{journal}{\emph{Nature}} \bibinfo{volume}{542},
  \bibinfo{number}{7639} (\bibinfo{year}{2017}), \bibinfo{pages}{115--118}.
\newblock
\urldef\tempurl%
\url{https://doi.org/10.1038/nature21056}
\showDOI{\tempurl}


\bibitem[\protect\citeauthoryear{{European Union}}{{European Union}}{2016}]%
        {regulation2016regulation}
\bibfield{author}{\bibinfo{person}{{European Union}}.}
  \bibinfo{year}{2016}\natexlab{}.
\newblock \showarticletitle{Regulation {(EU)} 2016/679 of the European
  Parliament and of the Council of 27 April 2016 on the protection of natural
  persons with regard to the processing of personal data and on the free
  movement of such data, and repealing Directive 95/46/EC {(General Data
  Protection Regulation)}}.
\newblock \bibinfo{journal}{\emph{Official Journal of the European Union}}
  \bibinfo{volume}{L 119} (\bibinfo{year}{2016}), \bibinfo{pages}{1--88}.
\newblock
\newblock
\shownote{Available at: https://eur-lex.europa.eu/eli/reg/2016/679/oj.}


\bibitem[\protect\citeauthoryear{Figueiredo, Santos, and Moreno}{Figueiredo
  et~al\mbox{.}}{2018}]%
        {figueiredo_automatic_2018}
\bibfield{author}{\bibinfo{person}{Joana Figueiredo},
  \bibinfo{person}{Cristina~P. Santos}, {and} \bibinfo{person}{Juan~C.
  Moreno}.} \bibinfo{year}{2018}\natexlab{}.
\newblock \showarticletitle{Automatic recognition of gait patterns in human
  motor disorders using machine learning: {A} review}.
\newblock \bibinfo{journal}{\emph{Medical Engineering and Physics}}
  \bibinfo{volume}{53} (\bibinfo{year}{2018}), \bibinfo{pages}{1--12}.
\newblock
\showISSN{1350-4533}
\urldef\tempurl%
\url{https://doi.org/10.1016/j.medengphy.2017.12.006}
\showDOI{\tempurl}


\bibitem[\protect\citeauthoryear{Fong and Vedaldi}{Fong and Vedaldi}{2017}]%
        {fong2017interpretable}
\bibfield{author}{\bibinfo{person}{Ruth~C Fong} {and} \bibinfo{person}{Andrea
  Vedaldi}.} \bibinfo{year}{2017}\natexlab{}.
\newblock \showarticletitle{Interpretable explanations of black boxes by
  meaningful perturbation}. In \bibinfo{booktitle}{\emph{2017 IEEE
  International Conference on Computer Vision (ICCV)}}.
  \bibinfo{publisher}{IEEE}, \bibinfo{pages}{3429--3437}.
\newblock
\urldef\tempurl%
\url{https://doi.org/10.1109/ICCV.2017.371}
\showDOI{\tempurl}


\bibitem[\protect\citeauthoryear{Haenssle, Fink, Schneiderbauer, Toberer, Buhl,
  Blum, Kalloo, Hassen, Thomas, Enk, et~al\mbox{.}}{Haenssle
  et~al\mbox{.}}{2018}]%
        {haenssle2018man}
\bibfield{author}{\bibinfo{person}{Holger~A Haenssle},
  \bibinfo{person}{Christine Fink}, \bibinfo{person}{R Schneiderbauer},
  \bibinfo{person}{Ferdinand Toberer}, \bibinfo{person}{Timo Buhl},
  \bibinfo{person}{A Blum}, \bibinfo{person}{A Kalloo},
  \bibinfo{person}{A~Ben~Hadj Hassen}, \bibinfo{person}{Luc Thomas},
  \bibinfo{person}{A Enk}, {et~al\mbox{.}}} \bibinfo{year}{2018}\natexlab{}.
\newblock \showarticletitle{Man against machine: diagnostic performance of a
  deep learning convolutional neural network for dermoscopic melanoma
  recognition in comparison to 58 dermatologists}.
\newblock \bibinfo{journal}{\emph{Annals of Oncology}} \bibinfo{volume}{29},
  \bibinfo{number}{8} (\bibinfo{year}{2018}), \bibinfo{pages}{1836--1842}.
\newblock


\bibitem[\protect\citeauthoryear{Halilaj, Rajagopal, Fiterau, Hicks, Hastie,
  and Delp}{Halilaj et~al\mbox{.}}{2018}]%
        {halilaj_machine_2018}
\bibfield{author}{\bibinfo{person}{Eni Halilaj}, \bibinfo{person}{Apoorva
  Rajagopal}, \bibinfo{person}{Madalina Fiterau}, \bibinfo{person}{Jennifer~L
  Hicks}, \bibinfo{person}{Trevor~J Hastie}, {and} \bibinfo{person}{Scott~L
  Delp}.} \bibinfo{year}{2018}\natexlab{}.
\newblock \showarticletitle{Machine learning in human movement biomechanics:
  best practices, common pitfalls, and new opportunities}.
\newblock \bibinfo{journal}{\emph{Journal of Biomechanics}}
  \bibinfo{volume}{81} (\bibinfo{year}{2018}), \bibinfo{pages}{1--11}.
\newblock


\bibitem[\protect\citeauthoryear{He, Baxter, Xu, Xu, Zhou, and Zhang}{He
  et~al\mbox{.}}{2019}]%
        {He_practical_2019}
\bibfield{author}{\bibinfo{person}{Jianxing He}, \bibinfo{person}{Sally~L.
  Baxter}, \bibinfo{person}{Jie Xu}, \bibinfo{person}{Jiming Xu},
  \bibinfo{person}{Xingtao Zhou}, {and} \bibinfo{person}{Kang Zhang}.}
  \bibinfo{year}{2019}\natexlab{}.
\newblock \showarticletitle{The practical implementation of artificial
  intelligence technologies in medicine}.
\newblock \bibinfo{journal}{\emph{Nature Medicine}} \bibinfo{volume}{25},
  \bibinfo{number}{1} (\bibinfo{year}{2019}), \bibinfo{pages}{30--36}.
\newblock
\urldef\tempurl%
\url{https://doi.org/10.1038/s41591-018-0307-0}
\showDOI{\tempurl}


\bibitem[\protect\citeauthoryear{Hendricks, Akata, Rohrbach, Donahue, Schiele,
  and Darrell}{Hendricks et~al\mbox{.}}{2016}]%
        {hendricks2016generating}
\bibfield{author}{\bibinfo{person}{Lisa~Anne Hendricks},
  \bibinfo{person}{Zeynep Akata}, \bibinfo{person}{Marcus Rohrbach},
  \bibinfo{person}{Jeff Donahue}, \bibinfo{person}{Bernt Schiele}, {and}
  \bibinfo{person}{Trevor Darrell}.} \bibinfo{year}{2016}\natexlab{}.
\newblock \showarticletitle{Generating visual explanations}. In
  \bibinfo{booktitle}{\emph{European Conference on Computer Vision (ECCV)}}.
  Springer, \bibinfo{pages}{3--19}.
\newblock
\urldef\tempurl%
\url{https://doi.org/10.1007/978-3-319-46493-0_1}
\showDOI{\tempurl}


\bibitem[\protect\citeauthoryear{Holzinger, Biemann, Pattichis, and
  Kell}{Holzinger et~al\mbox{.}}{2017}]%
        {holzinger_what_2017}
\bibfield{author}{\bibinfo{person}{Andreas Holzinger}, \bibinfo{person}{Chris
  Biemann}, \bibinfo{person}{Constantinos~S. Pattichis}, {and}
  \bibinfo{person}{Douglas~B. Kell}.} \bibinfo{year}{2017}\natexlab{}.
\newblock \showarticletitle{What do we need to build explainable {AI} systems
  for the medical domain?}
\newblock \bibinfo{journal}{\emph{arXiv:1712.09923 [Preprint].}}
  (\bibinfo{date}{Dec.} \bibinfo{year}{2017}).
\newblock
\urldef\tempurl%
\url{http://arxiv.org/abs/1712.09923}
\showURL{%
\tempurl}
\newblock
\shownote{Available at: https://arxiv.org/abs/1712.09923.}


\bibitem[\protect\citeauthoryear{Holzinger, Langs, Denk, Zatloukal, and
  Müller}{Holzinger et~al\mbox{.}}{2019}]%
        {holzinger_causability_2019}
\bibfield{author}{\bibinfo{person}{Andreas Holzinger}, \bibinfo{person}{Georg
  Langs}, \bibinfo{person}{Helmut Denk}, \bibinfo{person}{Kurt Zatloukal},
  {and} \bibinfo{person}{Heimo Müller}.} \bibinfo{year}{2019}\natexlab{}.
\newblock \showarticletitle{Causability and explainability of artificial
  intelligence in medicine}.
\newblock \bibinfo{journal}{\emph{Wiley Interdisciplinary Reviews: Data Mining
  and Knowledge Discovery}} \bibinfo{volume}{9}, \bibinfo{number}{4}
  (\bibinfo{date}{July} \bibinfo{year}{2019}), \bibinfo{pages}{e1312}.
\newblock
\showISSN{1942-4787}
\urldef\tempurl%
\url{https://doi.org/10.1002/widm.1312}
\showDOI{\tempurl}


\bibitem[\protect\citeauthoryear{Horsak, Slijepcevic, Raberger, Schwab,
  Worisch, and Zeppelzauer}{Horsak et~al\mbox{.}}{2020}]%
        {horsak_gaitrec_2020}
\bibfield{author}{\bibinfo{person}{Brian Horsak}, \bibinfo{person}{Djordje
  Slijepcevic}, \bibinfo{person}{Anna-Maria Raberger},
  \bibinfo{person}{Caterine Schwab}, \bibinfo{person}{Marianne Worisch}, {and}
  \bibinfo{person}{Matthias Zeppelzauer}.} \bibinfo{year}{2020}\natexlab{}.
\newblock \showarticletitle{{GaitRec}, a large-scale ground reaction force
  dataset of healthy and impaired gait}.
\newblock \bibinfo{journal}{\emph{Scientific Data}} \bibinfo{volume}{7},
  \bibinfo{number}{1} (\bibinfo{date}{May} \bibinfo{year}{2020}),
  \bibinfo{pages}{1--8}.
\newblock
\showISSN{2052-4463}
\urldef\tempurl%
\url{https://doi.org/10.1038/s41597-020-0481-z}
\showDOI{\tempurl}


\bibitem[\protect\citeauthoryear{Horst, Lapuschkin, Samek, M{\"u}ller, and
  Sch{\"o}llhorn}{Horst et~al\mbox{.}}{2019}]%
        {horst2019explaining}
\bibfield{author}{\bibinfo{person}{Fabian Horst}, \bibinfo{person}{Sebastian
  Lapuschkin}, \bibinfo{person}{Wojciech Samek}, \bibinfo{person}{Klaus-Robert
  M{\"u}ller}, {and} \bibinfo{person}{Wolfgang~I Sch{\"o}llhorn}.}
  \bibinfo{year}{2019}\natexlab{}.
\newblock \showarticletitle{Explaining the unique nature of individual gait
  patterns with deep learning}.
\newblock \bibinfo{journal}{\emph{Scientific Reports}} \bibinfo{volume}{9},
  \bibinfo{number}{1} (\bibinfo{year}{2019}), \bibinfo{pages}{2391}.
\newblock
\urldef\tempurl%
\url{https://doi.org/10.1038/s41598-019-38748-8}
\showDOI{\tempurl}


\bibitem[\protect\citeauthoryear{Hsu, Chang, and Lin}{Hsu
  et~al\mbox{.}}{2016}]%
        {Hsu.2016}
\bibfield{author}{\bibinfo{person}{Chih-Wei Hsu}, \bibinfo{person}{Chih-Chung
  Chang}, {and} \bibinfo{person}{Chih-Jen Lin}.}
  \bibinfo{year}{2016}\natexlab{}.
\newblock \emph{\bibinfo{title}{A Practical Guide to Support Vector
  Classification}}.
\newblock Technical Report. \bibinfo{school}{{National Taiwan University}}.
\newblock
\newblock
\shownote{Available at:
  https://www.csie.ntu.edu.tw/~cjlin/papers/guide/guide.pdf.}


\bibitem[\protect\citeauthoryear{Kohlbrenner, Bauer, Nakajima, Binder, Samek,
  and Lapuschkin}{Kohlbrenner et~al\mbox{.}}{2019}]%
        {kohlbrenner2019towards}
\bibfield{author}{\bibinfo{person}{Maximilian Kohlbrenner},
  \bibinfo{person}{Alexander Bauer}, \bibinfo{person}{Shinichi Nakajima},
  \bibinfo{person}{Alexander Binder}, \bibinfo{person}{Wojciech Samek}, {and}
  \bibinfo{person}{Sebastian Lapuschkin}.} \bibinfo{year}{2019}\natexlab{}.
\newblock \showarticletitle{Towards best practice in explaining neural network
  decisions with LRP}.
\newblock \bibinfo{journal}{\emph{arXiv:1910.09840 [Preprint].}}
  (\bibinfo{year}{2019}).
\newblock
\newblock
\shownote{Available at: http://arxiv.org/abs/1910.09840.}


\bibitem[\protect\citeauthoryear{Lapuschkin, W{\"a}ldchen, Binder, Montavon,
  Samek, and M{\"u}ller}{Lapuschkin et~al\mbox{.}}{2019}]%
        {lapuschkin2019unmasking}
\bibfield{author}{\bibinfo{person}{Sebastian Lapuschkin},
  \bibinfo{person}{Stephan W{\"a}ldchen}, \bibinfo{person}{Alexander Binder},
  \bibinfo{person}{Gr{\'e}goire Montavon}, \bibinfo{person}{Wojciech Samek},
  {and} \bibinfo{person}{Klaus-Robert M{\"u}ller}.}
  \bibinfo{year}{2019}\natexlab{}.
\newblock \showarticletitle{Unmasking Clever Hans Predictors and Assessing What
  Machines Really Learn}.
\newblock \bibinfo{journal}{\emph{Nature Communications}}  \bibinfo{volume}{10}
  (\bibinfo{year}{2019}), \bibinfo{pages}{1096}.
\newblock
\urldef\tempurl%
\url{https://doi.org/10.1038/s41467-019-08987-4}
\showDOI{\tempurl}


\bibitem[\protect\citeauthoryear{Lau, Tong, and Zhu}{Lau et~al\mbox{.}}{2009}]%
        {lau_support_2009}
\bibfield{author}{\bibinfo{person}{Hong-yin Lau}, \bibinfo{person}{Kai-yu
  Tong}, {and} \bibinfo{person}{Hailong Zhu}.} \bibinfo{year}{2009}\natexlab{}.
\newblock \showarticletitle{Support vector machine for classification of
  walking conditions of persons after stroke with dropped foot}.
\newblock \bibinfo{journal}{\emph{Human Movement Science}}
  \bibinfo{volume}{28}, \bibinfo{number}{4} (\bibinfo{date}{Aug.}
  \bibinfo{year}{2009}), \bibinfo{pages}{504--514}.
\newblock
\showISSN{01679457}
\urldef\tempurl%
\url{https://doi.org/10.1016/j.humov.2008.12.003}
\showDOI{\tempurl}


\bibitem[\protect\citeauthoryear{LeCun, Bottou, Orr, and M{\"{u}}ller}{LeCun
  et~al\mbox{.}}{2012}]%
        {lecun2012efficient}
\bibfield{author}{\bibinfo{person}{Yann LeCun}, \bibinfo{person}{L{\'{e}}on
  Bottou}, \bibinfo{person}{Genevieve~B. Orr}, {and}
  \bibinfo{person}{Klaus-Robert M{\"{u}}ller}.}
  \bibinfo{year}{2012}\natexlab{}.
\newblock \showarticletitle{Efficient BackProp}.
\newblock In \bibinfo{booktitle}{\emph{Neural Networks: Tricks of the Trade -
  Second Edition}}. \bibinfo{publisher}{Springer}, \bibinfo{pages}{9--48}.
\newblock
\urldef\tempurl%
\url{https://doi.org/10.1007/978-3-642-35289-8\_3}
\showDOI{\tempurl}


\bibitem[\protect\citeauthoryear{Lundberg and Lee}{Lundberg and Lee}{2017}]%
        {lundberg2017unified}
\bibfield{author}{\bibinfo{person}{Scott~M. Lundberg} {and}
  \bibinfo{person}{Su-In Lee}.} \bibinfo{year}{2017}\natexlab{}.
\newblock \showarticletitle{A unified approach to interpreting model
  predictions}. In \bibinfo{booktitle}{\emph{Advances in Neural Information
  Processing Systems (NIPS)}}. \bibinfo{publisher}{Curran Associates, Inc.},
  \bibinfo{pages}{4765--4774}.
\newblock
\newblock
\shownote{Available at:
  http://papers.nips.cc/paper/7062-a-unified-approach-to-interpreting-model-predictions.pdf.}


\bibitem[\protect\citeauthoryear{Maaten and Hinton}{Maaten and Hinton}{2008}]%
        {maaten2008visualizing}
\bibfield{author}{\bibinfo{person}{Laurens van~der Maaten} {and}
  \bibinfo{person}{Geoffrey Hinton}.} \bibinfo{year}{2008}\natexlab{}.
\newblock \showarticletitle{Visualizing data using t-SNE}.
\newblock \bibinfo{journal}{\emph{Journal of Machine Learning Research}}
  \bibinfo{volume}{9}, \bibinfo{number}{Nov} (\bibinfo{year}{2008}),
  \bibinfo{pages}{2579--2605.}
\newblock


\bibitem[\protect\citeauthoryear{McKinney, Sieniek, Godbole, Godwin, Antropova,
  Ashrafian, Back, Chesus, Corrado, Darzi, et~al\mbox{.}}{McKinney
  et~al\mbox{.}}{2020}]%
        {mckinney2020international}
\bibfield{author}{\bibinfo{person}{Scott~Mayer McKinney},
  \bibinfo{person}{Marcin Sieniek}, \bibinfo{person}{Varun Godbole},
  \bibinfo{person}{Jonathan Godwin}, \bibinfo{person}{Natasha Antropova},
  \bibinfo{person}{Hutan Ashrafian}, \bibinfo{person}{Trevor Back},
  \bibinfo{person}{Mary Chesus}, \bibinfo{person}{Greg~C Corrado},
  \bibinfo{person}{Ara Darzi}, {et~al\mbox{.}}}
  \bibinfo{year}{2020}\natexlab{}.
\newblock \showarticletitle{International evaluation of an AI system for breast
  cancer screening}.
\newblock \bibinfo{journal}{\emph{Nature}} \bibinfo{volume}{577},
  \bibinfo{number}{7788} (\bibinfo{year}{2020}), \bibinfo{pages}{89--94}.
\newblock


\bibitem[\protect\citeauthoryear{Montavon, Binder, Lapuschkin, Samek, and
  M{\"u}ller}{Montavon et~al\mbox{.}}{2019}]%
        {montavon2019layer}
\bibfield{author}{\bibinfo{person}{Gr{\'e}goire Montavon},
  \bibinfo{person}{Alexander Binder}, \bibinfo{person}{Sebastian Lapuschkin},
  \bibinfo{person}{Wojciech Samek}, {and} \bibinfo{person}{Klaus-Robert
  M{\"u}ller}.} \bibinfo{year}{2019}\natexlab{}.
\newblock \showarticletitle{Layer-wise Relevance Propagation: An Overview}. In
  \bibinfo{booktitle}{\emph{Explainable AI: Interpreting, Explaining and
  Visualizing Deep Learning}}. \bibinfo{publisher}{Springer},
  \bibinfo{pages}{193--209}.
\newblock
\urldef\tempurl%
\url{https://doi.org/10.1007/978-3-030-28954-6_10}
\showDOI{\tempurl}


\bibitem[\protect\citeauthoryear{Montavon, Samek, and M{\"u}ller}{Montavon
  et~al\mbox{.}}{2018}]%
        {montavon2018methods}
\bibfield{author}{\bibinfo{person}{Gr{\'e}goire Montavon},
  \bibinfo{person}{Wojciech Samek}, {and} \bibinfo{person}{Klaus-Robert
  M{\"u}ller}.} \bibinfo{year}{2018}\natexlab{}.
\newblock \showarticletitle{Methods for interpreting and understanding deep
  neural networks}.
\newblock \bibinfo{journal}{\emph{Digital Signal Processing}}
  \bibinfo{volume}{73} (\bibinfo{year}{2018}), \bibinfo{pages}{1--15}.
\newblock
\urldef\tempurl%
\url{https://doi.org/10.1016/j.dsp.2017.10.011}
\showDOI{\tempurl}


\bibitem[\protect\citeauthoryear{Nguyen, Dosovitskiy, Yosinski, Brox, and
  Clune}{Nguyen et~al\mbox{.}}{2016}]%
        {nguyen2016synthesizing}
\bibfield{author}{\bibinfo{person}{Anh Nguyen}, \bibinfo{person}{Alexey
  Dosovitskiy}, \bibinfo{person}{Jason Yosinski}, \bibinfo{person}{Thomas
  Brox}, {and} \bibinfo{person}{Jeff Clune}.} \bibinfo{year}{2016}\natexlab{}.
\newblock \showarticletitle{Synthesizing the preferred inputs for neurons in
  neural networks via deep generator networks}. In
  \bibinfo{booktitle}{\emph{Advances in Neural Information Processing
  Systems}}. \bibinfo{publisher}{Curran Associates, Inc.},
  \bibinfo{pages}{3387--3395}.
\newblock
\newblock
\shownote{Available at:
  http://papers.nips.cc/paper/6519-synthesizing-the-preferred-inputs-for-neurons-in-neural-networks-via-deep-generator-networks.pdf.}


\bibitem[\protect\citeauthoryear{Nieuwenhuys, Papageorgiou, Desloovere,
  Molenaers, and De~Laet}{Nieuwenhuys et~al\mbox{.}}{2017}]%
        {nieuwenhuys2017statistical}
\bibfield{author}{\bibinfo{person}{Angela Nieuwenhuys}, \bibinfo{person}{Eirini
  Papageorgiou}, \bibinfo{person}{Kaat Desloovere}, \bibinfo{person}{Guy
  Molenaers}, {and} \bibinfo{person}{Tinne De~Laet}.}
  \bibinfo{year}{2017}\natexlab{}.
\newblock \showarticletitle{Statistical parametric mapping to identify
  differences between consensus-based joint patterns during gait in children
  with cerebral palsy}.
\newblock \bibinfo{journal}{\emph{PLoS One}} \bibinfo{volume}{12},
  \bibinfo{number}{1} (\bibinfo{year}{2017}).
\newblock


\bibitem[\protect\citeauthoryear{Nüesch, Valderrabano, Huber, von Tscharner,
  and Pagenstert}{Nüesch et~al\mbox{.}}{2012}]%
        {nuesch_gait_2012}
\bibfield{author}{\bibinfo{person}{Corina Nüesch}, \bibinfo{person}{Victor
  Valderrabano}, \bibinfo{person}{Cora Huber}, \bibinfo{person}{Vinzenz von
  Tscharner}, {and} \bibinfo{person}{Geert Pagenstert}.}
  \bibinfo{year}{2012}\natexlab{}.
\newblock \showarticletitle{Gait patterns of asymmetric ankle osteoarthritis
  patients}.
\newblock \bibinfo{journal}{\emph{Clinical Biomechanics}} \bibinfo{volume}{27},
  \bibinfo{number}{6} (\bibinfo{date}{July} \bibinfo{year}{2012}),
  \bibinfo{pages}{613--618}.
\newblock
\showISSN{0268-0033}
\urldef\tempurl%
\url{https://doi.org/10.1016/j.clinbiomech.2011.12.016}
\showDOI{\tempurl}


\bibitem[\protect\citeauthoryear{Pataky}{Pataky}{2010}]%
        {pataky_generalized_2010}
\bibfield{author}{\bibinfo{person}{Todd~C. Pataky}.}
  \bibinfo{year}{2010}\natexlab{}.
\newblock \showarticletitle{Generalized n-dimensional biomechanical field
  analysis using statistical parametric mapping}.
\newblock \bibinfo{journal}{\emph{Journal of Biomechanics}}
  \bibinfo{volume}{43}, \bibinfo{number}{10} (\bibinfo{date}{July}
  \bibinfo{year}{2010}), \bibinfo{pages}{1976--1982}.
\newblock
\showISSN{00219290}
\urldef\tempurl%
\url{https://doi.org/10.1016/j.jbiomech.2010.03.008}
\showDOI{\tempurl}


\bibitem[\protect\citeauthoryear{Pataky}{Pataky}{2012}]%
        {pataky_one-dimensional_2012}
\bibfield{author}{\bibinfo{person}{Todd~C. Pataky}.}
  \bibinfo{year}{2012}\natexlab{}.
\newblock \showarticletitle{One-dimensional statistical parametric mapping in
  Python}.
\newblock \bibinfo{journal}{\emph{Computer Methods in Biomechanics and
  Biomedical Engineering}} \bibinfo{volume}{15}, \bibinfo{number}{3}
  (\bibinfo{date}{March} \bibinfo{year}{2012}), \bibinfo{pages}{295--301}.
\newblock
\showISSN{1025-5842, 1476-8259}
\urldef\tempurl%
\url{https://doi.org/10.1080/10255842.2010.527837}
\showDOI{\tempurl}


\bibitem[\protect\citeauthoryear{Phinyomark, Petri, Ib{\'a}{\~n}ez-Marcelo,
  Osis, and Ferber}{Phinyomark et~al\mbox{.}}{2018}]%
        {phinyomark_analysis_2018}
\bibfield{author}{\bibinfo{person}{Angkoon Phinyomark},
  \bibinfo{person}{Giovanni Petri}, \bibinfo{person}{Esther
  Ib{\'a}{\~n}ez-Marcelo}, \bibinfo{person}{Sean~T. Osis}, {and}
  \bibinfo{person}{Reed Ferber}.} \bibinfo{year}{2018}\natexlab{}.
\newblock \showarticletitle{Analysis of big data in gait biomechanics: Current
  trends and future directions}.
\newblock \bibinfo{journal}{\emph{Journal of Medical and Biological
  Engineering}} \bibinfo{volume}{38}, \bibinfo{number}{2}
  (\bibinfo{year}{2018}), \bibinfo{pages}{244--260}.
\newblock
\showISSN{1609-0985}
\urldef\tempurl%
\url{https://doi.org/10.1007/s40846-017-0297-2}
\showDOI{\tempurl}


\bibitem[\protect\citeauthoryear{Ribeiro, Singh, and Guestrin}{Ribeiro
  et~al\mbox{.}}{2016}]%
        {ribeiro2016model}
\bibfield{author}{\bibinfo{person}{Marco~Tulio Ribeiro},
  \bibinfo{person}{Sameer Singh}, {and} \bibinfo{person}{Carlos Guestrin}.}
  \bibinfo{year}{2016}\natexlab{}.
\newblock \showarticletitle{Model-agnostic interpretability of machine
  learning}.
\newblock \bibinfo{journal}{\emph{arXiv:1606.05386 [Preprint].}}
  (\bibinfo{year}{2016}).
\newblock
\newblock
\shownote{Available at: http://arxiv.org/abs/1606.05386.}


\bibitem[\protect\citeauthoryear{Rosenthal}{Rosenthal}{1986}]%
        {rosenthal_meta-analytic_1986}
\bibfield{author}{\bibinfo{person}{Robert Rosenthal}.}
  \bibinfo{year}{1986}\natexlab{}.
\newblock \showarticletitle{Meta-{Analytic} {Procedures} for {Social} {Science}
  {Research} {Sage} {Publications}: {Beverly} {Hills}, 1984, 148 pp.}
\newblock \bibinfo{journal}{\emph{Educational Researcher}}
  \bibinfo{volume}{15}, \bibinfo{number}{8} (\bibinfo{date}{Oct.}
  \bibinfo{year}{1986}), \bibinfo{pages}{18--20}.
\newblock
\showISSN{0013-189X}
\urldef\tempurl%
\url{https://doi.org/10.3102/0013189X015008018}
\showDOI{\tempurl}


\bibitem[\protect\citeauthoryear{Samek, Binder, Montavon, Lapuschkin, and
  M{\"u}ller}{Samek et~al\mbox{.}}{2016}]%
        {samek2016evaluating}
\bibfield{author}{\bibinfo{person}{Wojciech Samek}, \bibinfo{person}{Alexander
  Binder}, \bibinfo{person}{Gr{\'e}goire Montavon}, \bibinfo{person}{Sebastian
  Lapuschkin}, {and} \bibinfo{person}{Klaus-Robert M{\"u}ller}.}
  \bibinfo{year}{2016}\natexlab{}.
\newblock \showarticletitle{Evaluating the visualization of what a deep neural
  network has learned}.
\newblock \bibinfo{journal}{\emph{IEEE Transactions on Neural Networks and
  Learning Systems}} \bibinfo{volume}{28}, \bibinfo{number}{11}
  (\bibinfo{year}{2016}), \bibinfo{pages}{2660--2673}.
\newblock


\bibitem[\protect\citeauthoryear{{Samek}, {Binder}, {Montavon}, {Lapuschkin},
  and {Müller}}{{Samek} et~al\mbox{.}}{2017}]%
        {samek_2017_evaluating}
\bibfield{author}{\bibinfo{person}{Wojciech {Samek}},
  \bibinfo{person}{Alexander {Binder}}, \bibinfo{person}{Gr{\'e}goire
  {Montavon}}, \bibinfo{person}{Sebastian {Lapuschkin}}, {and}
  \bibinfo{person}{Klaus-Robert {Müller}}.} \bibinfo{year}{2017}\natexlab{}.
\newblock \showarticletitle{Evaluating the Visualization of What a Deep Neural
  Network Has Learned}.
\newblock \bibinfo{journal}{\emph{IEEE Transactions on Neural Networks and
  Learning Systems}} \bibinfo{volume}{28}, \bibinfo{number}{11}
  (\bibinfo{date}{Nov} \bibinfo{year}{2017}), \bibinfo{pages}{2660--2673}.
\newblock
\urldef\tempurl%
\url{https://doi.org/10.1109/TNNLS.2016.2599820}
\showDOI{\tempurl}


\bibitem[\protect\citeauthoryear{Samek, Montavon, Lapuschkin, Anders, and
  M{\"u}ller}{Samek et~al\mbox{.}}{2020}]%
        {samek2020toward}
\bibfield{author}{\bibinfo{person}{Wojciech Samek},
  \bibinfo{person}{Gr{\'e}goire Montavon}, \bibinfo{person}{Sebastian
  Lapuschkin}, \bibinfo{person}{Christopher~J Anders}, {and}
  \bibinfo{person}{Klaus-Robert M{\"u}ller}.} \bibinfo{year}{2020}\natexlab{}.
\newblock \showarticletitle{Toward Interpretable Machine Learning: Transparent
  Deep Neural Networks and Beyond}.
\newblock \bibinfo{journal}{\emph{arXiv preprint arXiv:2003.07631}}
  (\bibinfo{year}{2020}).
\newblock


\bibitem[\protect\citeauthoryear{Samek, Wiegand, and M{\"{u}}ller}{Samek
  et~al\mbox{.}}{2017}]%
        {samek_explaining_2017}
\bibfield{author}{\bibinfo{person}{Wojciech Samek}, \bibinfo{person}{Thomas
  Wiegand}, {and} \bibinfo{person}{Klaus-Robert M{\"{u}}ller}.}
  \bibinfo{year}{2017}\natexlab{}.
\newblock \showarticletitle{Explainable Artificial Intelligence: Understanding,
  Visualizing and Interpreting Deep Learning Models}.
\newblock \bibinfo{journal}{\emph{ITU Journal: ICT Discoveries}}
  \bibinfo{volume}{1}, \bibinfo{number}{1} (\bibinfo{year}{2017}),
  \bibinfo{pages}{39--48}.
\newblock


\bibitem[\protect\citeauthoryear{Sch{\"o}llhorn}{Sch{\"o}llhorn}{2004}]%
        {schollhorn_applications_2004}
\bibfield{author}{\bibinfo{person}{Wolfgang~I Sch{\"o}llhorn}.}
  \bibinfo{year}{2004}\natexlab{}.
\newblock \showarticletitle{Applications of artificial neural nets in clinical
  biomechanics}.
\newblock \bibinfo{journal}{\emph{Clinical Biomechanics}} \bibinfo{volume}{19},
  \bibinfo{number}{9} (\bibinfo{year}{2004}), \bibinfo{pages}{876--898}.
\newblock
\showISSN{02680033}
\urldef\tempurl%
\url{https://doi.org/10.1016/j.clinbiomech.2004.04.005}
\showDOI{\tempurl}


\bibitem[\protect\citeauthoryear{Shi, Huang, Yu, Liang, Zhang, Yu, Liu, and
  Ao}{Shi et~al\mbox{.}}{2018}]%
        {shi2018effect}
\bibfield{author}{\bibinfo{person}{Huijuan Shi}, \bibinfo{person}{Hongshi
  Huang}, \bibinfo{person}{Yuanyuan Yu}, \bibinfo{person}{Zixuan Liang},
  \bibinfo{person}{Si Zhang}, \bibinfo{person}{Bing Yu}, \bibinfo{person}{Hui
  Liu}, {and} \bibinfo{person}{Yingfang Ao}.} \bibinfo{year}{2018}\natexlab{}.
\newblock \showarticletitle{Effect of dual task on gait asymmetry in patients
  after anterior cruciate ligament reconstruction}.
\newblock \bibinfo{journal}{\emph{Scientific Reports}} \bibinfo{volume}{8},
  \bibinfo{number}{1} (\bibinfo{year}{2018}), \bibinfo{pages}{1--10}.
\newblock


\bibitem[\protect\citeauthoryear{Shrikumar, Greenside, and Kundaje}{Shrikumar
  et~al\mbox{.}}{2017}]%
        {shrikumar2017learning}
\bibfield{author}{\bibinfo{person}{Avanti Shrikumar}, \bibinfo{person}{Peyton
  Greenside}, {and} \bibinfo{person}{Anshul Kundaje}.}
  \bibinfo{year}{2017}\natexlab{}.
\newblock \showarticletitle{Learning important features through propagating
  activation differences}. In \bibinfo{booktitle}{\emph{Proceedings of the 34th
  International Conference on Machine Learning}}. \bibinfo{publisher}{PMLR},
  \bibinfo{pages}{3145--3153.}
\newblock


\bibitem[\protect\citeauthoryear{Simonyan, Vedaldi, and Zisserman}{Simonyan
  et~al\mbox{.}}{2013}]%
        {simonyan2013deep}
\bibfield{author}{\bibinfo{person}{Karen Simonyan}, \bibinfo{person}{Andrea
  Vedaldi}, {and} \bibinfo{person}{Andrew Zisserman}.}
  \bibinfo{year}{2013}\natexlab{}.
\newblock \showarticletitle{Deep inside convolutional networks: Visualising
  image classification models and saliency maps}.
\newblock \bibinfo{journal}{\emph{arXiv:1312.6034 [Preprint].}}
  (\bibinfo{year}{2013}).
\newblock
\newblock
\shownote{Available at: http://arxiv.org/abs/1312.6034.}


\bibitem[\protect\citeauthoryear{Slijepcevic, Zeppelzauer, Gorgas, Schwab,
  Sch{\"u}ller, Baca, Breiteneder, and Horsak}{Slijepcevic
  et~al\mbox{.}}{2017}]%
        {slijepcevic2017automatic}
\bibfield{author}{\bibinfo{person}{Djordje Slijepcevic},
  \bibinfo{person}{Matthias Zeppelzauer}, \bibinfo{person}{Anna-Maria Gorgas},
  \bibinfo{person}{Caterine Schwab}, \bibinfo{person}{Michael Sch{\"u}ller},
  \bibinfo{person}{Arnold Baca}, \bibinfo{person}{Christian Breiteneder}, {and}
  \bibinfo{person}{Brian Horsak}.} \bibinfo{year}{2017}\natexlab{}.
\newblock \showarticletitle{Automatic classification of functional gait
  disorders}.
\newblock \bibinfo{journal}{\emph{IEEE Journal of Biomedical and Health
  Informatics}} \bibinfo{volume}{22}, \bibinfo{number}{5}
  (\bibinfo{year}{2017}), \bibinfo{pages}{1653--1661}.
\newblock
\urldef\tempurl%
\url{https://doi.org/10.1109/JBHI.2017.2785682}
\showDOI{\tempurl}


\bibitem[\protect\citeauthoryear{Slijepcevic, Zeppelzauer, Schwab, Raberger,
  Breiteneder, and Horsak}{Slijepcevic et~al\mbox{.}}{2019}]%
        {SLIJEPCEVIC2019}
\bibfield{author}{\bibinfo{person}{Djordje Slijepcevic},
  \bibinfo{person}{Matthias Zeppelzauer}, \bibinfo{person}{Caterine Schwab},
  \bibinfo{person}{Anna-Maria Raberger}, \bibinfo{person}{Christian
  Breiteneder}, {and} \bibinfo{person}{Brian Horsak}.}
  \bibinfo{year}{2019}\natexlab{}.
\newblock \showarticletitle{{Input Representations and Classification
  Strategies for Automated Human Gait Analysis}}.
\newblock \bibinfo{journal}{\emph{Gait {\&} Posture}} (\bibinfo{year}{2019}).
\newblock
\showISSN{0966-6362}
\urldef\tempurl%
\url{https://doi.org/10.1016/j.gaitpost.2019.10.021}
\showDOI{\tempurl}


\bibitem[\protect\citeauthoryear{Slijepcevic, Zeppelzauer, Schwab, Raberger,
  Dumphart, Baca, Breiteneder, and Horsak}{Slijepcevic et~al\mbox{.}}{2018}]%
        {slijepcevic2018p}
\bibfield{author}{\bibinfo{person}{Djordje Slijepcevic},
  \bibinfo{person}{Matthias Zeppelzauer}, \bibinfo{person}{Caterine Schwab},
  \bibinfo{person}{Anna-Maria Raberger}, \bibinfo{person}{Bernhard Dumphart},
  \bibinfo{person}{Arnold Baca}, \bibinfo{person}{Christian Breiteneder}, {and}
  \bibinfo{person}{Brian Horsak}.} \bibinfo{year}{2018}\natexlab{}.
\newblock \showarticletitle{P 011-{Towards} an optimal combination of input
  signals and derived representations for gait classification based on ground
  reaction force measurements.}
\newblock \bibinfo{journal}{\emph{Gait {\&} Posture}}  \bibinfo{volume}{65}
  (\bibinfo{year}{2018}), \bibinfo{pages}{249}.
\newblock
\urldef\tempurl%
\url{https://doi.org/10.1016/j.gaitpost.2018.06.155}
\showDOI{\tempurl}


\bibitem[\protect\citeauthoryear{{\v{S}}trumbelj and Kononenko}{{\v{S}}trumbelj
  and Kononenko}{2014}]%
        {vstrumbelj2014explaining}
\bibfield{author}{\bibinfo{person}{Erik {\v{S}}trumbelj} {and}
  \bibinfo{person}{Igor Kononenko}.} \bibinfo{year}{2014}\natexlab{}.
\newblock \showarticletitle{Explaining prediction models and individual
  predictions with feature contributions}.
\newblock \bibinfo{journal}{\emph{Knowledge and Information Systems}}
  \bibinfo{volume}{41}, \bibinfo{number}{3} (\bibinfo{year}{2014}),
  \bibinfo{pages}{647--665}.
\newblock
\urldef\tempurl%
\url{https://doi.org/10.1007/s10115-013-0679-x}
\showDOI{\tempurl}


\bibitem[\protect\citeauthoryear{Tjoa and Guan}{Tjoa and Guan}{2019}]%
        {tjoa_survey_2019}
\bibfield{author}{\bibinfo{person}{Erico Tjoa} {and} \bibinfo{person}{Cuntai
  Guan}.} \bibinfo{year}{2019}\natexlab{}.
\newblock \showarticletitle{A Survey on Explainable Artificial Intelligence
  (XAI): Towards Medical XAI}.
\newblock \bibinfo{journal}{\emph{arXiv:1907.07374 [Preprint].}}
  (\bibinfo{year}{2019}).
\newblock
\newblock
\shownote{Available at: https://arxiv.org/abs/1907.07374.}


\bibitem[\protect\citeauthoryear{Topol}{Topol}{2019}]%
        {topol_medicine_2019}
\bibfield{author}{\bibinfo{person}{Eric~J Topol}.}
  \bibinfo{year}{2019}\natexlab{}.
\newblock \showarticletitle{High-performance medicine: The convergence of human
  and artificial intelligence}.
\newblock \bibinfo{journal}{\emph{Nature Medicine}} \bibinfo{volume}{25},
  \bibinfo{number}{1} (\bibinfo{year}{2019}), \bibinfo{pages}{44--56}.
\newblock
\urldef\tempurl%
\url{https://doi.org/10.1038/s41591-018-0300-7}
\showDOI{\tempurl}


\bibitem[\protect\citeauthoryear{Van~Gestel, De~Laet, Di~Lello, Bruyninckx,
  Molenaers, Van~Campenhout, Aertbeliën, Schwartz, Wambacq, De~Cock, and
  Desloovere}{Van~Gestel et~al\mbox{.}}{2011}]%
        {van_gestel_probabilistic_2011}
\bibfield{author}{\bibinfo{person}{Leen Van~Gestel}, \bibinfo{person}{Tinne
  De~Laet}, \bibinfo{person}{Enrico Di~Lello}, \bibinfo{person}{Herman
  Bruyninckx}, \bibinfo{person}{Guy Molenaers}, \bibinfo{person}{Anja
  Van~Campenhout}, \bibinfo{person}{Erwin Aertbeliën}, \bibinfo{person}{Mike
  Schwartz}, \bibinfo{person}{Hans Wambacq}, \bibinfo{person}{Paul De~Cock},
  {and} \bibinfo{person}{Kaat Desloovere}.} \bibinfo{year}{2011}\natexlab{}.
\newblock \showarticletitle{Probabilistic gait classification in children with
  cerebral palsy: {A} {Bayesian} approach}.
\newblock \bibinfo{journal}{\emph{Research in Developmental Disabilities}}
  \bibinfo{volume}{32}, \bibinfo{number}{6} (\bibinfo{date}{Nov.}
  \bibinfo{year}{2011}), \bibinfo{pages}{2542--2552}.
\newblock
\showISSN{0891-4222}
\urldef\tempurl%
\url{https://doi.org/10.1016/j.ridd.2011.07.004}
\showDOI{\tempurl}


\bibitem[\protect\citeauthoryear{Wagner, Slijepcevic, Horsak, Rind,
  Zeppelzauer, and Aigner}{Wagner et~al\mbox{.}}{2018}]%
        {wagner_KAVAG}
\bibfield{author}{\bibinfo{person}{Markus Wagner}, \bibinfo{person}{Djordje
  Slijepcevic}, \bibinfo{person}{Brian Horsak}, \bibinfo{person}{Alexander
  Rind}, \bibinfo{person}{Matthias Zeppelzauer}, {and}
  \bibinfo{person}{Wolfgang Aigner}.} \bibinfo{year}{2018}\natexlab{}.
\newblock \showarticletitle{KAVAGait: Knowledge-assisted visual analytics for
  clinical gait analysis}.
\newblock \bibinfo{journal}{\emph{IEEE Transactions on Visualization and
  Computer Graphics}} \bibinfo{volume}{25}, \bibinfo{number}{3}
  (\bibinfo{year}{2018}), \bibinfo{pages}{1528--1542}.
\newblock


\bibitem[\protect\citeauthoryear{Wahid, Begg, Hass, Halgamuge, and
  Ackland}{Wahid et~al\mbox{.}}{2015}]%
        {wahid_classification_2015}
\bibfield{author}{\bibinfo{person}{Ferdous Wahid}, \bibinfo{person}{Rezaul~K
  Begg}, \bibinfo{person}{Chris~J Hass}, \bibinfo{person}{Saman Halgamuge},
  {and} \bibinfo{person}{David~C Ackland}.} \bibinfo{year}{2015}\natexlab{}.
\newblock \showarticletitle{Classification of Parkinson's disease gait using
  spatial-temporal gait features}.
\newblock \bibinfo{journal}{\emph{IEEE Journal of Biomedical and Health
  Informatics}} \bibinfo{volume}{19}, \bibinfo{number}{6}
  (\bibinfo{year}{2015}), \bibinfo{pages}{1794--1802}.
\newblock


\bibitem[\protect\citeauthoryear{Wilhelm, V{\"o}gele, Zsoldos, Licka,
  Kr{\"u}ger, and Bernard}{Wilhelm et~al\mbox{.}}{2015}]%
        {wilhelm2015furyexplorer}
\bibfield{author}{\bibinfo{person}{Nils Wilhelm}, \bibinfo{person}{Anna
  V{\"o}gele}, \bibinfo{person}{Rebeka Zsoldos}, \bibinfo{person}{Theresia
  Licka}, \bibinfo{person}{Bj{\"o}rn Kr{\"u}ger}, {and}
  \bibinfo{person}{J{\"u}rgen Bernard}.} \bibinfo{year}{2015}\natexlab{}.
\newblock \showarticletitle{Furyexplorer: {Visual}-interactive exploration of
  horse motion capture data}. In \bibinfo{booktitle}{\emph{Visualization and
  Data Analysis 2015}}. International Society for Optics and Photonics,
  \bibinfo{pages}{93970F}.
\newblock
\urldef\tempurl%
\url{https://doi.org/10.1117/12.2080001}
\showDOI{\tempurl}


\bibitem[\protect\citeauthoryear{Wolf, Loose, Schablowski, D{\"o}derlein, Rupp,
  Gerner, Bretthauer, and Mikut}{Wolf et~al\mbox{.}}{2006}]%
        {wolf_automated_2006}
\bibfield{author}{\bibinfo{person}{Sebastian Wolf}, \bibinfo{person}{Tobias
  Loose}, \bibinfo{person}{Matthias Schablowski}, \bibinfo{person}{Leonhard
  D{\"o}derlein}, \bibinfo{person}{R{\"u}diger Rupp},
  \bibinfo{person}{Hans~J{\"u}rgen Gerner}, \bibinfo{person}{Georg Bretthauer},
  {and} \bibinfo{person}{Ralf Mikut}.} \bibinfo{year}{2006}\natexlab{}.
\newblock \showarticletitle{Automated feature assessment in instrumented gait
  analysis}.
\newblock \bibinfo{journal}{\emph{Gait {\&} Posture}} \bibinfo{volume}{23},
  \bibinfo{number}{3} (\bibinfo{year}{2006}), \bibinfo{pages}{331--338}.
\newblock
\showISSN{09666362}
\urldef\tempurl%
\url{https://doi.org/10.1016/j.gaitpost.2005.04.004}
\showDOI{\tempurl}


\bibitem[\protect\citeauthoryear{Zintgraf, Cohen, Adel, and Welling}{Zintgraf
  et~al\mbox{.}}{2017}]%
        {zintgraf2017visualizing}
\bibfield{author}{\bibinfo{person}{Luisa~M Zintgraf}, \bibinfo{person}{Taco~S
  Cohen}, \bibinfo{person}{Tameem Adel}, {and} \bibinfo{person}{Max Welling}.}
  \bibinfo{year}{2017}\natexlab{}.
\newblock \showarticletitle{Visualizing deep neural network decisions:
  Prediction difference analysis}.
\newblock \bibinfo{journal}{\emph{arXiv:1702.04595 [Preprint].}}
  (\bibinfo{year}{2017}).
\newblock
\newblock
\shownote{Available at: http://arxiv.org/abs/1702.04595.}


\bibitem[\protect\citeauthoryear{Zurada, Malinowski, and Cloete}{Zurada
  et~al\mbox{.}}{1994}]%
        {zurada1994sensitivity}
\bibfield{author}{\bibinfo{person}{Jacek~M. Zurada},
  \bibinfo{person}{Aleksander Malinowski}, {and} \bibinfo{person}{Ian Cloete}.}
  \bibinfo{year}{1994}\natexlab{}.
\newblock \showarticletitle{Sensitivity analysis for minimization of input data
  dimension for feedforward neural network}. In
  \bibinfo{booktitle}{\emph{Proceedings of IEEE International Symposium on
  Circuits and Systems (ISCAS)}}. IEEE, \bibinfo{pages}{447--450}.
\newblock
\urldef\tempurl%
\url{https://doi.org/10.1109/ISCAS.1994.409622}
\showDOI{\tempurl}


\end{thebibliography}

\appendix
\newpage
\section*{Supplementary Tables and Figures}

\newcommand{\hbAppendixPrefix}{S}

\renewcommand{\thefigure}{\hbAppendixPrefix\arabic{figure}}
\setcounter{figure}{0}
\renewcommand{\thetable}{\hbAppendixPrefix\arabic{table}} 
\setcounter{table}{0}

The present Supplementary Material is intended to present additional results we generated for the paper \textbf{"On the Explanation of Machine Learning Predictions in Clinical Gait Analysis"}. The primary aim of this article is to investigate to which degree Explainable Artificial Intelligence (XAI) methods can make predictions of machine learning (ML) approaches more explainable to clinical experts in clinical gait analysis. 

For this purpose, we investigate different gait classification tasks, employ a representative set of classification methods~--~(linear)~Support Vector Machine~(SVM), Multi-layer Perceptron~(MLP), and Convolutional Neural Network~(CNN)~--, and a well-established XAI method – Layer-wise Relevance Propagation (LRP) – to explain predictions at the signal (input) level. Since there is no ground truth for automatically generated explanations in this context, we evaluate the explanations from a clinical point of view by a clinical expert. In addition, as a second reference, we propose the use of Statistical Parametric Mapping (SPM) to compare the obtained results from a statistical point of view. 

The dataset employed, comprises ground reaction force~(GRF) measurements from 132 patients with gait disorders~($GD$) and data from 62 healthy controls~($HC$). The $GD$ class is furthermore differentiated into three classes of gait disorders associated with the hip~($H$), knee~($K$), and ankle~($A$). The classification tasks, which represent the basis of the XAI investigation, due to high classification accuracies obtained, include a binary classification between healthy controls and all gait disorders~($HC/GD$), and a binary classification between healthy controls and each gait disorder separately,~\ie, $HC/H$, $HC/K$, and $HC/A$. The classification results obtained for the six classification tasks, are presented in Supplementary Table~\ref{table:classification-results}.

The following figures visualize the relevance-based explanations obtained with LRP. The input vector for the classifiers comprises concatenated affected and unaffected GRF signals. These GRF signals are time-normalized to 101 points (100\% stance phase), thus the input vector contains 606 values. For each value LRP provides whether they are relevant or not for the classification. Sub-figure~(A) shows mean GRF signals averaged over each class of the classification task. The shaded areas in all sub-figures highlight areas in the input signals where SPM resulted in a statistically significant difference between both classes. Sub-figure~(B) shows mean GRF signals (including a band of one standard deviation) for the $HC$ class. The input relevance indicates which GRF characteristics were most relevant for (or contradictory to) the classification of a certain class. For visualization, input values neutral to the prediction~($R_i \approx 0$) are shown in black, while warm hues indicate input values supporting the prediction~($R_i \gg 0 $) of the analyzed class and cool hues identify contradictory input values~($R_i \ll 0 $). Sub-figure~(C) depicts mean GRF signals averaged over a pathological class ($H$, $K$, or $A$) or all gait disorders ($GD$), in the same format as in sub-figure~(B). Sub-figure~(D) shows the effect size obtained from SPM and the total relevance, which is calculated as the sum of the absolute input relevance values of both classes. The total relevance indicates the common relevance of the input signal for the classification task. 

\newpage
\section*{Classification Results}

\begin{table*}[ht!]
\centering
\caption{Overview of the prediction accuracy obtained for the three employed classification methods (CNN, SVM and MLP) and all six classification tasks with min-max normalized and non-normalized input signals, reported in pairs of mean (standard deviation) over the ten-fold cross validation in percent. Note that the Zero-Rule Baseline (ZRB) is task-specific.}
\label{table:classification-results}
\begin{tabular}{lccccc}
\hline
Task        & Normalization     & ZRB & SVM & MLP & CNN \\
\hline
HC/GD        & no norm.   & 68.0  & 88.6 (4.9) & 88.1 (4.8) & 87.8 (4.5) \\
HC/GD 	    & min-max        & 68.0  & 88.4 (5.3) & 88.8 (5.0) & 88.0 (5.0) \\
\hline
HC/H         & no norm.   & 62.6  & 85.9 (8.4) & 86.6 (7.9) & 85.1 (8.2) \\
HC/H 	    & min-max        & 62.6  & 87.1 (7.6) & 86.7 (8.5) & 85.5 (8.0) \\
\hline
HC/K         & no norm.   & 54.4  & 85.7 (9.0) & 86.1 (7.9) & 84.8 (9.9) \\
HC/K 	    & min-max        & 54.4  & 88.5 (7.2) & 88.5 (7.6) & 85.9 (9.3) \\
\hline
HC/A         & no norm.   & 59.0  & 89.1 (5.9) & 88.3 (6.3) & 88.7 (5.5) \\
HC/A 	    & min-max        & 59.0  & 87.6 (7.4) & 86.5 (8.1) & 86.7 (8.3) \\
\hline
H/K/A       & no norm.   & 39.4  & 46.4 (9.5) & 45.9 (11.0) & 48.0 (10.1) \\
H/K/A  	    & min-max        & 39.4  & 51.8 (9.6) & 47.4 (10.9) & 50.7 (9.8) \\
\hline
HC/H/K/A     & no norm.   & 32.0  & 58.7 (7.5) & 55.6 (7.6) & 55.0 (8.7) \\
HC/H/K/A 	& min-max        & 32.0  & 59.5 (8.5) & 59.2 (7.6) & 57.5 (7.0) \\
\hline
\end{tabular}
\end{table*}

\newpage

\section*{Explainability Results}
\subsection*{Classification Task: $HC/GD$ | Classification method: $CNN$}

\begin{figure}[ht!]
  \centering
	\includegraphics[width=0.85\linewidth]{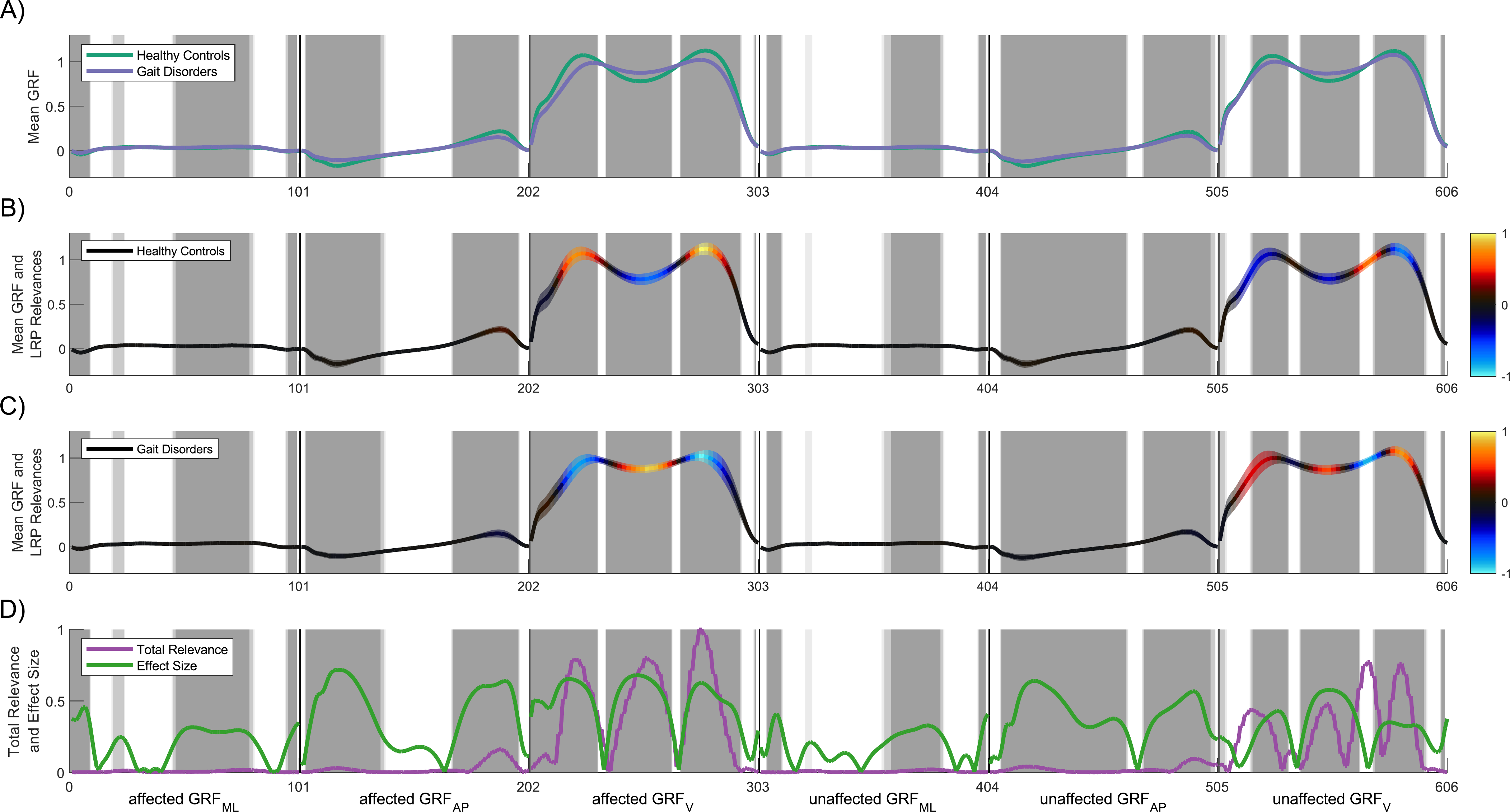} 
    \caption{Result overview for the classification of healthy controls and the aggregated class of all three gait disorders~($HC/GD$) based on non-normalized GRF signals using a CNN as classifier.}
    \label{img:supl_cnn-nonorm-NGD}
\end{figure}

\begin{figure}[ht!]
  \centering
	\includegraphics[width=0.85\linewidth]{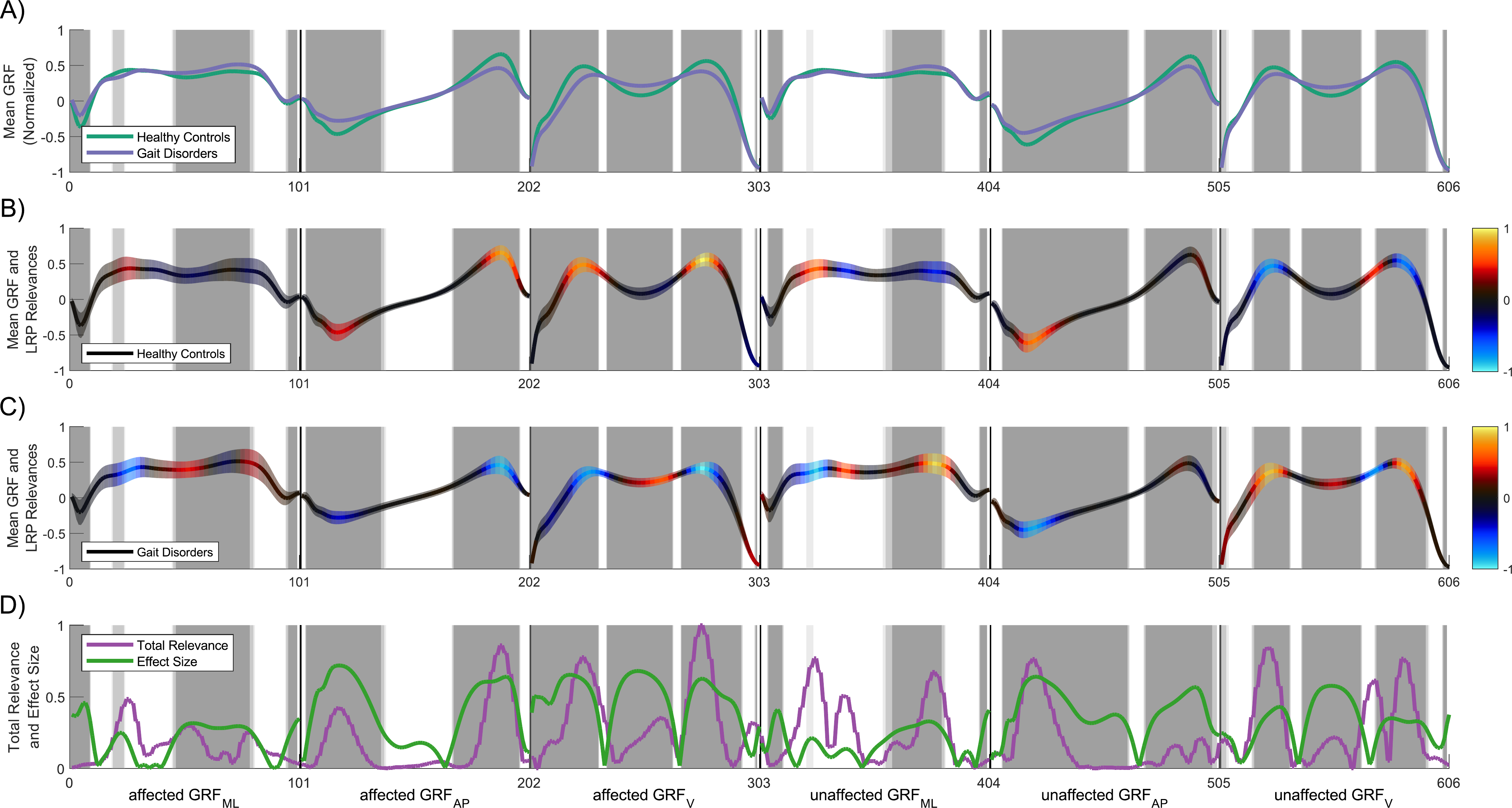} 
    \caption{Result overview for the classification of healthy controls and the aggregated class of all three gait disorders~($HC/GD$) based on min-max normalized GRF signals using a CNN as classifier.}
    \label{img:supl_cnn-norm-NGD}
\end{figure}

\newpage
\subsection*{Classification Task: $HC/GD$ | Classification method: $MLP$}

\begin{figure}[ht!]
  \centering
	\includegraphics[width=0.85\linewidth]{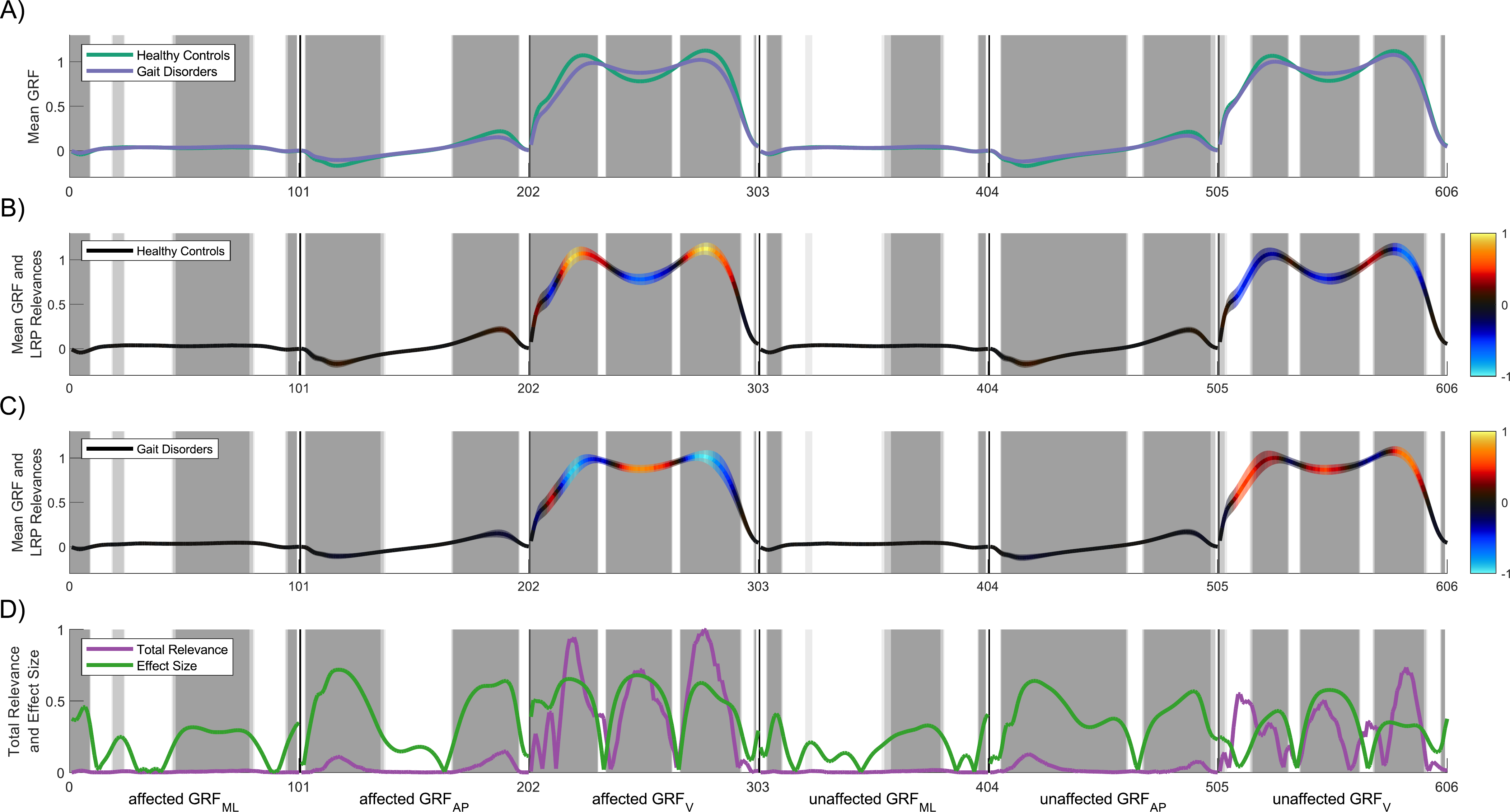} 
    \caption{Result overview for the classification of healthy controls and the aggregated class of all three gait disorders~($HC/GD$) based on non-normalized GRF signals using an MLP as classifier.}
    \label{img:supl_mlp-nonorm-NGD}
\end{figure}

\begin{figure}[ht!]
  \centering
	\includegraphics[width=0.85\linewidth]{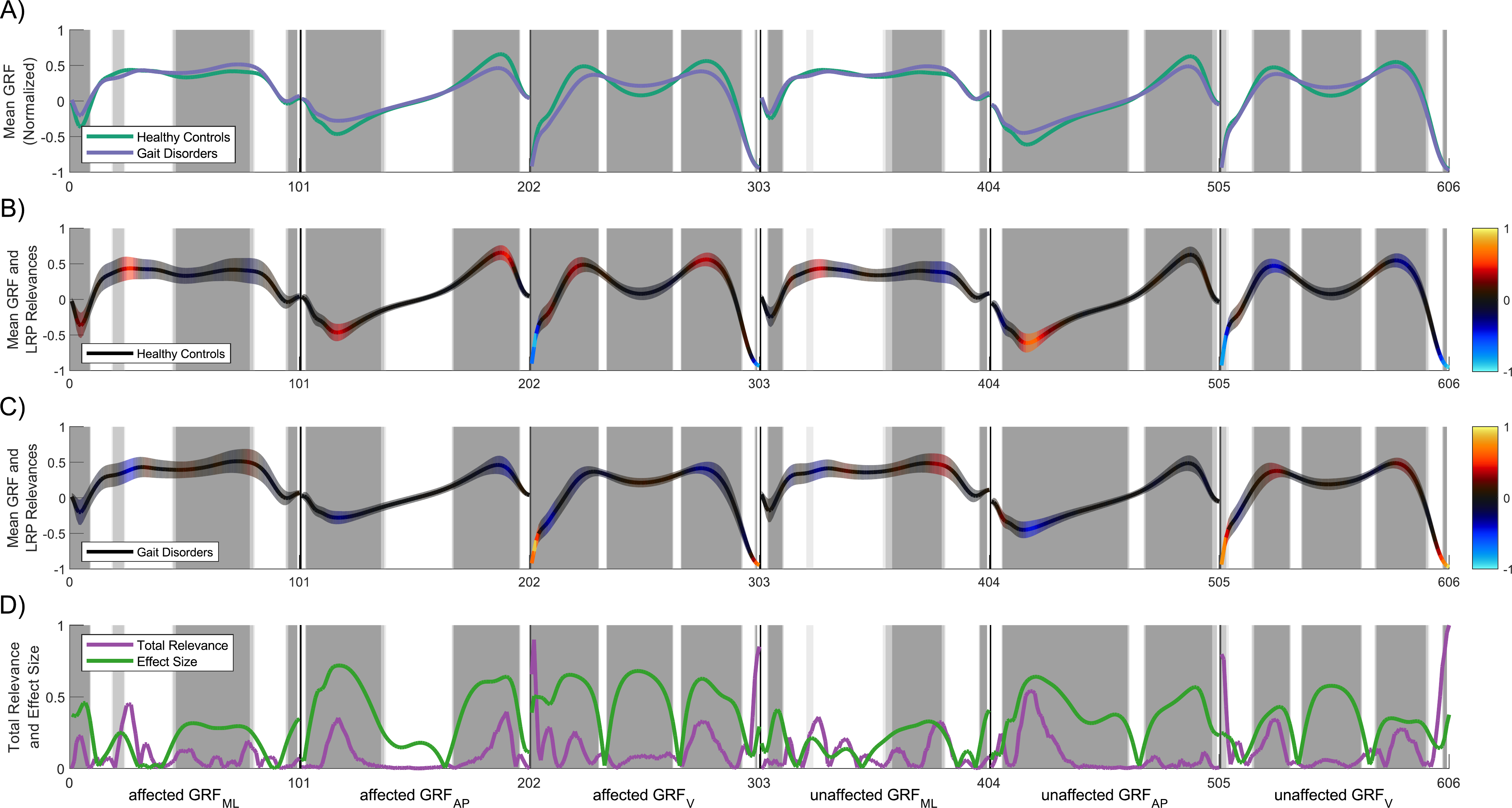} 
    \caption{Result overview for the classification of healthy controls and the aggregated class of all three gait disorders~($HC/GD$) based on min-max normalized GRF signals using an MLP as classifier.}
    \label{img:supl_mlp-norm-NGD}
\end{figure}

\newpage
\subsection*{Classification Task: $HC/GD$ | Classification method: $SVM$}

\begin{figure}[ht!]
  \centering
	\includegraphics[width=0.85\linewidth]{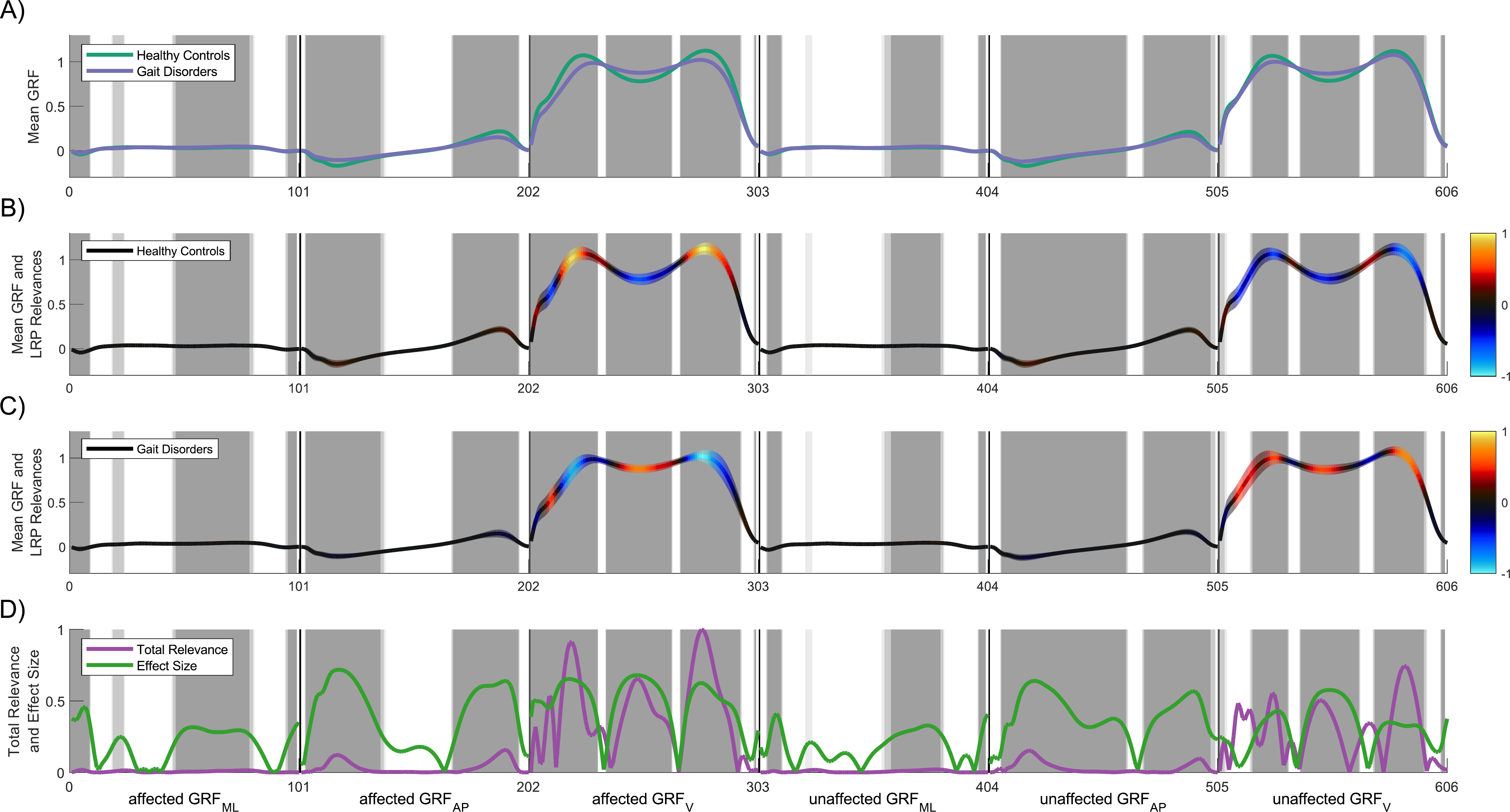} 
    \caption{Result overview for the classification of healthy controls and the aggregated class of all three gait disorders~($HC/GD$) based on non-normalized GRF signals using a SVM as classifier.}
    \label{img:supl_svm-nonorm-NGD}
\end{figure}

\begin{figure}[ht!]
  \centering
	\includegraphics[width=0.85\linewidth]{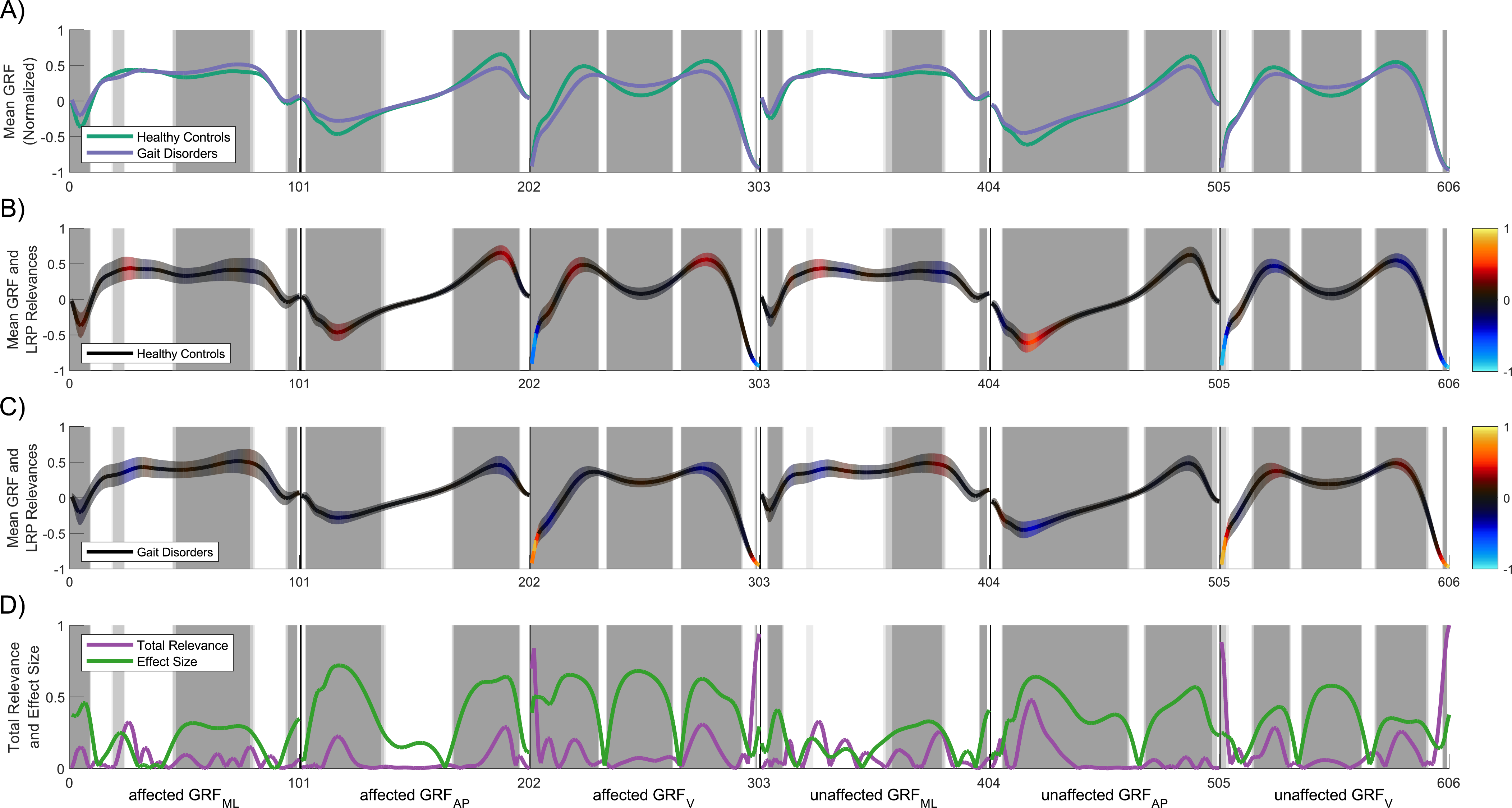} 
    \caption{Result overview for the classification of healthy controls and the aggregated class of all three gait disorders~($HC/GD$) based on min-max normalized GRF signals using a SVM as classifier.}
    \label{img:supl_svm-norm-NGD}
\end{figure}

\newpage
\subsection*{Classification Task: $HC/H$ | Classification method: $CNN$}
\begin{figure}[ht!]
  \centering
	\includegraphics[width=0.85\linewidth]{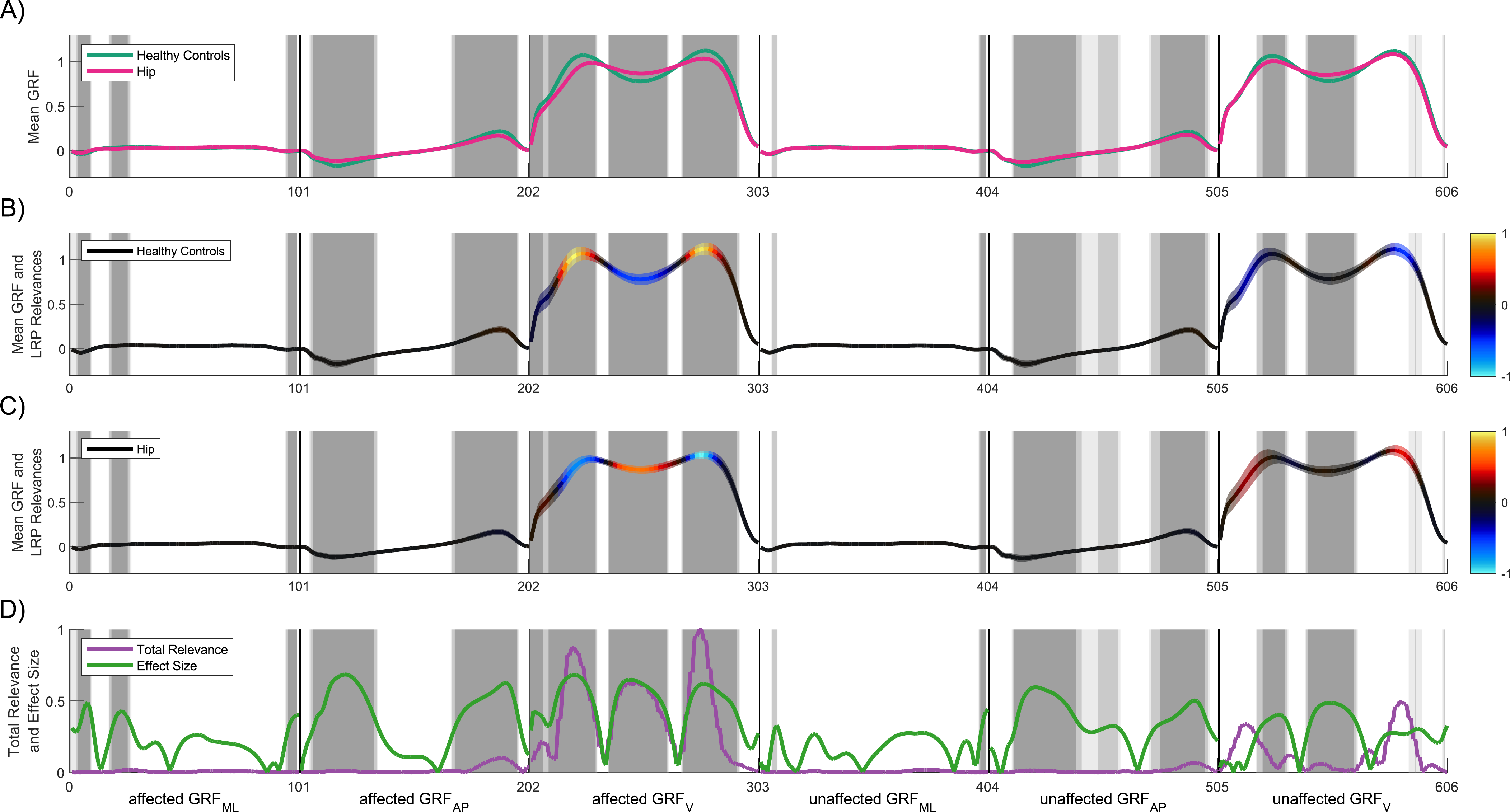} 
    \caption{Result overview for the classification of healthy controls ($HC$) and hip injury class ($H$) based on non-normalized GRF signals using a CNN as classifier.}
    \label{img:supl_cnn-nonorm-NH}
\end{figure}

\begin{figure}[ht!]
  \centering
	\includegraphics[width=0.85\linewidth]{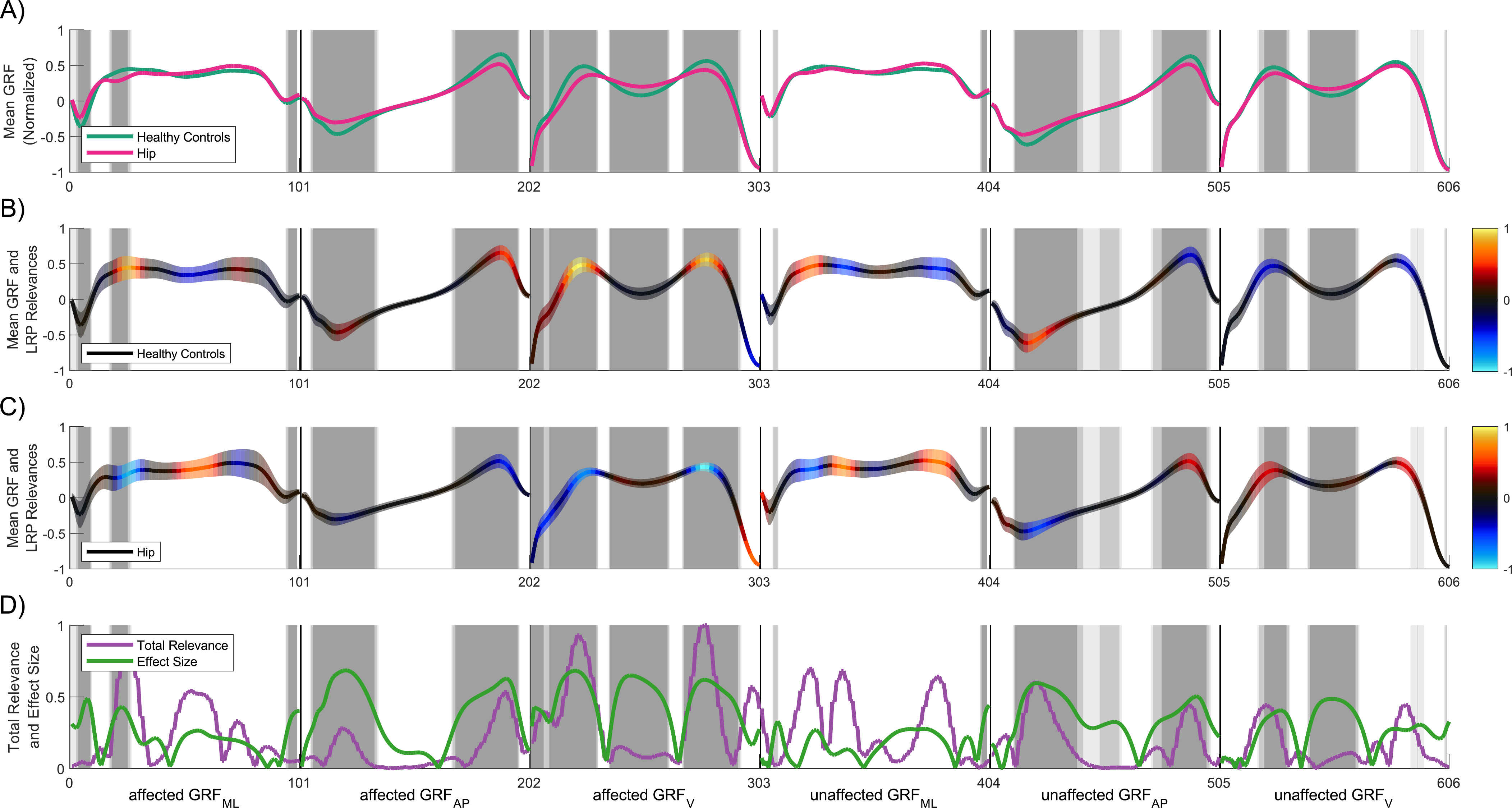} 
    \caption{Result overview for the classification of healthy controls ($HC$) and hip injury class ($H$) based on min-max normalized GRF signals using a CNN as classifier.}
    \label{img:supl_cnn-norm-NH}
\end{figure}

\newpage
\subsection*{Classification Task: $HC/H$ | Classification method: $MLP$}

\begin{figure}[ht!]
  \centering
	\includegraphics[width=0.85\linewidth]{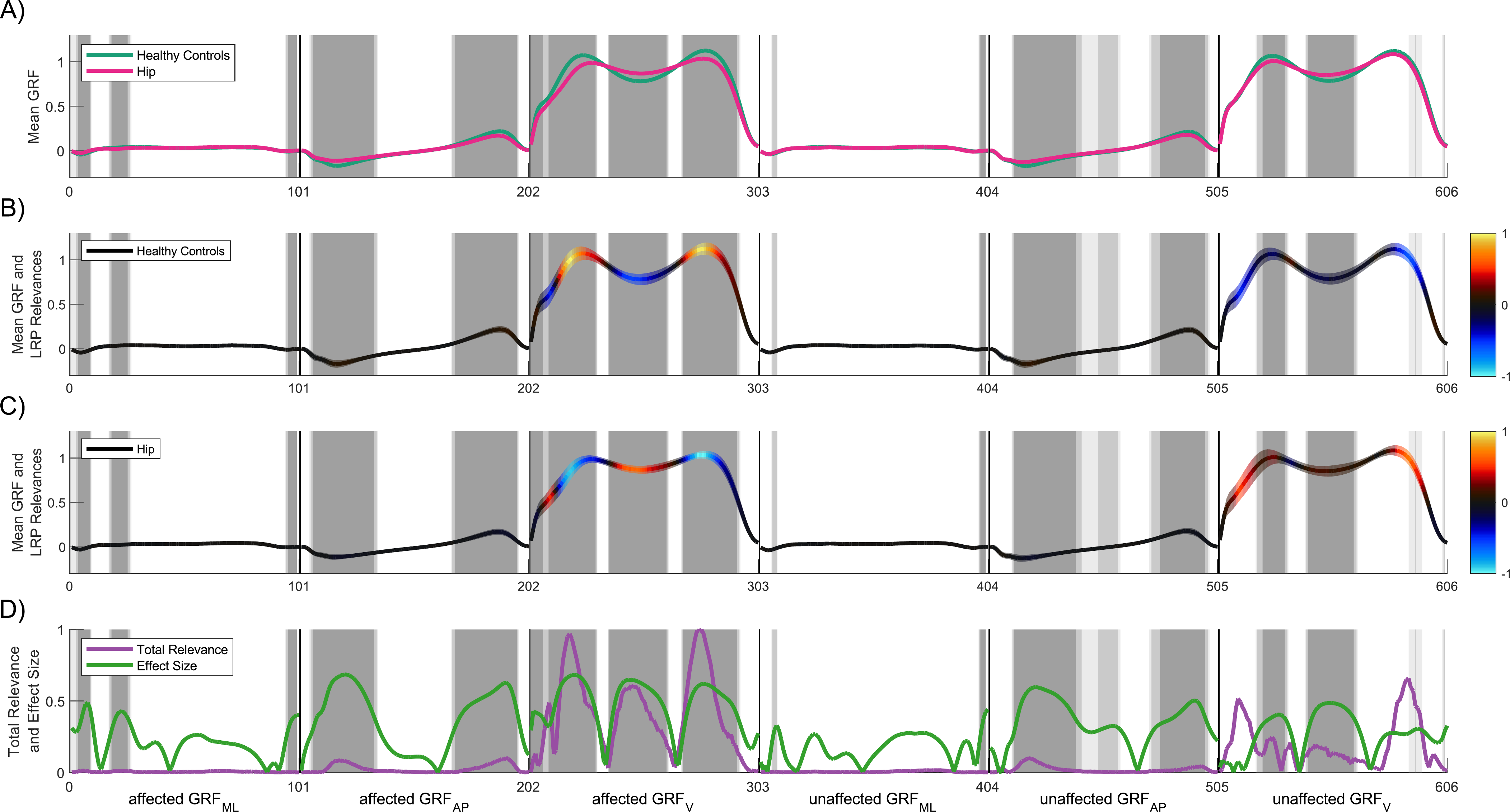} 
    \caption{Result overview for the classification of healthy controls ($HC$) and hip injury class ($H$) based on non-normalized GRF signals using an MLP as classifier.}
    \label{img:supl_mlp-nonorm-NH}
\end{figure}

\begin{figure}[ht!]
  \centering
	\includegraphics[width=0.85\linewidth]{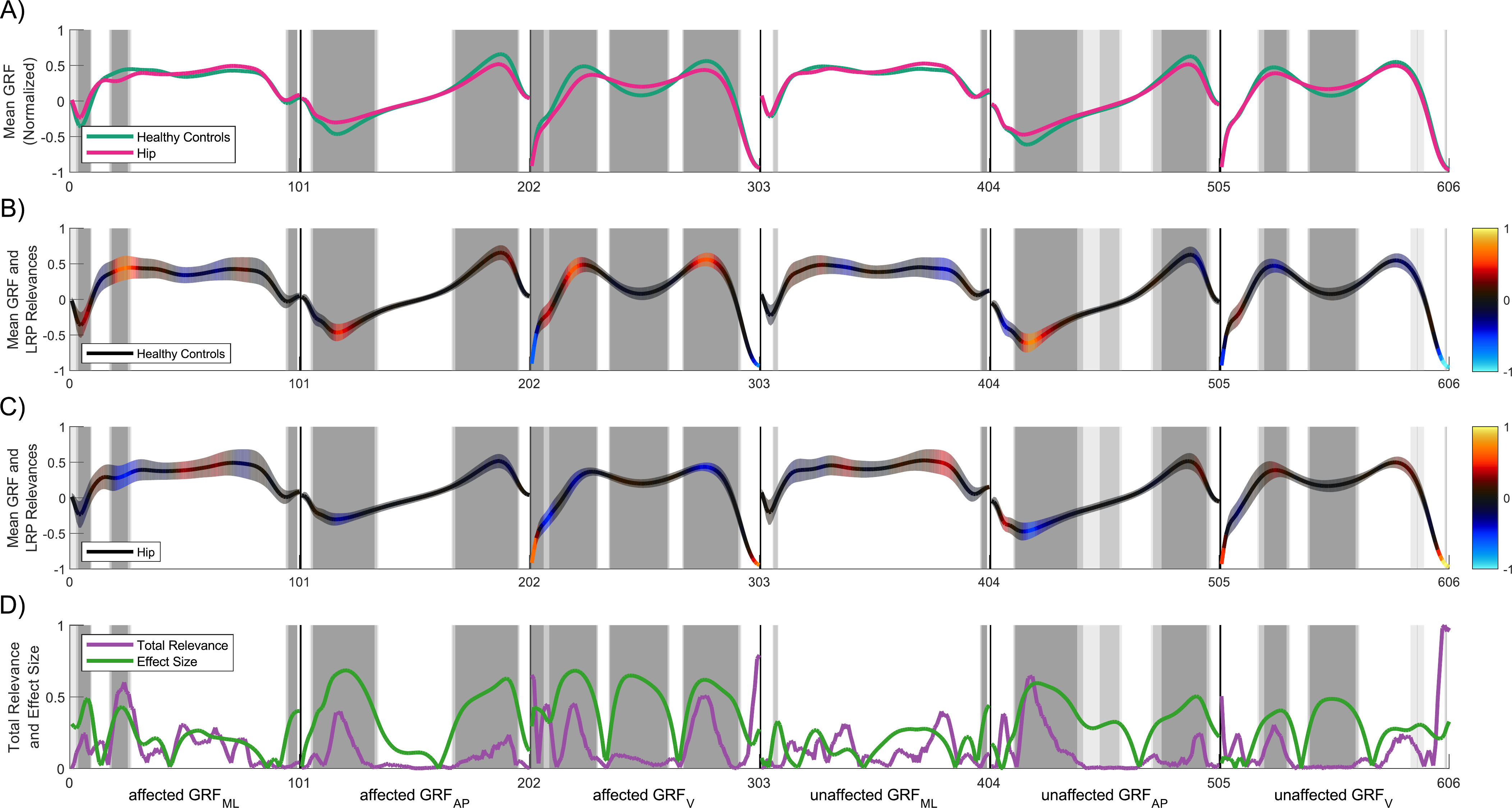} 
    \caption{Result overview for the classification of healthy controls ($HC$) and hip injury class ($H$) based on min-max normalized GRF signals using an MLP as classifier.}
    \label{img:supl_mlp-norm-NH}
\end{figure}

\newpage
\subsection*{Classification Task: $HC/H$ | Classification method: $SVM$}

\begin{figure}[ht!]
  \centering
	\includegraphics[width=0.85\linewidth]{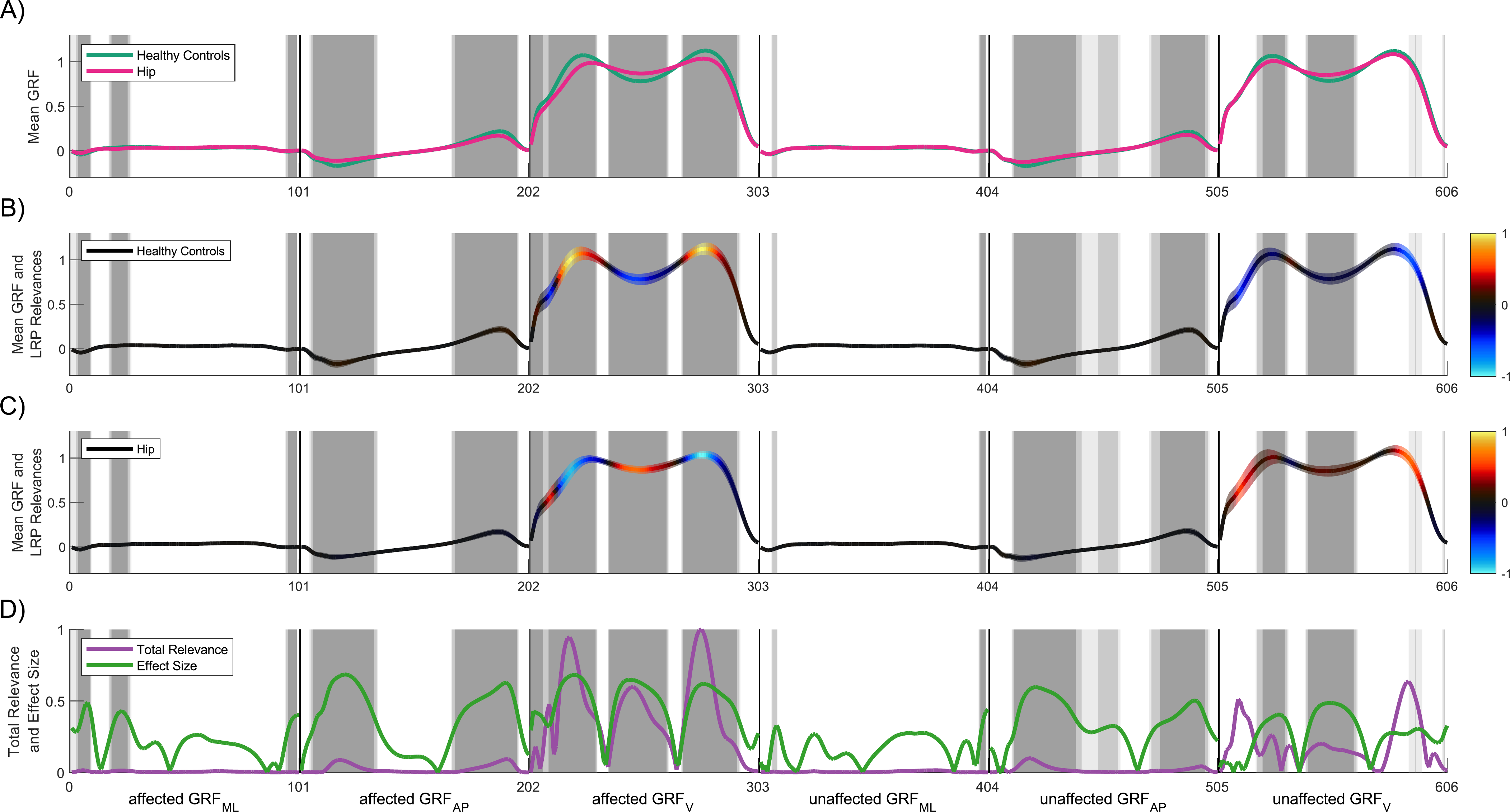} 
    \caption{Result overview for the classification of healthy controls ($HC$) and hip injury class ($H$) based on non-normalized GRF signals using a SVM as classifier.}
    \label{img:supl_svm-nonorm-NH}
\end{figure}

\begin{figure}[ht!]
  \centering
	\includegraphics[width=0.85\linewidth]{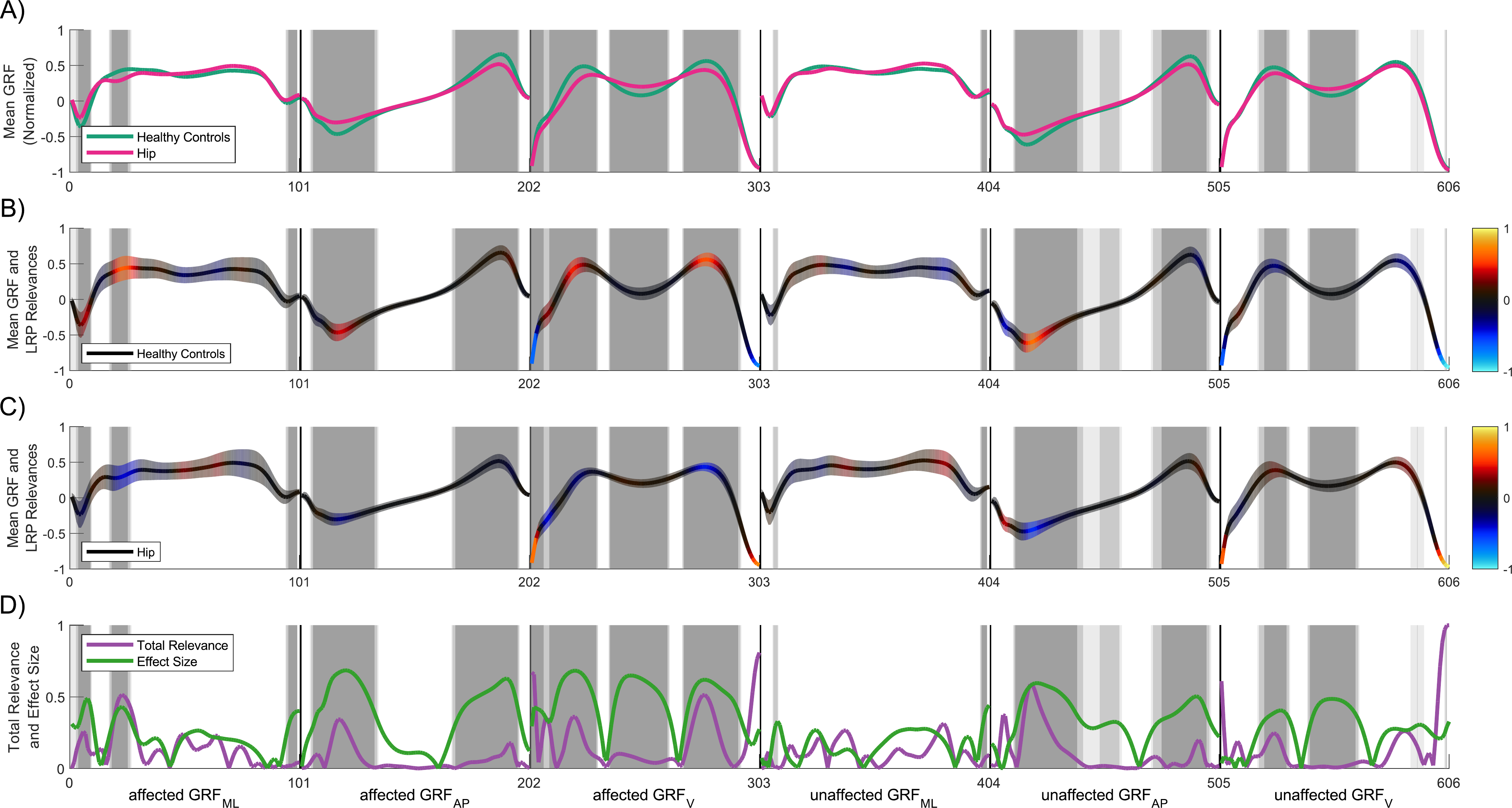} 
    \caption{Result overview for the classification of healthy controls ($HC$) and hip injury class ($H$) based on min-max normalized GRF signals using a SVM as classifier.}
    \label{img:supl_svm-norm-NH}
\end{figure}

\newpage
\subsection*{Classification Task: $HC/K$ | Classification method: $CNN$}

\begin{figure}[ht!]
  \centering
	\includegraphics[width=0.85\linewidth]{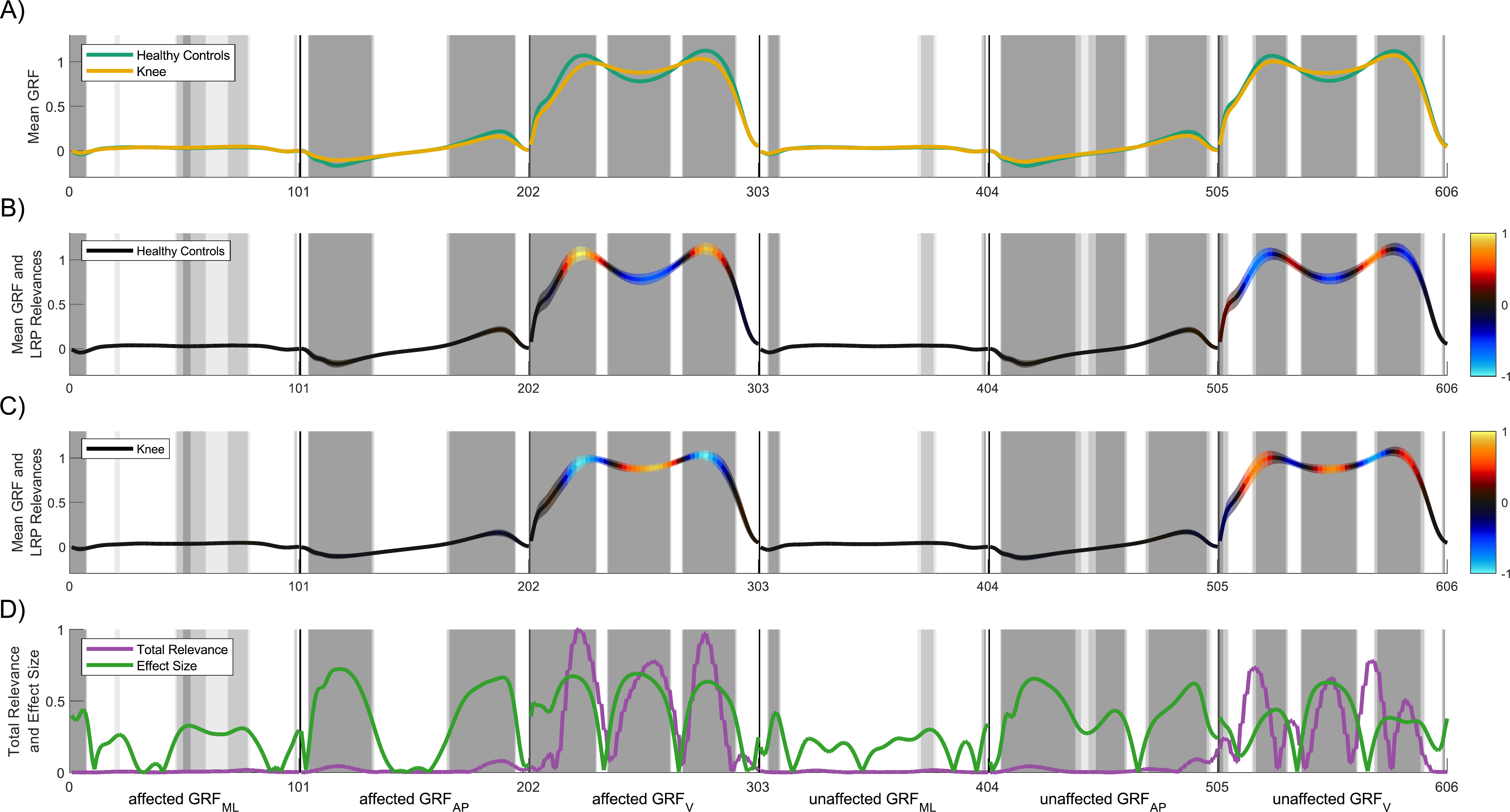} 
    \caption{Result overview for the classification of healthy controls~($HC$) and knee injury class~($K$) based on non-normalized GRF signals using a CNN as classifier.}
    \label{img:supl_cnn-nonorm-NK}
\end{figure}

\begin{figure}[ht!]
  \centering
	\includegraphics[width=0.85\linewidth]{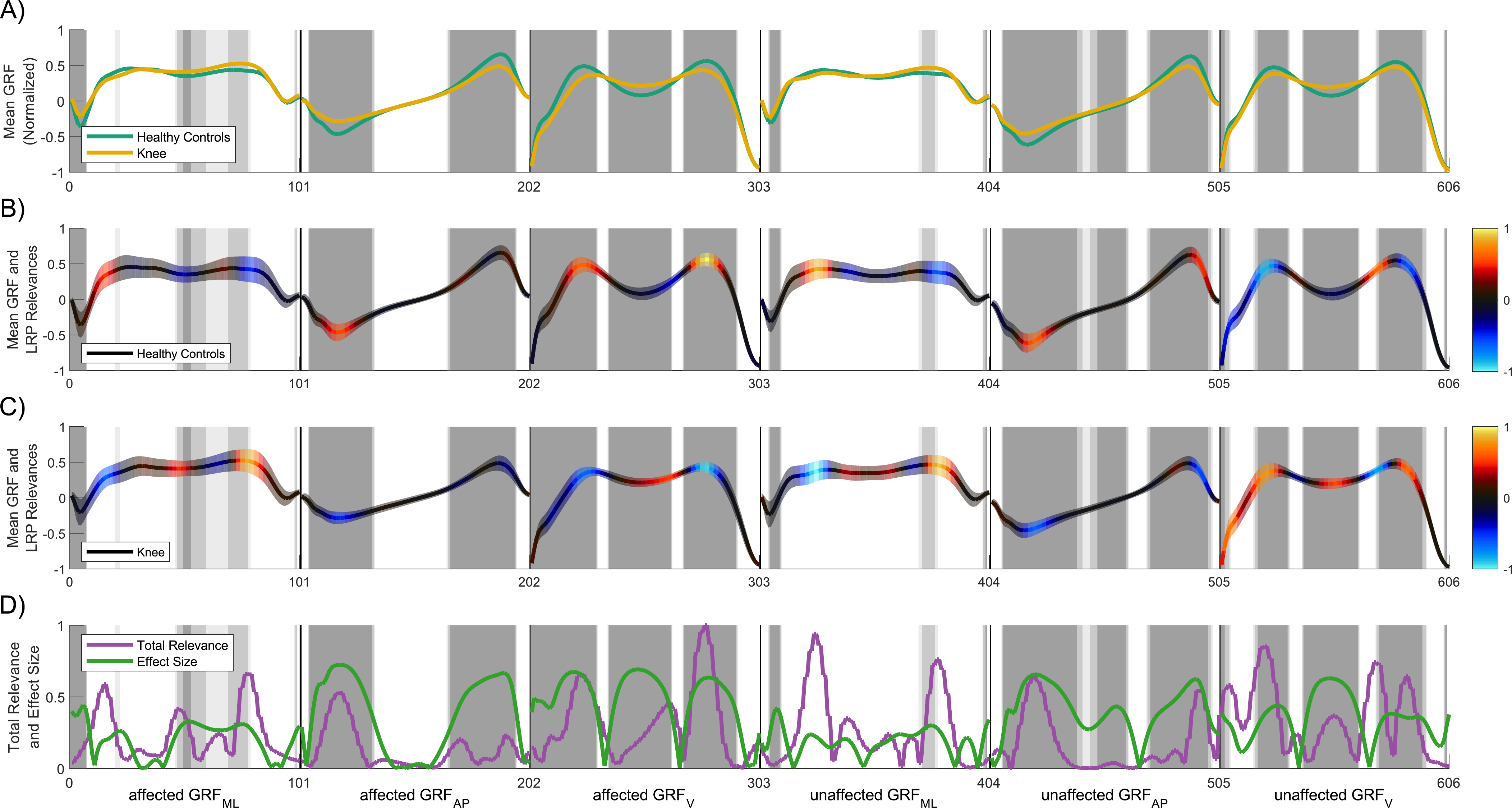} 
    \caption{Result overview for the classification of healthy controls~($HC$) and knee injury class~($K$) based on min-max normalized GRF signals using a CNN as classifier.}
    \label{img:supl_cnn-norm-NK}
\end{figure}

\newpage
\subsection*{Classification Task: $HC/K$ | Classification method: $MLP$}

\begin{figure}[ht!]
  \centering
	\includegraphics[width=0.85\linewidth]{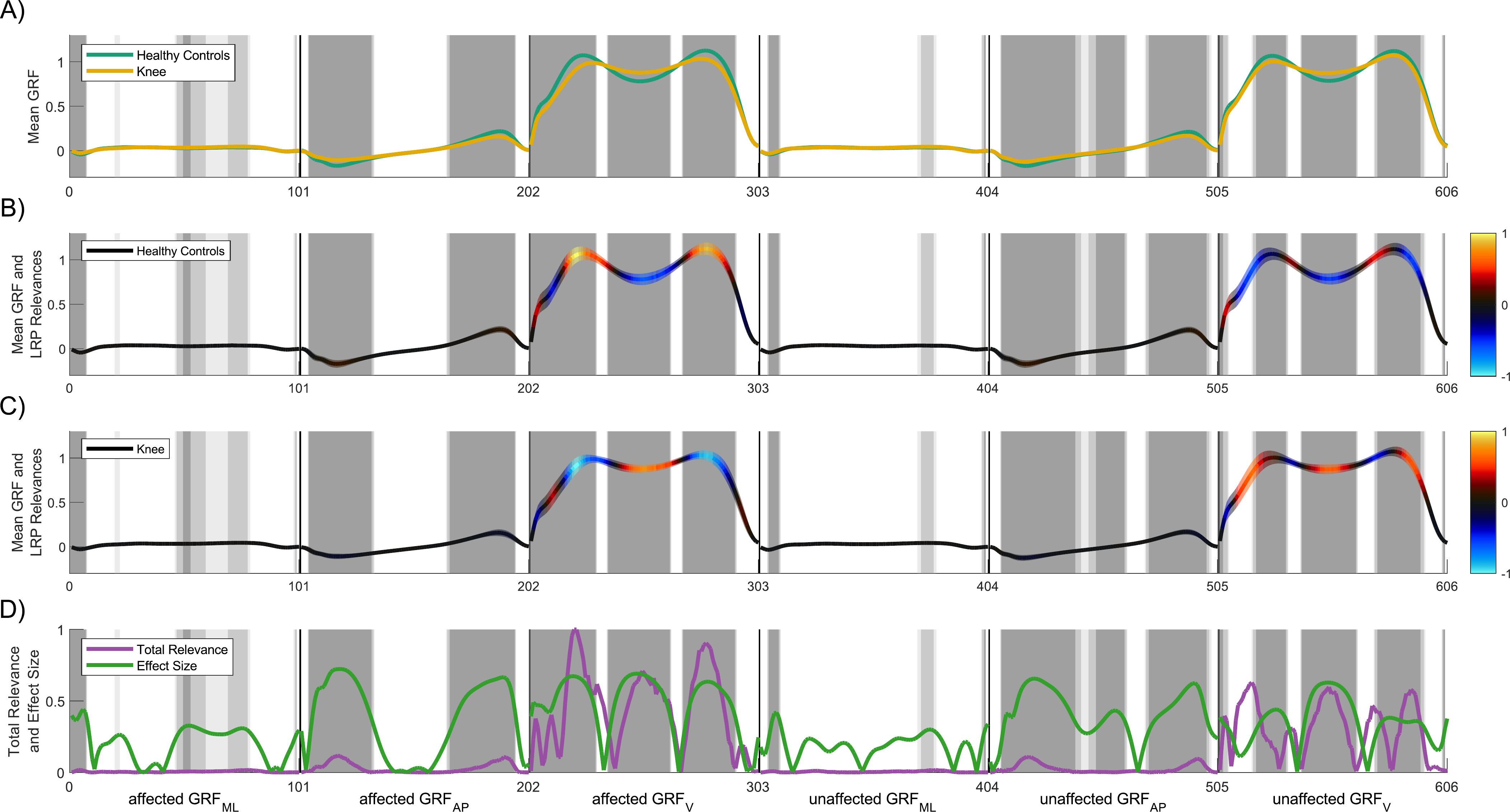} 
    \caption{Result overview for the classification of healthy controls~($HC$) and knee injury class~($K$) based on non-normalized GRF signals using an MLP as classifier.}
    \label{img:supl_mlp-nonorm-NK}
\end{figure}

\begin{figure}[ht!]
  \centering
	\includegraphics[width=0.85\linewidth]{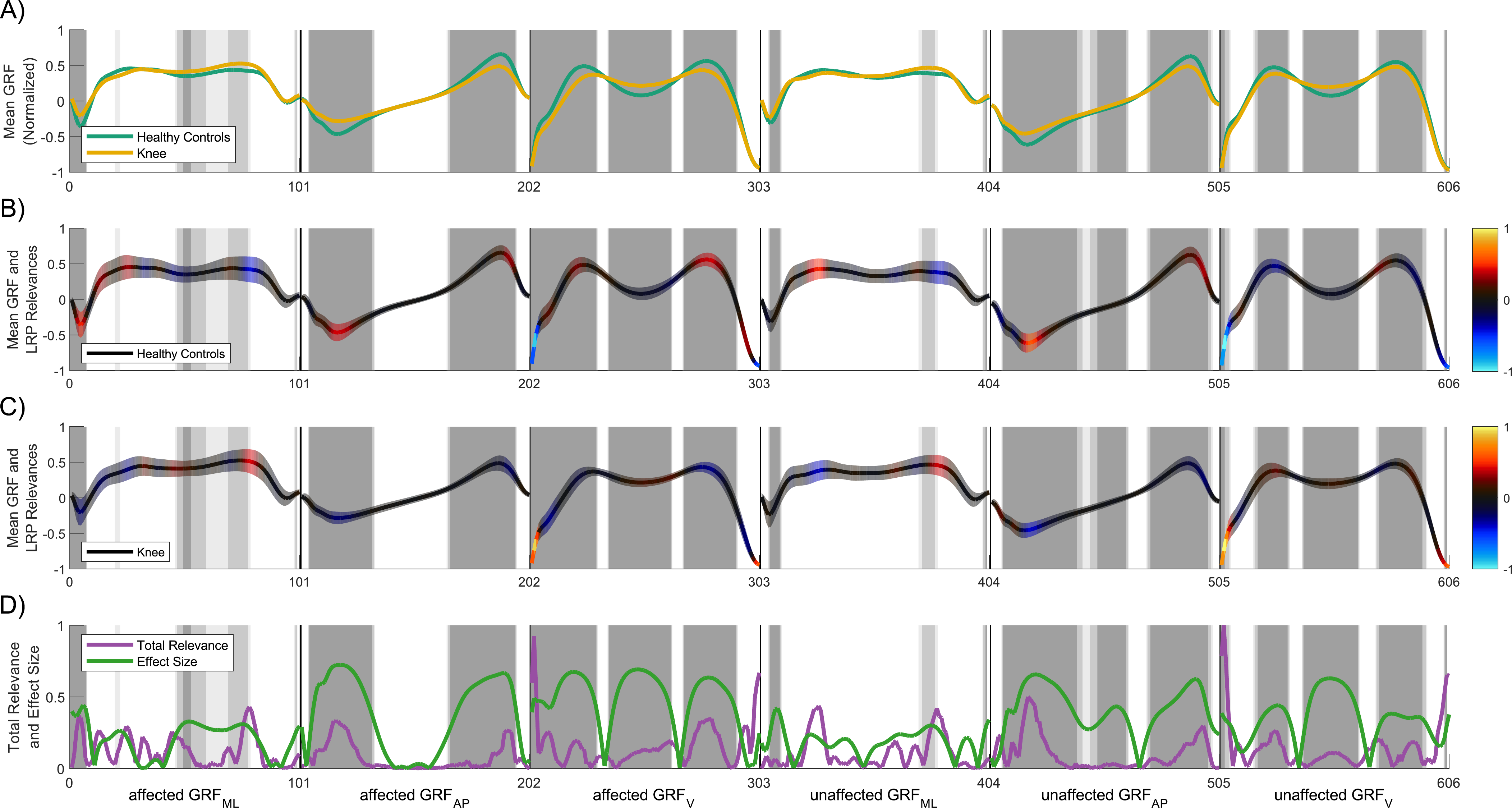} 
    \caption{Result overview for the classification of healthy controls~($HC$) and knee injury class~($K$) based on min-max normalized GRF signals using an MLP as classifier.}
    \label{img:supl_mlp-norm-NK}
\end{figure}

\newpage
\subsection*{Classification Task: $HC/K$ | Classification method: $SVM$}

\begin{figure}[ht!]
  \centering
	\includegraphics[width=0.85\linewidth]{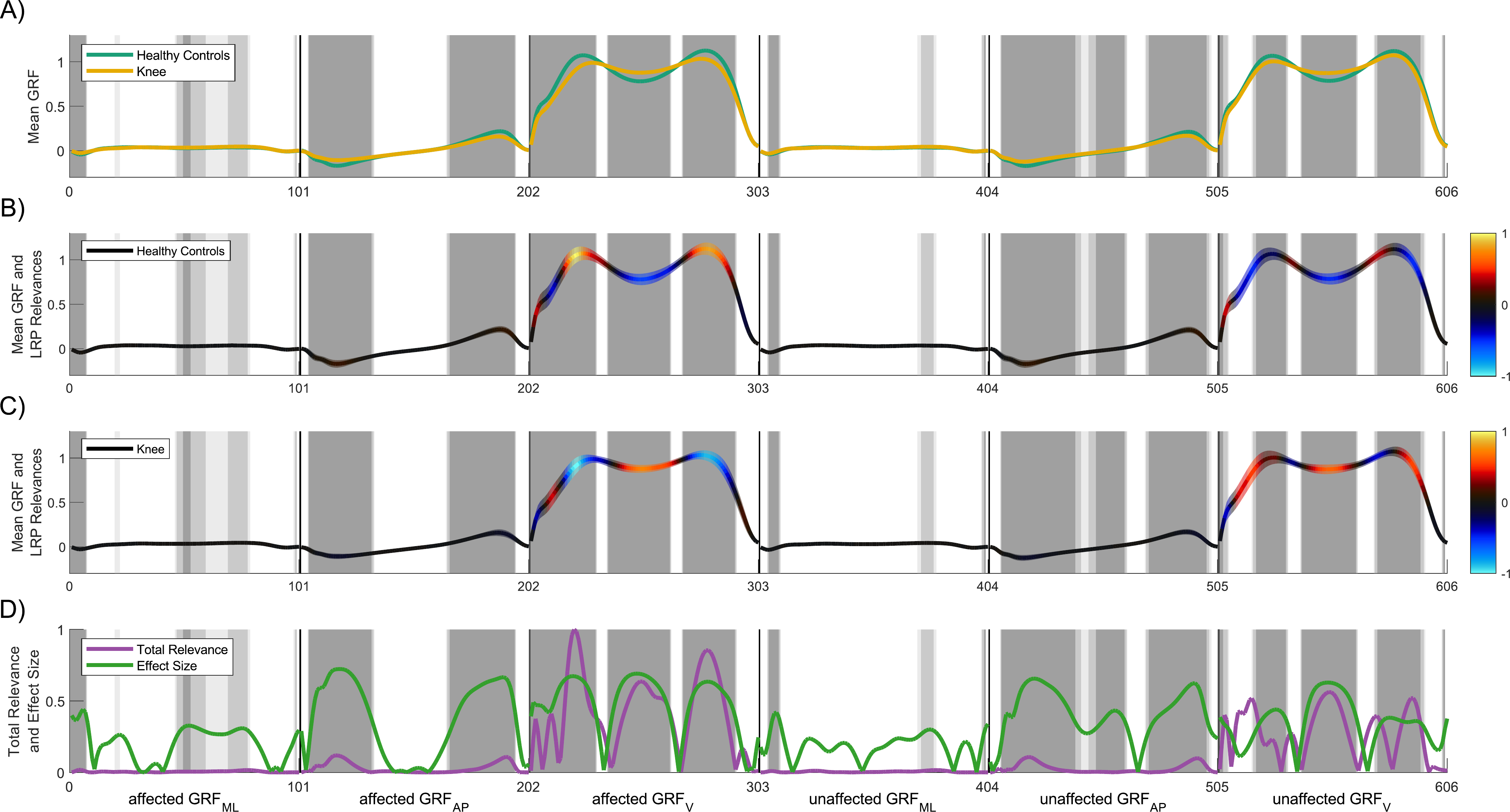} 
    \caption{Result overview for the classification of healthy controls~($HC$) and knee injury class~($K$) based on non-normalized GRF signals using a SVM as classifier.}
    \label{img:supl_svm-nonorm-NK}
\end{figure}

\begin{figure}[ht!]
  \centering
	\includegraphics[width=0.85\linewidth]{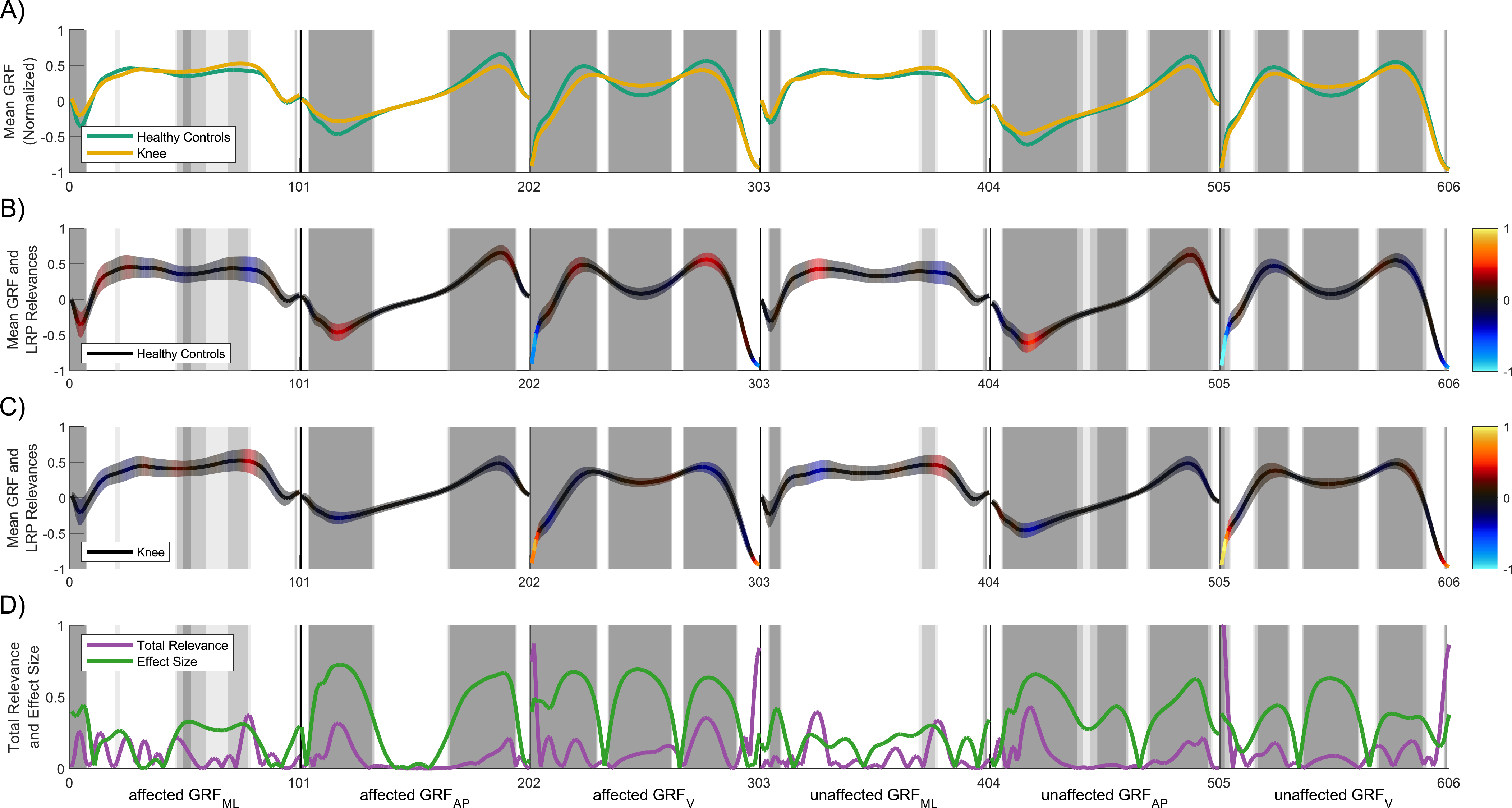} 
    \caption{Result overview for the classification of healthy controls~($HC$) and knee injury class~($K$) based on min-max normalized GRF signals using a SVM as classifier.}
    \label{img:supl_svm-norm-NK}
\end{figure}

\newpage
\subsection*{Classification Task: $HC/A$ | Classification method: $CNN$}

\begin{figure}[ht!]
  \centering
	\includegraphics[width=0.85\linewidth]{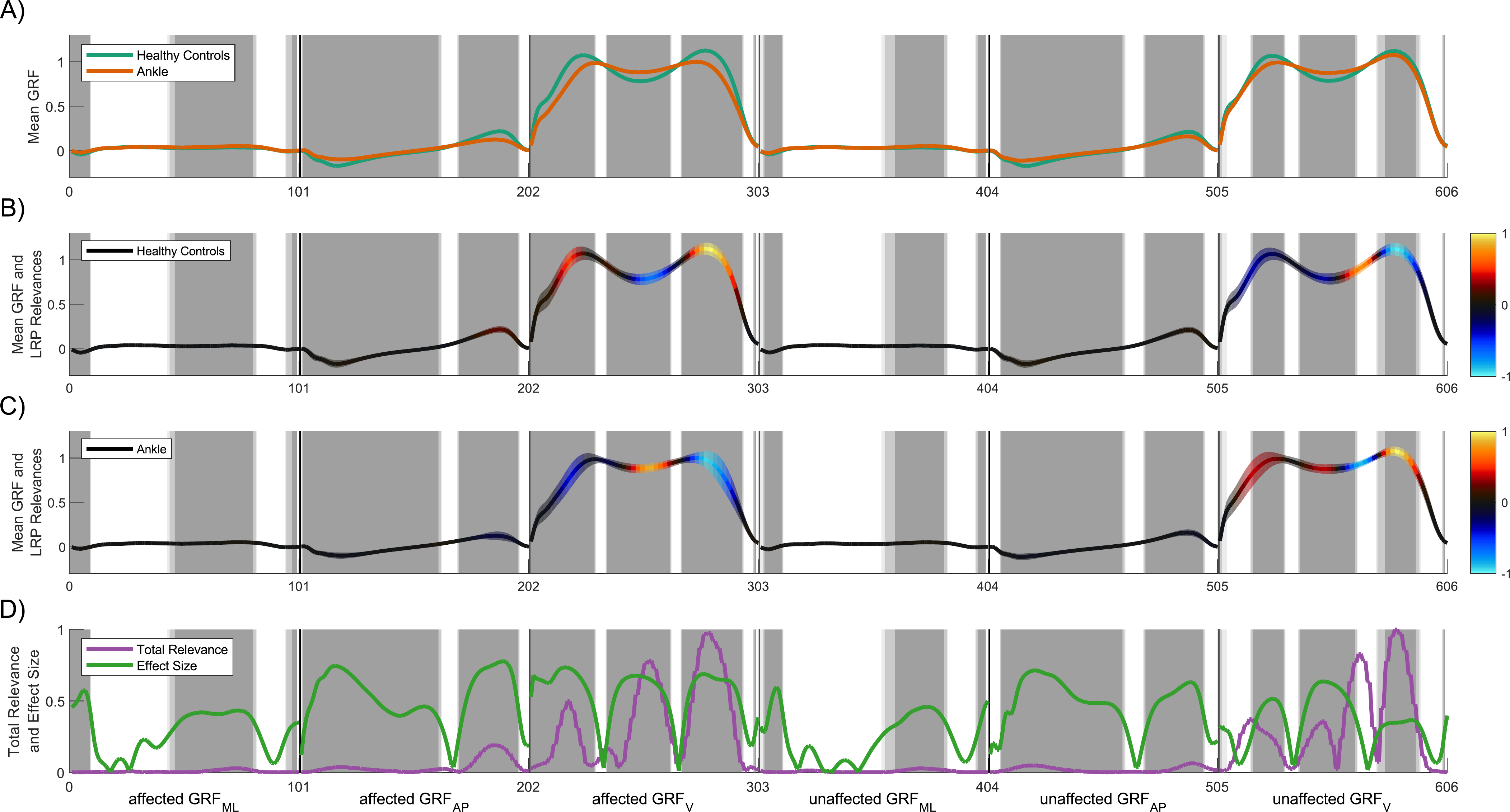} 
    \caption{Result overview for the classification of healthy controls~($HC$) and ankle injury class~($A$) based on non-normalized GRF signals using a CNN as classifier.}
    \label{img:supl_cnn-nonorm-NA}
\end{figure}

\begin{figure}[ht!]
  \centering
	\includegraphics[width=0.85\linewidth]{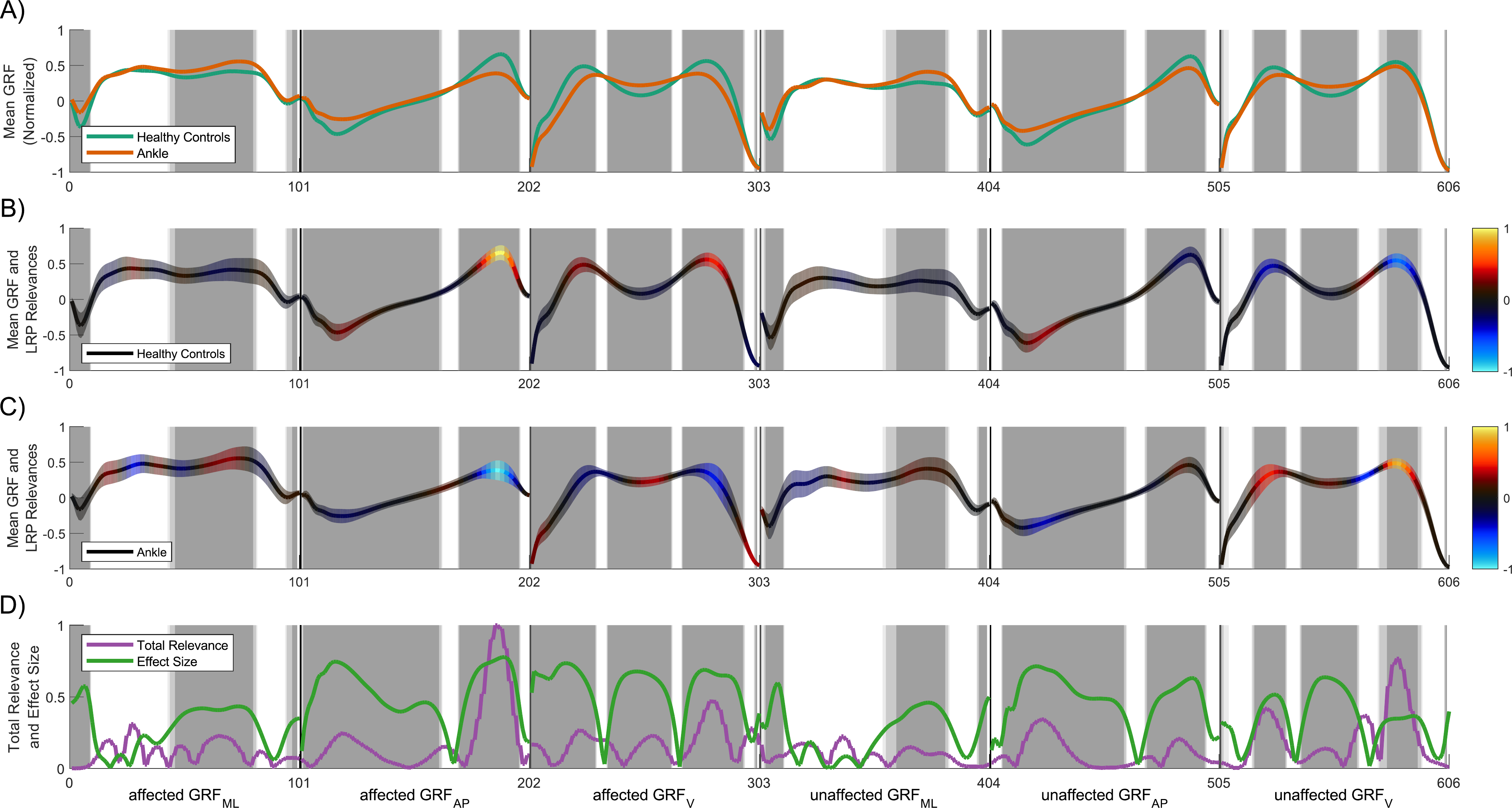} 
    \caption{Result overview for the classification of healthy controls~($HC$) and ankle injury class~($A$) based on min-max normalized GRF signals using a CNN as classifier.}
    \label{img:supl_cnn-norm-NA}
\end{figure}

\newpage
\subsection*{Classification Task: $HC/A$ | Classification method: $MLP$}

\begin{figure}[ht!]
  \centering
	\includegraphics[width=0.85\linewidth]{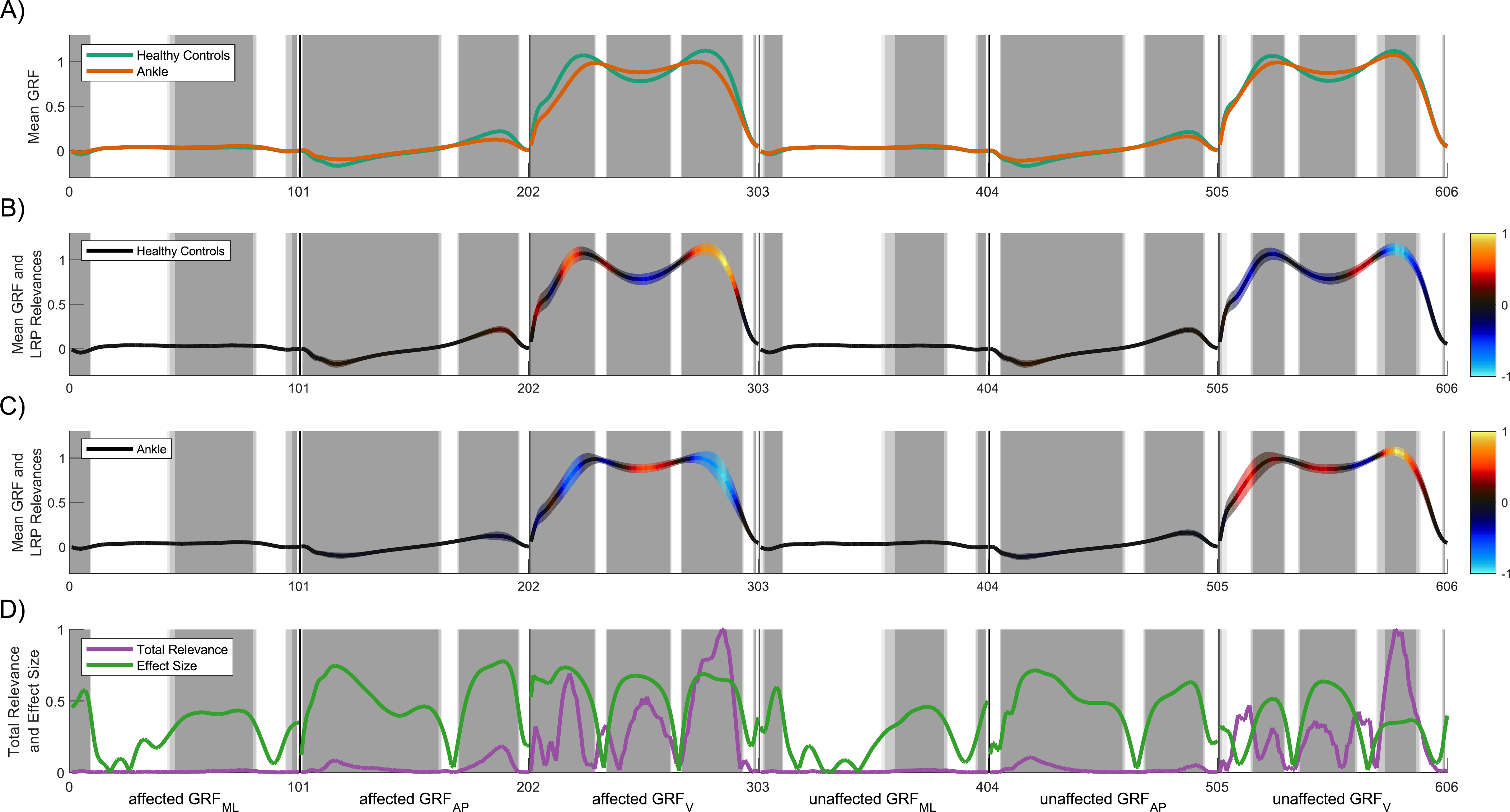} 
    \caption{Result overview for the classification of healthy controls~($HC$) and ankle injury class~($A$) based on non-normalized GRF signals using an MLP as classifier.}
    \label{img:supl_mlp-nonorm-NA}
\end{figure}

\begin{figure}[ht!]
  \centering
	\includegraphics[width=0.85\linewidth]{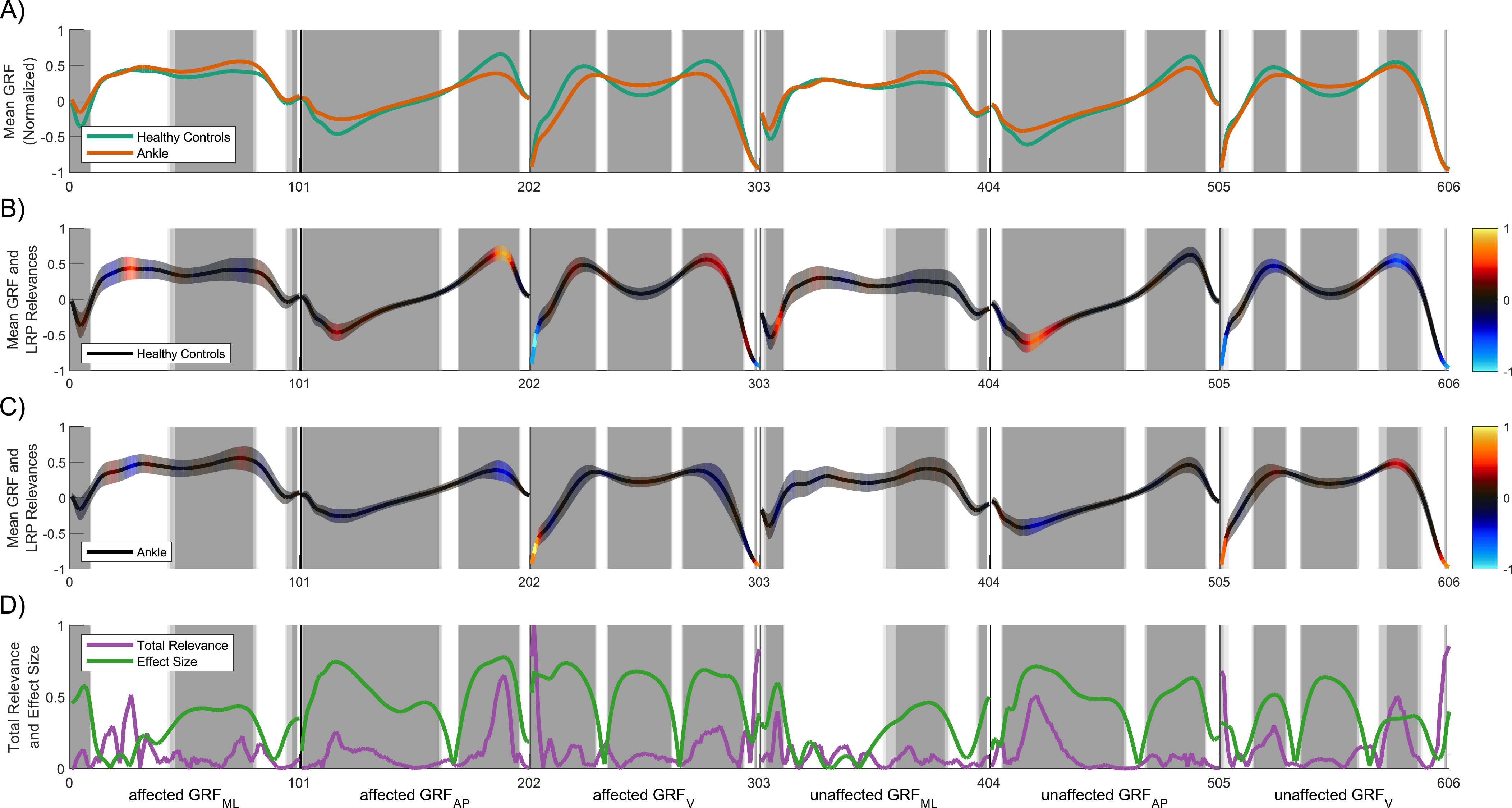} 
    \caption{Result overview for the classification of healthy controls~($HC$) and ankle injury class~($A$) based on min-max normalized GRF signals using an MLP as classifier.}
    \label{img:supl_mlp-norm-NA}
\end{figure}

\newpage
\subsection*{Classification Task: $HC/A$ | Classification method: $SVM$}

\begin{figure}[ht!]
  \centering
	\includegraphics[width=0.85\linewidth]{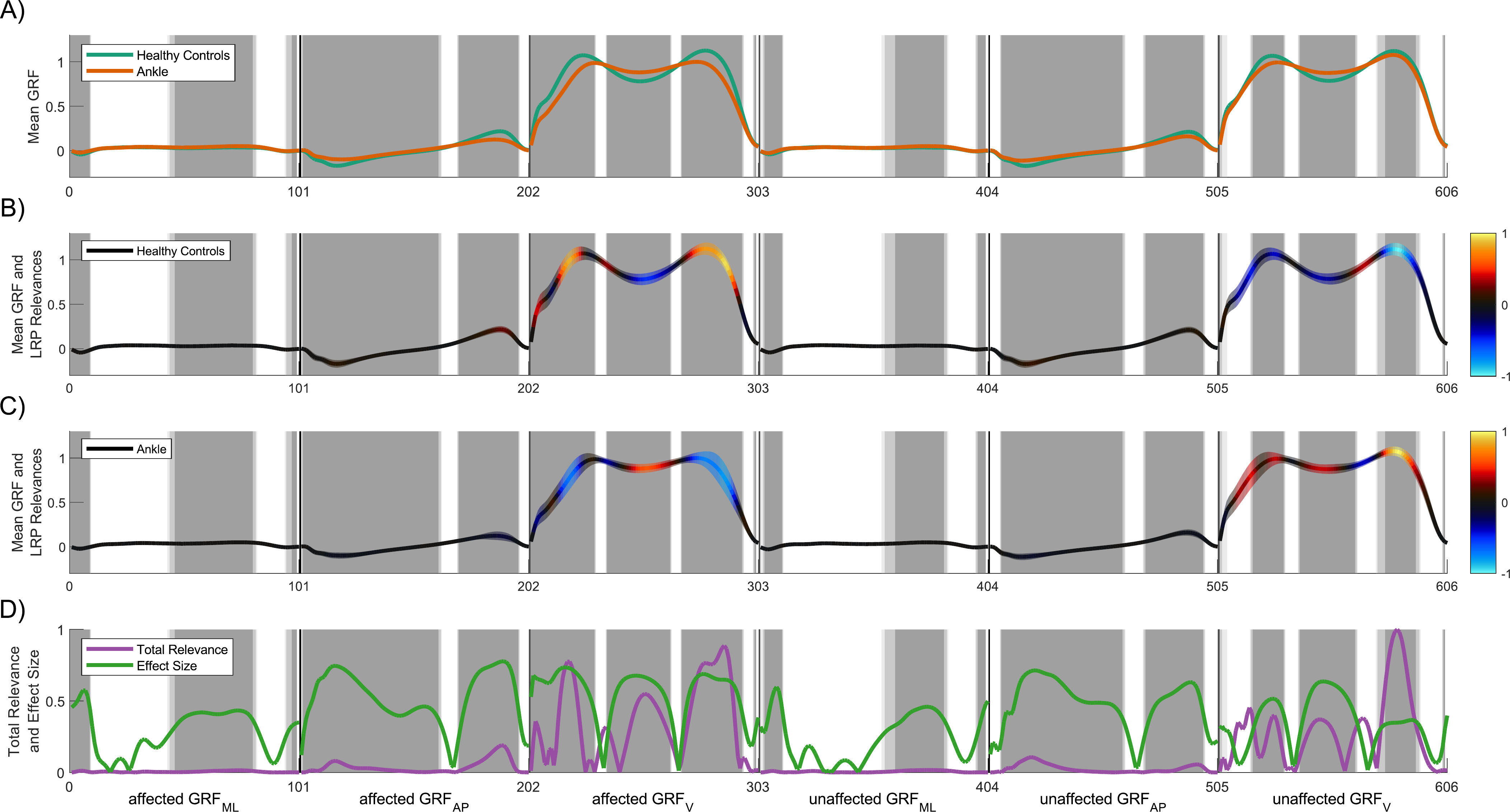} 
    \caption{Result overview for the classification of healthy controls~($HC$) and ankle injury class~($A$) based on non-normalized GRF signals using a SVM as classifier.}
    \label{img:supl_svm-nonorm-NA}
\end{figure}

\begin{figure}[ht!]
  \centering
	\includegraphics[width=0.85\linewidth]{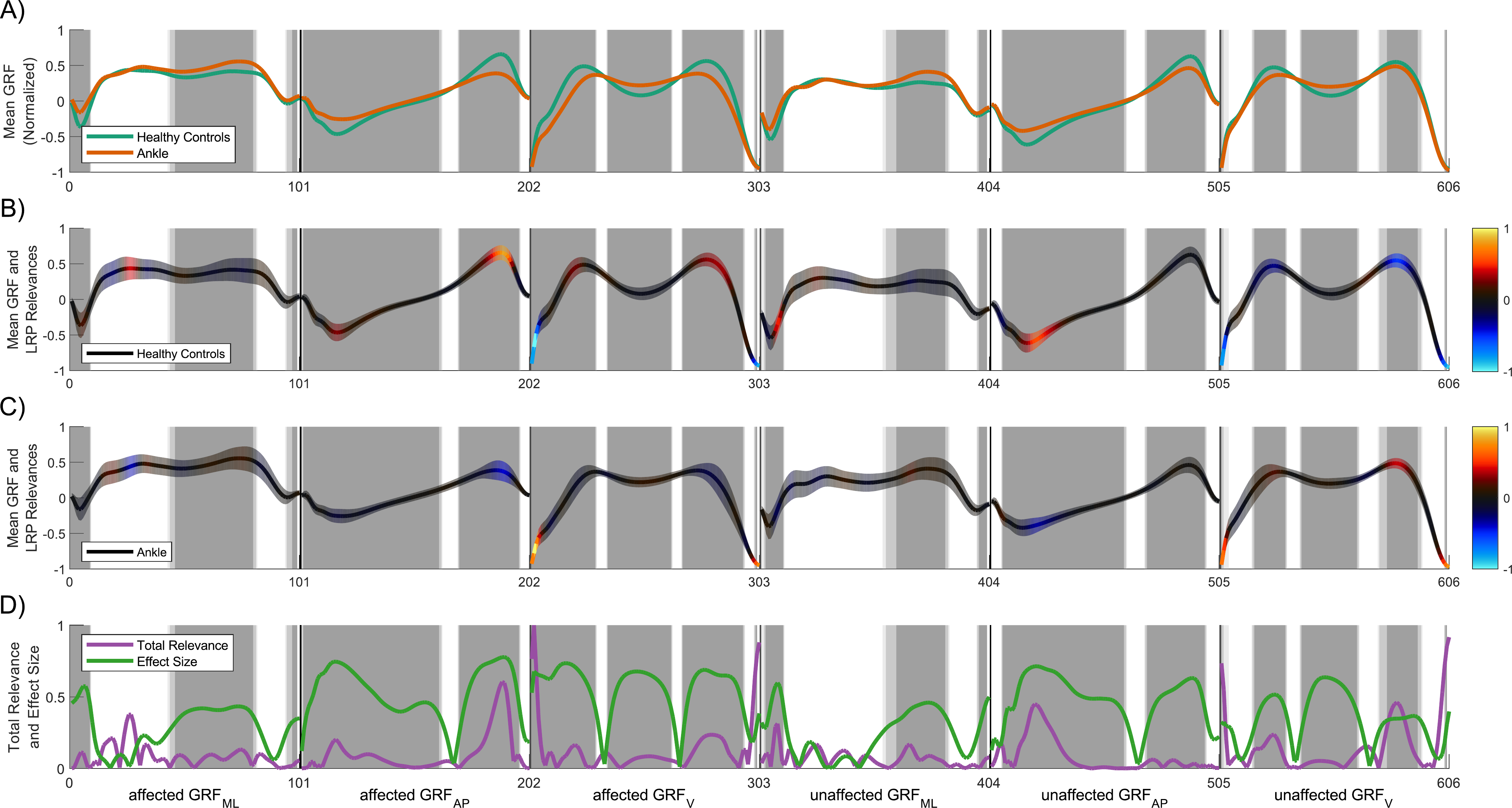} 
    \caption{Result overview for the classification of healthy controls~($HC$) and ankle injury class~($A$) based on min-max normalized GRF signals using a SVM as classifier.}
    \label{img:supl_svm-norm-NA}
\end{figure}

\end{document}